\documentclass[preprint,3p,authoryear]{elsarticle}

\usepackage{amssymb}
\usepackage{amsmath}
\usepackage{amsthm}

\usepackage{graphicx}
\usepackage{epsfig}
\usepackage{enumitem}
\usepackage{natbib}
\usepackage{multirow}
\usepackage{comment}

\usepackage{algorithm}
\usepackage{algpseudocode}

\usepackage{booktabs} 
\usepackage[hyphens]{url}
\usepackage{hyperref}

\usepackage{subcaption} 

\usepackage{mathtools}

\newcommand{\bx}{\mathbf{x}}

\newcommand{\bbeta}{\boldsymbol{\beta}}

\newtheorem{theorem}{Theorem}
\newtheorem{lemma}{Lemma}
\newtheorem{proposition}{Proposition}

\newtheorem{assumption}{Assumption}
\newtheorem{remark}{Remark}

\bibpunct[, ]{(}{)}{,}{a}{}{,}%
\makeatletter
\def\ps@pprintTitle{%
	\let\@oddhead\@empty
	\let\@evenhead\@empty
	\let\@oddfoot\@empty
	\let\@evenfoot\@empty
}

\begin{document}

\begin{frontmatter}



\title{Conditional Generative Modeling for Enhanced Credit Risk Management in Supply Chain Finance} 

\author[label1,label2,label4]{Qingkai Zhang}
\ead{22110690021@m.fudan.edu.cn}

\author[label3]{L. Jeff Hong \corref{cor1}}
\ead{lhong@umn.edu}

\author[label2,label4]{Houmin Yan}
\ead{houminyan@cityu.edu.hk}

\cortext[cor1]{Corresponding author}
 
\affiliation[label1]{organization={School of Management, Fudan University},
             city={Shanghai},
             postcode={200433},
             country={China}}

\affiliation[label2]{organization={Department of Decision Analytics and Operations, College of Business, City University of Hong Kong},
             city={Hong Kong},
             country={China}}
\affiliation[label3]{organization={Department of Industrial and Systems Engineering, University of Minnesota},
	city={Minneapolis},
	postcode={MN 55455},
	country={USA}}
\affiliation[label4]{organization={The Laboratory for AI-Powered Financial Technologies},
	city={Hong Kong},
	country={China}}

%

\begin{abstract}
The rapid expansion of cross-border e-commerce (CBEC) has created significant opportunities for small- and medium-sized sellers, yet financing remains a critical challenge due to their limited credit histories. Third-party logistics (3PL)-led supply chain finance (SCF) has emerged as a promising solution, leveraging in-transit inventory as collateral. We propose an advanced credit risk management framework tailored for 3PL-led SCF, addressing the dual challenges of credit risk assessment and loan size determination. Specifically, we leverage conditional generative modeling of sales distributions through Quantile-Regression-based Generative Metamodeling (QRGMM) as the foundation for risk measures estimation. We propose a unified framework that enables flexible estimation of multiple risk measures while introducing a functional risk measure formulation that systematically captures the relationship between these risk measures and varying loan levels, supported by theoretical guarantees. To capture complex covariate interactions in e-commerce sales data, we integrate QRGMM with Deep Factorization Machines (DeepFM). Extensive experiments on synthetic and real-world data validate the efficacy of our model for credit risk assessment and loan size determination. This study explores the use of generative models in CBEC SCF risk management, illustrating their potential to strengthen credit assessment and support financing for small- and medium-sized sellers.
\end{abstract}




\end{frontmatter}

\section{Introduction}

E-commerce has become an increasingly important component of global trade, with digital platforms enabling businesses of all sizes to reach international customers. Over the past decade, online retail has expanded rapidly: Business-to-Consumer (B2C) e-commerce sales have increased from approximately \$1.5 trillion in 2014 to over \$6 trillion in 2024, exceeding the GDPs of Japan and Germany \citep{WorldBank2019ecommerce,WorldBankEGATE}. Within this boom, cross-border e-commerce (CBEC) has emerged as a rapidly growing segment, increasing from around \$300 billion in 2015 to over \$1.2 trillion in 2024 \citep{MarketResearchFuture2025}.
Despite this scale, access to financing remains a significant barrier, especially for small- and medium-sized sellers. For instance, although over 70\% of Amazon's merchants source products internationally, only 1\% obtain financial support from traditional banks \citep{junglescout2024}. These financing gaps are driven by small- and medium-sized sellers' limited collateral, thin credit histories, and the complexities of cross-border operations \citep{Chiu2024,cai2024supply}.

Supply chain finance (SCF), particularly the third-party logistics (3PL)-led model, has emerged as a promising solution. In this model, 3PL providers collaborate with financial institutions to manage both logistics and financing, offering inventory-backed loans by leveraging real-time in-transit goods as collateral \citep{chen2011joint,ji2023financial}. A salient example is UPS Capital, which combines logistics infrastructure with financial services \citep{huang2019impact}.

Despite the promise of 3PL-led SCF, the practice of CBEC-oriented supply chain finance remains in its early stage, with fragmented adoption and limited standardization across markets. Industry developments reflect this nascent state: for instance, in 2024, the Hong Kong Export Credit Insurance Corporation just launched the market-first trade credit insurance programs specifically tailored to Hong Kong-based CBEC loans \citep{Chiu2024}. This paper focuses on a specific 3PL-led CBEC-SCF model where small- and medium-sized sellers must pre-stock inventory in overseas warehouses to meet consumer expectations for rapid delivery, resulting in substantial working capital pressure.

In this 3PL-led CBEC-SCF model, small- and medium-sized sellers consign goods to 3PL-managed overseas warehouses, which serve as collateral for financial institutions (FI). Lenders then assess credit risk based on the real-time value of in-transit inventory and predictive analytics of sales performance. The loan amount is adaptively adjusted: if sales forecasts indicate high volatility or low market demand, lenders may set loan-to-value ratios as low as 20\% or reject financing entirely, while favorable forecasts may lead to higher loan amounts.
This loan is secured against the products, which are managed within the 3PL's coordinated logistics network. As the products are eventually delivered, sales transactions on e-commerce platforms generate real-time cash flows. The revenue generated from these sales is then allocated primarily to repay the loan (with any surplus then remitted to the seller). This model is akin to the CBEC-SCF models proposed by \cite{shi4450732evaluating} and \cite{ji2023financial}, with the added innovation of lenders utilizing predictive sales analytics for credit risk control.

A central challenge in any lending model (and especially in CBEC-SCF) is effective risk management \citep{qiao2023highlight}. Two risk management questions stand out: (1) \textit{Credit risk assessment} --- how to reliably evaluate the probability of a borrower defaulting on a loan and estimate the potential financial loss and interested risk measures, and (2) \textit{Loan size determination} --- how to set credit limits or loan amounts in line with the borrower's repayment capacity. These two questions are inherently interrelated and jointly shape the overall risk profile of lending decisions: the outcome of one directly affects the other. As such, addressing them in an integrated manner is essential for robust risk management.

In the 3PL-led CBEC-SCF model, where loan repayments depend on future sales revenues, accurate credit risk assessment and loan sizing hinge on modeling the conditional sales distribution \(P(Y \,|\,\bx)\), given contextual features \(\bx\) such as product attributes (e.g., ratings, pricing, category), seller characteristics (e.g., review scores, store tenure, fulfillment efficiency), and market factors (e.g., buyer preferences, industry trends, seasonality). Traditional methods often provide only point estimates like the mean of sales \(Y\) \citep{hastie2009elements, mohri2018foundations}, whereas learning the full conditional distribution $P(Y\,|\,\bx)$ enables a nuanced view of sales variability, including dispersion and tail risk. This richer representation is key for assessing risk measures like the default probability and the expected loss, as it captures how input features $\bx$ influence the full range of potential sales outcomes $Y$, thereby supporting more informed and robust credit decisions.

We identify two promising and distinctive opportunities within this 3PL-led CBEC SCF model. The first opportunity arises from the immense volume and richness of sales data from e-commerce platforms. Such data comes with a variety of easily collectible features, offering a granular view of consumer behavior and sales trends. In contrast, existing credit risk assessment methods in SCF often center on the “small- and medium-sized-enterprise-oriented” factors (e.g., credit histories, financial ratios, and operational metrics) and the “supply-chain-finance-oriented” factors (e.g., core enterprise credit ratings, industry trends, and supply chain concentration) to predict default risk via various classification models \citep{zhu2019forecasting,hou2024predicting,belhadi2025ensemble}. These conventional approaches face several limitations: the sample size is often limited (data about such a diverse range of SCF influencing factors is frequently scattered across multiple entities, especially for small- and medium-sized enterprises
); defaults are highly imbalanced (given the low incidence of bankruptcies), decreasing the accuracy of models \citep{hou2024predicting};
moreover, most of these models assume a fixed credit loan amount, lacking the flexibility required for adaptive loan sizing.

The second opportunity emerges from transformative advances in generative models, which enable learning the entire conditional distribution that plays a critical role in risk quantification. While early machine learning excelled at point prediction, estimating conditional expectations through supervised learning \citep{hastie2009elements, mohri2018foundations}, such approaches inherently obscure outcome variability. Generative models fundamentally redefine this paradigm. Pioneered by architectures like Generative Adversarial Networks \citep[GANs,][]{goodfellow2014generative} and its conditional variations \citep[CGAN,][]{mirza2014conditional} and later advanced through Generative Pre-trained Transformers \citep[GPTs,][]{radford2018improving}, these generative models are designed to directly learn the entire conditional distribution of \(Y\) given \(\bx\).

Given these opportunities, there is now a clear need for an accurate, robust, and efficient generative model that can learn the conditional distribution of the sales for specific products and covariates. Despite significant progress in generative models, as highlighted by \cite{hong2023learning}, most existing generative models are primarily designed for high-dimensional data (e.g., images, texts, audios, and videos), with an emphasis on generating outputs that are visually or semantically realistic rather than capturing the full underlying data distribution. In the context of SCF risk management, however, accurately and robustly acquiring the conditional distribution of low-dimensional variables like sales volume is critical. Poor forecasting can lead to significant misjudgments: overestimating the right tail of the distribution may result in a loan amount that exceeds the seller's actual repayment capacity, heightening default risk, while underestimating sales potential might leave the seller under-financed, curtailing its growth and causing lenders to miss safe lending opportunities. 

However, popular generative models based on GAN structures, such as the CWGAN with gradient penalty \citep[CWGAN-GP,][]{athey2021using}, often do not exhibit the desired level of accuracy and robustness in this setting \citep{hong2023learning}. The current state-of-the-art generative models for accurately and robustly capturing low-dimensional conditional distributions is Quantile-Regression-based Generative Metamodeling \citep[QRGMM,][]{hong2023learning}. In addition to its precision and robustness, QRGMM provides a significant advantage of efficiently generating a substantial number of random observations in real time, which is particularly valuable for estimating risk measures, facilitating the use of Monte Carlo methods to calculate various risk measures of interest, thereby enhancing our flexibility for risk measures estimation while minimizing estimation errors caused by sampling variance.
In this paper, we explore the application of QRGMM for SCF credit risk management, aiming to leverage its capabilities to overcome the limitations inherent in traditional credit risk assessment methods. It is important to note that, throughout the paper, we use the term ``generative model'' in its broader statistical sense, referring to methods that approximate probability distributions and generate random observations, not limited to recent deep-learning-based architectures.

While QRGMM offers a promising tool for learning the conditional distribution of sales volume,  its application to SCF credit risk management is far from straightforward and presents several non-trivial challenges. Conventional SCF risk models are largely grounded in binary classification, estimating static default probabilities that fail to account for varying loan sizes or richer risk profiles. Even recent generative approaches using GANs focus on data augmentation for training classification models, remaining confined to classification frameworks and failing to capture responsive loan size-risk relationships \citep{zhang2025credit}. In real-world SCF settings, financial institutions demand adaptive tools that capture how risk measures evolve with different loan levels, enabling real-time, data-driven loan sizing. Bridging generative modeling with practical SCF credit risk management thus requires more than data augmentation: it demands a unified framework that translates the full conditional sales distribution output of QRGMM into actionable, loan-level-specific risk measures.

To meet this challenge, we develop a functional formulation as a bridge that effectively links the learned sales distribution to various credit risk measures relevant to SCF. Specifically, we introduce a unified framework that systematically formulates key risk measures as explicit functions of the loan level: \( r_1(l) \) for default probability, \( r_2(l) \) for expected loss, and the versatile \( r_3(l) \) for more generalized risk measures. Importantly, this formulation accommodates a wide range of risk measures within a unified and coherent structure, eliminating the need to construct and analyze cumbersome and inefficient separate static models for each risk measure individually, and thereby significantly enhancing decision-making flexibility. This flexibility allows practitioners to adaptively assess diverse credit risk measures across different loan sizes in real time, a capability that has been largely absent in prior methodologies.

Nevertheless, ensuring the reliability of this generative approach requires robust theoretical guarantees, which introduce another separate set of non-trivial challenges. While QRGMM ensures convergence in distribution for the estimated sales distribution, this alone does not directly guarantee the accurate estimation of many complex credit risk measures. In particular, key risk measures like \( r_2(l) \) and \( r_3(l) \) involve non-linear, non-smooth transformations of the sales distribution, often expressed through complex expectations. Such structures prevent the direct application of standard convergence results from QRGMM, as ensuring accurate risk measures estimation requires uniform control not just over distribution estimates but also over the induced expectations across all loan levels. We overcome these difficulties by carefully exploiting the structures inherent to the SCF setting. Finally, building upon the linear quantile regression model of \cite{hong2023learning}, we establish uniform convergence results for all formulated risk measures, providing a solid theoretical foundation for our unified framework, ensuring its utility and credibility for SCF credit risk management.

In practice, the above linear assumption seems to be rigid, potentially overlooking the complex relationships between product sales and their covariates in SCF. This complexity arises from the multifaceted nature of sales, which are influenced not only by the products and their attributes but also by the combinatorial effects of products, stores, and other contextual factors that shape sales distributions. For example, in a cross-border e-commerce company that we worked with, product selection and sales strategies executed by diverse groups significantly affect outcomes due to varied skills and levels of collaboration. Traditional models, such as those used in recommendation systems like matrix factorization \citep[MF,][]{lee2000algorithms} and factorization machines \citep[FM,][]{rendle2010factorization}, primarily focus on predicting expected values rather than distributions. They are effective for learning combinatorial features but fall short in capturing the full range of the targeted distribution. To address this, we propose a novel approach that integrates QRGMM with deep factorization machines \citep[DeepFM,][]{guo2017deepfm}, enhancing accuracy in predicting target distributions. Extensive numerical experiments with both synthetic and real-world data validate our method's ability to capture complex covariate interactions and precisely predict the conditional distribution of sales, which facilitates the estimation of pertinent credit risk measures and supports well-informed decision-making processes.

The main contributions of this paper can be summarized as follows. 
\begin{itemize}
	\item We develop a generative modeling framework for supply chain finance that provides a distributional characterization of sales and a unified analytical formulation linking the sales distribution to credit risk assessment and loan sizing.
	\item We establish uniform convergence results for a broad class of credit risk measures, provide theoretical guarantees that extend beyond distributional convergence, and ensure consistency across the loan decision space.
	\item We extend QRGMM with DeepFM to capture complex feature interactions while retaining sampling efficiency, and demonstrate through extensive experiments that it outperforms QRGMM and alternative generative approaches in distributional accuracy, risk measures estimation, and stability.
\end{itemize}

The remainder of the paper is structured as follows. Section 2 presents the 3PL-led CBEC-SCF model and our functional risk formulation, followed by Section 3, which introduces the quantile-regression-based generative model for SCF risk management, and Section 4 details its theoretical properties. Section 5 extends our framework by integrating DeepFM to effectively capture complex covariate interactions. Comprehensive numerical experiments with both synthetic and real-world data are presented in Section 6 to validate our approach. Finally, Section 7 concludes the paper and discusses potential avenues for future research.

\section{3PL-led CBEC SCF Model and Functional Risk Formulation}

This section presents both the business model of 3PL-led CBEC-SCF and its functional risk formulation. 
Section~\ref{sec:Bus_mode} details the financing model and illustrates the flow of goods, cash, data, and control among the key actors. 
Section~\ref{sec:riskformulation} then develops the mathematical risk formulation, linking the operational 
mechanism to quantitative risk analysis.

\subsection{3PL-led CBEC SCF Model}\label{sec:Bus_mode}

We begin by outlining the 3PL-led CBEC supply chain finance model, which describes the interactions among the seller, the e-commerce platform, the third-party logistics provider, and the financial institution. Specifically, we explain how this model operates by tracing the flows of goods, cash, data, and control, and by highlighting the financial institution's role in coordinating financing and data-driven risk management. Finally, we contrast this approach with other SCF practices to clarify its distinctive features and positioning.

In practice, the 3PL-led CBEC SCF process unfolds as follows. The seller first pre-positions goods in an overseas warehouse operated by the 3PL provider. These goods are pledged as collateral, and the warehouse provides real-time visibility into inventory movements and stock levels. Customers place orders on the e-commerce platform and pay the platform directly. The 3PL warehouse then fulfills the orders. The platform routes the seller's proceeds into a repayment account controlled by the financial institution or by a payment service provider (PSP) appointed by the financial institution. From this account, principal and interest are automatically deducted before any surplus is remitted to the seller, which is a standard “cash-dominion” or lockbox arrangement in asset-based lending. In parallel, the financial institution advances loans to the seller to support working-capital needs, with loan sizes determined on the basis of projected sales and the value of the pledged inventory. 

This structure does not require direct information sharing between the 3PL and the e-commerce platform. Instead, the financial institution orchestrates the process: it obtains the seller's authorization to access the platform's sales and settlement data through the platform's official application programming interfaces, e.g., Amazon's Selling Partner API \citep{AmazonSellingPartnerAPI}. Without access to these data, the financial institution cannot properly assess repayment risk and may therefore decline to extend credit to the seller. Beyond that, the financial institution can also build a large database of historical sales records through licensed dataset purchases, strategic partnerships, or experience accumulated from previous lending, which provides a valuable corpus for credit-risk modeling.

\begin{figure*}[t]
	\centering
	\vspace{-10pt}
	\includegraphics[width=0.95\linewidth]{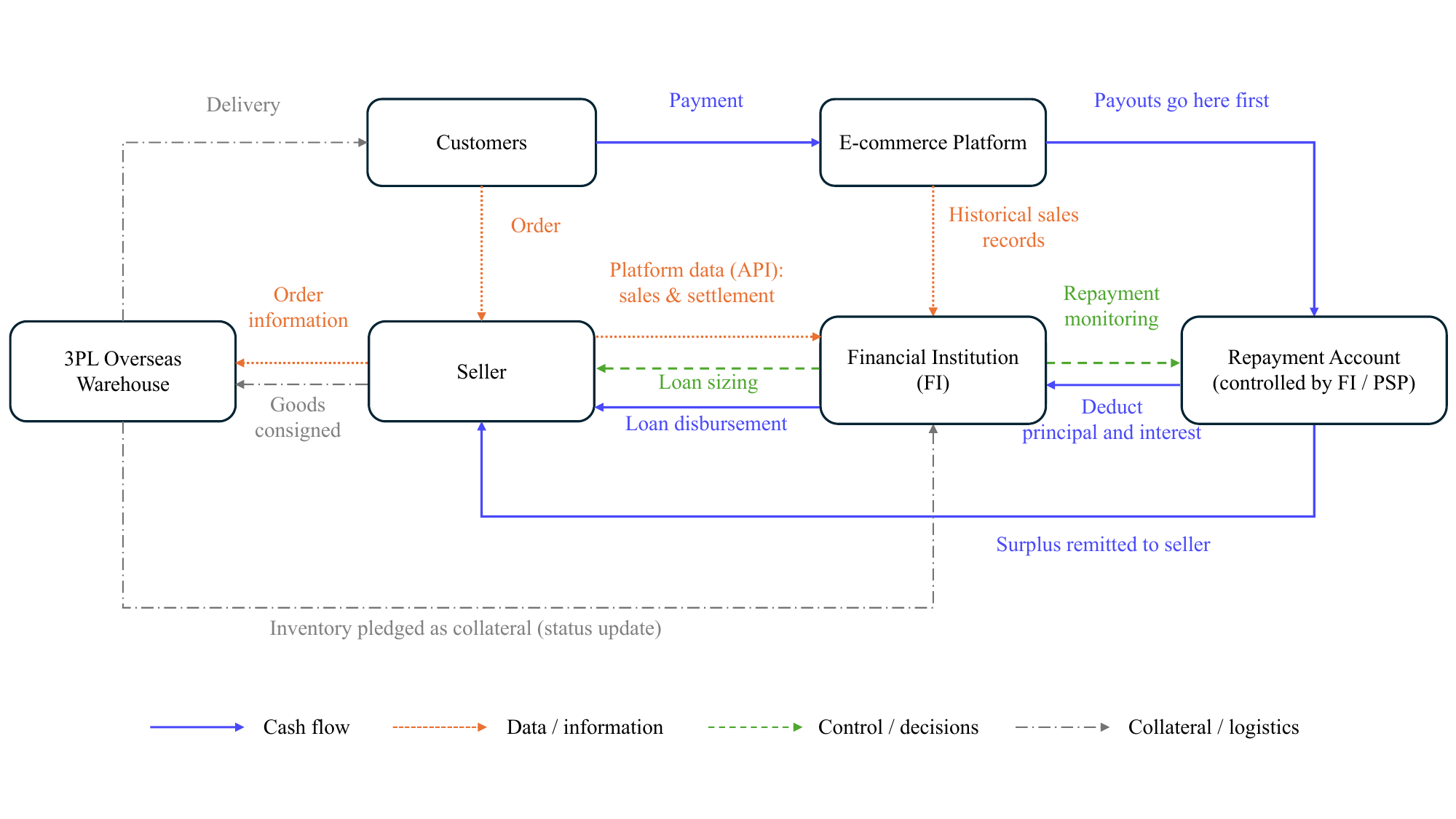}
	\vspace{-25pt}
	\caption{Schematic of the 3PL-led CBEC-SCF process.}
	\label{fig:3pl_schematic_color}
\end{figure*}

This financial-institution-led design of data and cash-flow control sidesteps potential data-sharing frictions between independent logistics providers and platforms. It also accommodates cases in which logistics or platform operators offer their own financing, such as the supply chain finance services of UPS Capital or Amazon Lending. Figure~\ref{fig:3pl_schematic_color} illustrates the corresponding flows of goods, cash, data, and control among the seller, platform, 3PL provider, financial institution, and customers.

Compared with other SCF models, the 3PL-led CBEC structure offers several advantages:

\begin{itemize}
	\item 
	It enables financing even for sellers with limited credit histories. 
	Unlike buyer-led receivables programs (reverse factoring) that depend on the creditworthiness of large buyers, this model enables financing as long as the seller's e-commerce sales can be verified and monitored.
	\item 
	It allows data-driven and adaptive collateralization.
	Traditional inventory- or receivables-based lending in domestic trade typically relies on a static borrowing base. By contrast, the CBEC model uses e-commerce sales forecasts and goods in transit or stored in 3PL warehouses as collateral, broadening financing access for smaller sellers. 
	\item 
	It remains applicable to both independent and captive settings. 
	While some logistics or platform operators provide their own lending, such as UPS Capital or Amazon Lending, the financial-institution-led control of data and cash flows captures the broader range of CBEC transactions where logistics and platforms are organizationally separate.
\end{itemize}

This model aligns with the broader movement toward data-driven inclusive finance, in which alternative information substitutes for traditional collateral or credit files and expands financing opportunities for small- and medium-sized e-commerce sellers.

\subsection{Functional Risk Formulation}\label{sec:riskformulation}

Building on the operational description of the 3PL-led CBEC-SCF model, we now formulate the problem from the perspective of the financial institution. Consider a small- and medium-sized seller engaged in cross-border e-commerce that sells products via an online platform and requires 3PL-led supply chain financing. In this setting, the goods are managed by the 3PL provider during transit, and the revenue generated from eventual sales is primarily allocated to repay the loan. As the financial institution partnering with the 3PL, our goal is to determine an appropriate loan level \( l \) based on both the collateral value and projected sales performance, ensuring adequate risk control while aligning financing with the seller's revenue potential.

Let $Y$ represent the random sales volume (in units) of the product over a fixed time horizon, 
let \( r \) denote the net unit revenue per product sold, adjusted for shipping, handling, and platform operating costs. The effective revenue available for loan repayment is thus \( rY \). We define the financial institution's loss function as
\begin{equation} 
	L(l) = (l - rY)^+, \label{eq:loss}
\end{equation}
where $(x)^+ = \max\{x,0\}$. Intuitively, $L(l)$ captures the shortfall incurred when the loan level $l$ exceeds the total revenue generated from sales, thereby encapsulating the risk associated with insufficient sales performance. This formulation abstracts the problem to its essential components, enabling us to evaluate how the loss (and hence the risk) varies as a function of the loan level.

The financial institution seeks to evaluate risk measures across the loan decision space \( l \in [0, \bar{l}] \), where \(\bar{l}\) denotes the maximum loan amount (often equal to the total collateral value to prevent arbitrage opportunities). We introduce a family of key risk measures that vary with \( l \), offering flexibility in capturing different dimensions of credit risk. The first is the \textit{default probability}, which is defined as:
\begin{equation*} \label{eq:r1}
	r_1(l) = \Pr\{L(l) > 0\} = \Pr\{l - rY > 0\} = \Pr\{Y < l/r\},
\end{equation*}
representing the likelihood that sales revenue fails to cover the loan. The second is the \textit{expected loss}, which is given by:
\begin{equation*} \label{eq:r2}
	r_2(l) = \mathbb{E}[L(l)] = \mathbb{E}[(l - rY)^+].
\end{equation*}
quantifying the average shortfall in the event of default.

These two risk measures are fundamental to credit risk evaluation. For example, widely used credit risk measure, such as the Probability of Default (PD), is directly given by $r_1(l)$, and the Loss Given Default (LGD) can be computed as
\begin{equation*} \label{eq:LGD}
	\text{LGD} =\frac{\mathbb{E}[L(l) \,|\, L(l) > 0]}{l} = \frac{\mathbb{E}[L(l)]}{l\,\Pr\{L(l)>0\}} = \frac{r_2(l)}{l\,r_1(l)}.
\end{equation*}
Moreover, we might be interested in the probability and the expected loss when the loss exceeds a threshold level $\xi$, where $0<\xi<l$. Then,
\begin{align*}
	\Pr\{L(l)>\xi\} &= \Pr\{(l - rY)^+ > \xi\} \nonumber\\
	&= \Pr\{l - rY > \xi\} \nonumber\\
	&= \Pr\{Y < (l-\xi)/r\} \nonumber\\
	&= r_1(l-\xi),\label{eq:threshold_prob}
\end{align*}
and 
\begin{align*}
	\mathbb{E}\left[L(l)\,\mathbb{I}\{L(l)>\xi\}\right] 
	= &~ \mathbb{E}\left[(l - rY)\cdot\mathbb{I}\{l - rY > \xi\}\right] \nonumber\\
	= &~ \mathbb{E}\left[(l-\xi - rY)\cdot\mathbb{I}\{rY < l-\xi\}\right] \nonumber\\
	&+ \xi\,\Pr\{rY < l-\xi\} \nonumber\\
	= &~ r_2(l-\xi) + \xi\,r_1(l-\xi). 
\end{align*}
This demonstrates that \( r_1(l) \) and \( r_2(l) \) serve as building blocks for broader risk measures. 

In addition, to further broaden the scope of our framework, we propose a more generalized risk measure, \( r_3(l) \), defined as:
\[
r_3(l) = \mathbb{E} \left[ g_l(Y) \cdot \mathbb{I}\{Y < a(l)\} \right],
\]
where \( g_l(\cdot) \) is a loan-level-dependent function that quantifies specific risk forms, and \( a(l) \) is a threshold that varies with \( l \), often tied to the default point (e.g., \( a(l) = l/r \)). This formulation further enhances the modeling flexibility, enabling \( r_3(l) \) to adapt to a wide range of risk measures by tailoring \( g_l(\cdot) \) and \( a(l) \).
\begin{itemize}[]
	\item \textit{Recovering basic measures.} Both $r_1(l)$ and $r_2(l)$ are special cases of $r_3(l)$. By setting \( g_l(Y) = 1 \) and \( a(l) = l/r \), we obtain $r_1(l)$; and by setting \( g_l(Y) = l - rY \) and \( a(l) = l/r \), we arrive at $r_2(l)$.
	\item \textit{Nonlinear extensions.} \( r_3(l) \) extends to more complex risk measures where the loss function takes on nonlinear forms. For instance, choosing \(g_l(Y)=(l-rY)^2\) and \(a(l)=l/r\), we have \( r_3(l) = \mathbb{E}\left[\left(\left(l - rY\right)^+\right)^2\right] \), which amplifies larger losses quadratically and emphasizes the severity of extreme shortfalls.
	\item \textit{Distortion-based measures.} Distortion risk measures often take the form of $\mathbb E_{\mathbb Q}\!\left[(l-rY)^+\right]$, where \(\mathbb Q\) is a new distorted distribution of the sale $Y$ \citep{wang2000class, prelec1998probability}. Let \(\mathbb P\) denote the baseline distribution. Then 	
	\[
	\mathbb E_{\mathbb Q}\!\left[(l-rY)^+\right]=\mathbb E_{\mathbb P}\big[\left(l-rY\right)\cdot L(Y)\cdot\mathbb{I}\left\{Y<l/r\right\}\big],
	\]
	where \(L(Y)\) denotes the likelihood ratio. In cases where \(L(Y)\) may be computed (at least empirically), the distorting risk measure fits into the framework of \( r_3(l)\) by setting \(g_l(Y)=(l-rY)\cdot L(Y)\) and \(a(l)=l/r\).
	\item \textit{Revenue-adjusted losses.} So far, we assume that the total revenue of the seller is $rY$, which is linear in the sales $Y$. However, sellers may adjust the price of the product based on sales. In that case, we may let $v(Y)$ be the revenue, and the loss becomes $l-v(Y)$. Furthermore, assuming $v(Y)$ is an increasing function of the sales $Y$, then $\left\{l-v(Y)>0\right\}$ is equivalent to $\left\{Y<v^{-1}(l)\right\}$. Then, to compute the expected loss, we may set \(g_l(Y)=l-v(Y)\) and \(a(l)=v^{-1}(l)\) in the definition of \( r_3(l)\).
\end{itemize}	

Thus, \( r_3(l) \) provides a unified and versatile framework for capturing sophisticated risk measures, ensuring broad applicability in SCF applications.

\begin{remark}
	Here, our attention is focused on risk measures expressed in terms of the loan level \(l\) and the sales outcome \(Y\), which helps clarify the unified formulation of $r_1(l)$, $r_2(l)$, and $r_3(l)$. In the next section, the scope will be broadened to incorporate covariates into the generative model for sales $Y(\bx)$, where $\bx$ may represent product attributes (e.g., ratings, pricing, category), seller characteristics (e.g., review scores, store tenure, fulfillment efficiency), and market factors (e.g., industry trends, buyer preferences, seasonality). Incorporating these features allows us to capture product- and seller-specific heterogeneity, account for broader market influences, and support more refined credit risk assessments, including within-industry comparisons.
\end{remark}

Now, the key challenge becomes efficiently computing these risk measures across all \( l \in [0, \bar{l}] \). The crux of this challenge is to learn \textit{the distribution of $Y$}. Suppose we have a generative model $\hat{Y}$ capable of producing $K$ independent and identically distributed observations $\{\hat{Y}_i\}_{i=1}^K$ rapidly, such that the distribution of $\hat{Y}$ closely approximates that of $Y$. Then, we can estimate these risk measures using Monte Carlo simulation \citep{hong2014monte}:
\begin{align}\label{eq:mcestimation}
	&\hat{r}_{1,K}(l) = \frac{1}{K} \sum_{i=1}^K \mathbb{I}\{l - r \hat{Y}_i > 0\}, \nonumber \\
	& \hat{r}_{2,K}(l) = \frac{1}{K} \sum_{i=1}^K \left(l - r\hat{Y}_i\right)^+, \nonumber \\
	& \hat{r}_{3,K}(l) = \frac{1}{K} \sum_{i=1}^K g_l(\hat{Y}_i) \cdot \mathbb{I}\{\hat{Y}_i < a(l)\}. 
\end{align}
Therefore, the ability to generate these observations rapidly and accurately is crucial for evaluating risk measures across a continuum of loan levels. This motivates our use of a generative model to capture the full probability distribution of $Y$, enabling efficient and reliable risk assessments across all potential loan levels.

Our proposed functional risk assessment framework revolutionizes traditional SCF credit risk estimation, which typically rely on methods like logistic regression or support vector machines to estimate a single risk measure, often the probability of default, for a fixed loan amount. By modeling risk measures such as $ r_1(l) $, $ r_2(l) $, and the versatile $ r_3(l) $ as functions of loan size $ l $, our functional approach provides a unified and comprehensive risk profile across the range of loan levels. Supported by a fast and reliable generative model and the Monte Carlo method, this framework enables financial institutions to estimate diverse credit risk measures and make informed loan sizing decisions, aligning financing with sales potential and significantly enhancing SCF risk management compared to static, single-point traditional methods.

\section{Constructing a Conditional Generative Model for Sales Distribution}

In this section, we first formulate the problem of constructing a conditional generative model for sale distribution. We then present QRGMM as an effective generative modeling technique that meets the specific requirements of SCF risk management. 

\subsection{Problem Formulation}\label{sec:prob_Formulation}

When a small- and medium-sized seller raises financing through the 3PL-led CEBC-SCF model, the sales revenue of its product will generally be prioritized to repay the loan. Hence, the random variable representing product sales, \( Y(\mathbf{x}) \), directly determines the loan repayment outcome. Here, the multi-dimensional vector $\mathbf{x} \in \mathcal{X}\subset \mathbb{R}^p$ denotes the associated covariates, such as product features, seller characteristics, and other auxiliary information. We collect $n$ historical observations $\{(\mathbf{x}_i, y_i)\}_{i=1}^n$ from e-commerce platforms, where $y_i$ are realizations of $Y(\mathbf{x}_i)$. To effectively manage credit risk in SCF, our objective is to develop a generative model $\hat{Y}(\bx)$ utilizing the dataset $\{(\bx_i,y_i)\}_{i=1}^n$. This model should accurately and robustly generate random observations that replicate the conditional distribution of $Y(\bx)$ given the covariates $\bx$, while also ensuring rapid generation.

While generative models have significantly propelled advancements in areas such as image and video synthesis as well as language generation \citep{goodfellow2014generative, radford2018improving, ho2020denoising, liu2024sora}, many of them are designed for high-dimensional data, focusing on visual or semantic realism rather than \emph{accurately capturing low-dimensional distributions}, especially in the tails. Some models like GANs and variational autoencoders \citep[VAE,][]{kingma2013auto} also suffer from mode collapse issues \citep{saxena2021generative}, which reduces data diversity and fails to replicate the true distribution accurately, while others (e.g., diffusion-based approaches \citep{ho2020denoising,song2020denoising}) are too slow for large-scale Monte Carlo simulation. In contrast, QRGMM \citep{hong2023learning} is designed to address these issues, offering a compelling choice for SCF applications.

\subsection{The QRGMM Algorithm}
The foundation of QRGMM lies in the \textit{inverse transform method}, where a uniform random variable \(U\) on \((0,1)\) is mapped into a random observation from the target distribution by applying the conditional \textit{inverse cumulative distribution function (CDF)} (also known as quantile function) $F_Y^{-1}(\tau \,|\, \mathbf{x})$. QRGMM approximates this inverse CDF by fitting \textit{quantile regressions} on predefined grid points and then interpolating for intermediate quantiles.
QRGMM operates in two stages: an \textit{offline learning}  stage where quantile regressions are fitted using the dataset \(\{(\mathbf{x}_i, y_i)\}_{i=1}^n\) across a grid of quantile levels evenly spaced in the interval (0,1), and an \textit{online generation} stage that generates uniform random variables and transforms them into target random variables by applying the fitted quantile function with linear interpolation. Algorithm~\ref{alg:gmm} outlines the procedure.

\begin{algorithm}
	\caption{QRGMM}\label{alg:gmm}
	\begin{algorithmic}
		\State \textbf{Offline Stage:}
		\begin{itemize}
			\item Collect dataset \(\{(\mathbf{x}_i,y_i)\}_{i=1}^n\).
			\item Choose an integer \(m\), and discretize \([0,1]\) by \(\tau_{j}=\tfrac{j}{m}, \; j=1,\ldots,m-1\).
			\item Fit quantile regression models \(\hat{Q}_{\tau_j}(\mathbf{x})\) for all \(j=1,\ldots,m-1\).
		\end{itemize}
		
		\State \textbf{Online Stage:} 
		\begin{itemize}
			\item Given new covariates \(\bx\).
			\item Generate \(u_1,\ldots,u_K\) i.i.d.\ from \(\mathrm{Uniform}\,(0,1)\).
			\item For \(k=1,\ldots,K\), output $\hat{Y}_k(\bx)$ as follows:
			for $u_k \in [\tau_j, \tau_{j+1})$ with $j = 1,\ldots,m{-}2$, 
			\[
			\hat{Y}_k(\bx) = \hat{Q}_{\tau_j}(\bx) + m (u_k - \tau_j)\left[\hat{Q}_{\tau_{j+1}}(\bx) - \hat{Q}_{\tau_j}(\bx)\right].
			\]
			otherwise,
			\[
			\hat{Y}_k(\bx) =
			\begin{cases}
				\hat{Q}_{\tau_1}(\bx), & u_k < \tau_1, \\
				\hat{Q}_{\tau_{m-1}}(\bx), & u_k \ge \tau_{m-1}.
			\end{cases}
			\]
		\end{itemize}
	\end{algorithmic}
\end{algorithm}

Below, we elaborate on the key features of QRGMM in detail, highlighting why it is a compelling choice for risk management in SCF.

\begin{itemize}[leftmargin=2em]
	\item \textit{Accuracy and Robustness:} 
	By explicitly modeling conditional quantiles at multiple levels, QRGMM learns the \textit{full shape} of a distribution rather than focusing merely on mean predictions or specific performance measures. This quantile-based approach inherently avoids issues such as mode collapse, and it provides more reliable estimates of the tails of the distribution which have more impact on the SCF credit risk. Unlike adversarial models such as GANs that require solving challenging saddle-point optimization (notoriously difficult to tune), QRGMM merely involves fitting quantile regressions and selecting the main parameter, the number of grid points $m$, 
	resulting in a simpler and more stable training process.
	
	\item \textit{Speed:} 
	Numerical results in \cite{hong2023learning} show that QRGMM can generate a lot of observations in an extremely short time, e.g., \textit{100{,}000 random observations in under 0.01 seconds}. This efficiency stems from its reliance on simple linear interpolation in the online generation stage, after the quantile regressions have been fitted offline. Such rapid sampling is particularly valuable in SCF risk management, as it enables real-time generation of extensive observations, significantly reducing the computational cost of sample generation. This allows us to generate a substantial number of observations to support the Monte Carlo method, computing various risk measures of interest with enhanced flexibility, while minimizing estimation errors caused by sampling variance.

\end{itemize}

By framing SCF risk management through the lens of the full conditional distribution of product sales, we underscore the need for a generative model that is not only accurate and robust but also computationally efficient. QRGMM effectively meets these requirements, providing a promising tool for the integrated framework that links credit risk estimation with loan sizing decisions.

However, to fully realize the practical potential of QRGMM in SCF applications, it is essential to go beyond its generative capability and rigorously establish how its learning outcomes translate into reliable credit risk measures estimation. Specifically, building on the problem formulation in Section \ref{sec:riskformulation}, the next critical step, which we address in the following section, is to provide strong theoretical guarantees on the estimators of risk measures.
This ensures that the QRGMM-based framework is not only practically effective but also theoretically sound, setting the stage for robust real-world deployment.

\begin{remark}
	While QRGMM is well suited for univariate sales modeling in the SCF setting, two practical limitations are worth noting for potential future extensions and other applications. First, extending the method to multivariate outcomes is nontrivial because a standard and widely accepted definition of a multivariate quantile function is lacking; sequentially modeling each dimension is possible but may introduce error propagation and slower computation, especially when the dimensionality is high. Second, because QRGMM estimates conditional quantiles on a grid of quantile levels and constructs the full quantile function via interpolation without explicitly imposing monotonicity constraints, the resulting curve may occasionally violate its natural ordering (e.g., a higher quantile taking a smaller estimated value). When such non-monotonic patterns arise, the rearrangement procedure of \citet{chernozhukov2010quantile} provides a simple post-processing step that restores monotonicity without distorting its statistical validity. A more detailed discussion of these issues of QRGMM can be found in \citet{hong2023learning}.
\end{remark}

\section{Asymptotic Analysis of Risk Measures Estimation via QRGMM}\label{sec:theory}

In this section, we establish theoretical guarantees for the estimators of $r_1(l)$, $r_2(l)$, and $r_3(l)$ for all $l\in[0,\bar{l}]$, thereby ensuring asymptotic validity for a broad range of SCF risk measures estimation through QRGMM. Prior to proceeding, we emphasize several key considerations.

\paragraph*{\textbf{Model Assumption}} In this section, we adopt the linear quantile regression model assumption of \cite{hong2023learning}, which lays the theoretical groundwork for analyzing the asymptotic properties of QRGMM. Suppose $Y(\bx)$ is a continuous random variable on its support given the input $\bx$. Its conditional distribution and density function are denoted as $F_Y(\cdot\,|\,\bx)$ and $f_Y(\cdot\,|\,\bx)$. The linear quantile regression model assumption is stated as follows.

\begin{assumption}\label{ass:cons1} $F^{-1}_Y(\tau\,|\,\bx)=\bbeta(\tau)^\intercal\bx,~\forall ~\tau\in(0,1)$ and $\bbeta(\tau)$ is continuous in $\tau$.
\end{assumption}

Assumption~\ref{ass:cons1} posits that the linear quantile regression model accurately represents the true model of the conditional quantile function of \(Y\) given \(\bx\) for all \(\tau\) in \((0,1)\), implying that \(F^{-1}_Y(\tau\,|\,\bx)\) is linear with respect to the covariates \(\bx\). Although Assumption~\ref{ass:cons1} may seem restrictive, it is widely used in practice due to its simplicity and interpretability. This assumption is foundational not only because it simplifies the theoretical framework but also because it is underpinned by a substantial literature on classical quantile regression, which provides the necessary theoretical foundation for establishing the asymptotic analysis. Furthermore, in conjunction with domain knowledge, it can be expanded to capture nonlinearity by substituting $\bx$ with a set of nonlinear basis functions $\boldsymbol{b(\bx)}$. In terms of theoretical analysis, the theoretical results can be readily extended when employing basis functions. Supporting evidence from numerical experiments by \cite{hong2023learning} demonstrates that even when the linear model assumption does not strictly hold, the approximation using basis functions still produces favorable outcomes. In Section~\ref{sec:extension_qrgmm}, we will explore an extension of the linear model to the DeepFM model, enhancing its ability to capture complex combinatorial features. 

Under Assumption~\ref{ass:cons1},  the quantile function model $\hat{Q}_{\tau_j}(\bx)$ in Algorithm~\ref{alg:gmm} can then be obtained by $\hat Q_{\tau_j}(\bx)=\hat\bbeta({\tau_j})^\intercal\bx$ for all $j=1,\ldots,m-1$, where $\hat\bbeta(\tau_j)$ is the estimation of $\bbeta(\tau_j)$ using the linear quantile regression on the dataset  $\{(\bx_i,y_i)\}_{i=1}^n$ such that
\begin{equation*}
	\hat\bbeta({\tau_j})=\underset{\bbeta}{\text{argmin}}\sum_{i=1}^n\rho_{\tau_j}(y_i-\bbeta^\intercal\bx_i),\quad j=1,2,\ldots,m-1.
\end{equation*}
where $\rho_\tau(u)=\left[\tau-I(u\le 0)\right]u$ is the pinball loss function of quantile regression, this loss penalizes over- and under-predictions asymmetrically, guiding the model to estimate the desired quantile. And the above optimization problem can be efficiently solved via linear programming \citep{koenker2005}.

\paragraph*{\textbf{Sources of Error}} Our credit risk measures estimation procedure builds upon QRGMM to produce a generative model $\hat{Y}(\bx)$ that approximates the distribution of $Y(\bx)$. Given $\hat{Y}(\bx)$, we then rely on Monte Carlo simulation to estimate risk measures by generating $K$ observations $\{\hat{Y}_k(\bx)\}_{k=1}^K$ from the generative model and transforming them into the estimation of risk measures through \eqref{eq:mcestimation}. This process involves three distinct sources of error:

\begin{enumerate}
	\item \textit{Quantile Estimation Error:} The construction of $\hat{Y}(\bx)$ begins with using quantile regression to estimate a discrete set of conditional quantile functions $\hat{Q}_{\tau_j}(\bx)$ from an offline dataset of size $n$. These finite-sample quantile function estimates at grid points inherently deviate from the true quantile functions due to data variability and model assumptions.
	
	\item \textit{Interpolation Error:} To approximate $F^{-1}_Y(\tau\,|\,\bx)$ for all $\tau$, we rely on a set of $m$ quantile levels and piecewise linear interpolation. This discretization, governed by $m$, introduces interpolation error that diminishes as we increase the number of quantile levels.
	
	\item \textit{Monte Carlo Sampling Error:} After constructing $\hat{Y}(\bx)$, the final estimation of risk measures is computed by sampling from this approximate distribution. The finite number $K$ of Monte Carlo observations leads to sampling variability, which decreases as $K$ grows.
	
\end{enumerate}

Notice that the third source of error can be made arbitrarily small 
by choosing a sufficiently large $K$, which is computationally feasible since QRGMM enables extremely fast data generation. Therefore, we focus our theoretical analysis on the first two error sources: the quantile estimation error associated with the finite sample size $n$, and the interpolation error linked to the number of quantile levels $m$.

\subsection{Uniform Convergence of Distribution
}\label{sec:asymptotic_r1}

The asymptotic analysis is based on the uniform convergence of the estimated conditional distribution function $F_{\hat{Y}}(y\,|\,\bx)$, which serves as the key ingredient for analyzing all subsequent risk measures.

We begin by recalling the assumptions and results of \cite{hong2023learning} to obtain the uniform convergence of $F_{\hat{Y}}(y\,|\,\bx)$. 

\begin{assumption}\label{ass:cons2}	
	\hfill
	\begin{enumerate}[label=(2.\alph*), left=0pt]
		\item \label{ass:cons2a} $F_Y(\cdot\,|\,\bx)$ is absolutely continuous on a common support for all $\bx\in\mathcal X$.
		\item \label{ass:cons2b} For any given interval $[\tau_\ell,\tau_u]\subset(0,1)$ with $0<\tau_\ell<\tau_u<1$, the density functions $f_Y(\cdot\,|\,\bx)$ are uniformly bounded away from $0$ and $\infty$ in the sense that
		\begin{align*}
			0<\inf_{\tau\in[\tau_\ell,\tau_u]}\inf_{\bx\in\mathcal{X}} f_Y(F_Y^{-1}(\tau\,|\,\bx)\,|\,\bx) <\infty,\\
			0< \sup_{\tau\in[\tau_\ell,\tau_u]}\sup_{\bx\in\mathcal{X}} f_Y(F_Y^{-1}(\tau\,|\,\bx)\,|\,\bx)<\infty.
		\end{align*}
	\end{enumerate} 
\end{assumption}
\begin{assumption}\label{ass:cons3}
	\hfill
	\begin{enumerate}[label=(3.\alph*), left=0pt]
		\item \label{ass:cons3a}	 For any given interval $[\tau_\ell,\tau_u]\subset(0,1)$ with $0<\tau_\ell<\tau_u<1$, there exist positive definite matrices $D_0$ and $D_1(\tau)$ such that
		\begin{align*}
			&\lim_{n\to\infty}\frac{1}{n}\sum_{i=1}^n\bx_i\bx_i^\intercal=D_0,\\ &\lim_{n\to\infty}\frac{1}{n}\sum_{i=1}^nf_Y(F^{-1}_Y(\tau\,|\,\bx_i)\,|\,\bx_i)\bx_i\bx_i^\intercal=D_1(\tau)
		\end{align*}
		uniformly for $\tau \in[\tau_\ell,\tau_u].$ 
		\item \label{ass:cons3b} $\max_{1\le i\le n}\|\bx_i\|=o(\sqrt{n})$. 
		
	\end{enumerate} 
\end{assumption}

Assumption~\ref{ass:cons2} and \ref{ass:cons3} are commonly adopted in quantile regression analysis to ensure robust and reliable statistical inferences. These conditions facilitate the estimation of quantile regression coefficients and provide the theoretical underpinning for the asymptotic analysis.

With the above assumptions, the following uniform convergence of $F_{\hat{Y}}(y\,|\,\bx)$ has been established by \cite{hong2023learning}.

\begin{lemma}[Hong et.al, 2023]\label{thm:weakconvergence_Y}
	Under Assumptions~\ref{ass:cons1}--\ref{ass:cons3}, the following results hold:
	\begin{enumerate}[label=(1.\alph*), left=0pt]
		\item \label{thm:weakconvergence_a}	 For any $\bx\in \mathcal{X}$, $\hat{Y}(\bx)  \overset{d}{\to} Y(\bx)$ as $n,m\to\infty$, where ``$ \overset{d}{\to}$" denotes convergence in distribution.  
		\item \label{thm:weakconvergence_b}	 For any $\bx\in \mathcal{X}$,
		$
		\sup\limits_{y\in[0,\infty)}\left|F_{\hat{Y}}(y\,|\,\bx)-F_Y(y\,|\,\bx)\right| \to 0
		$
		as $n,m\to\infty$.
	\end{enumerate} 
\end{lemma}

The uniform convergence of $F_{\hat{Y}}(y\,|\,\bx)$ established in Lemma~\ref{thm:weakconvergence_Y} serves as the central tool in analyzing the asymptotic behaviors of estimators of risk measures, which will be discussed in the next subsection.

\subsection{Asymptotic Analysis for Estimators of Risk Measures}\label{sec:asymptotic_r3}

Building on the uniform convergence of the conditional distribution function established in Lemma~\ref{thm:weakconvergence_Y}, we now analyze the risk measure
$$
r_3(l)\;=\;\mathbb{E}\!\left[\,g_l(Y(\bx))\,\mathbb{I}\{Y(\bx)<a(l)\}\right],
\quad l\in[0,\bar l],
$$
and its QRGMM-based plug-in estimator
$$
\hat r_3(l)\;=\;\mathbb{E}\!\left[\,g_l(\hat Y(\bx))\,\mathbb{I}\{\hat Y(\bx)<a(l)\}\right],
$$
where we ignore Monte Carlo error from sampling $\hat Y(\bx)$. It is important to note that $r_3(l)$ is the most general form of risk measure that we consider. It includes $r_1(l)$ and $r_2(l)$ as special cases.

Two sources of difficulty arise in the analysis.
(i) \emph{Expectation of a transformation.} Even though $F_{\hat Y}(\cdot\,|\,\bx)$ converges uniformly to $F_Y(\cdot\,|\,\bx)$, weak convergence alone does not in general imply convergence of expectations for nonlinear transforms such as $g_l(y)$.
(ii) \emph{Discontinuity from the indicator.} The factor $\mathbb{I}\{y<a(l)\}$ renders the map $y\mapsto g_l(y)\mathbb{I}\{y<a(l)\}$ potentially discontinuous at $a(l)$. Consequently, standard asymptotic convergence theory for bounded continuous functions, such as the Portmanteau theorem \citep[see, e.g.,][p.~6]{van2000}, does not directly apply.

To study the asymptotic behavior of $\hat{r}_3(l)$,
we first impose the following mild regularity conditions on $g_l(\cdot)$ and $a(l)$:

\begin{assumption}\label{ass:cons4}	
	\hfill
	\begin{enumerate}[label=(4.\alph*), left=0pt]
		\item \label{ass:cons4a} \( a(l) \) is monotone on \( [0, \bar{l}] \), satisfies \( 0 \leq a(l) \leq a(\bar{l}) < \infty \), with \( a(l) \to 0 \) as \( l \to 0 \). 
		\item \label{ass:cons4b} 
		\( g_l(\cdot) \) is Lipschitz continuous with constant \( K \) within $y\in[0,a(\bar{l})]$ uniformly over $l\in[0,\bar{l}]$, and uniformly bounded by $g_{\bar{l}}(0)<\infty$.
	\end{enumerate} 
\end{assumption}

Assumption~\ref{ass:cons4a} requires \( a(l) \), a threshold function, to be monotonic in $l$ and vanish as $l\to0$, reflecting the intuition that smaller loans imply lower trigger levels.
Assumption~\ref{ass:cons4b} requires that the loss function \( g_l(\cdot) \) to be Lipschitz continuous with constant \( K \) on $[0,a(\bar{l})]$, which by Rademacher's theorem (see, e.g., \citealt[Theorem~3.2]{evans2018measure}) ensures that \( g_l(\cdot) \) is differentiable almost everywhere (a.e.) and that its derivative satisfies \(\left|g_l'(y)\right| \leq K\) a.e.. Moreover, \( g_l(\cdot) \) is uniformly bounded by \( M = g_{\bar{l}}(0) \), reflecting the maximum loss in the worst case when sales \( Y(\bx)=0 \) and loan \( l=\bar{l} \).
These conditions are in general intuitive and reasonable for SCF risk management. They ensure tractable analysis of \( \hat{r}_3(l) \) and lead to the following result.

\begin{theorem}\label{thm:asym_r3}
	Under Assumptions~\ref{ass:cons1}--\ref{ass:cons4}, for any given \( \mathbf{x} \), as \( n, m \to \infty \), we have
	\[
	\sup_{l \in [0, \bar{l}]} \left| \hat{r}_3(l) - r_3(l) \right| \to 0,
	\]
\end{theorem}

\begin{proof}
	For any \(\mathbf{x} \in \mathcal{X}\) and \(l \in [0, \bar{l}]\),
	\begin{align*}
		r_3(l) &= \mathbb{E} \left[ g_l(Y(\bx)) \cdot \mathbb{I}_{\{Y(\bx) < a(l)\}} \right], \\
		\hat{r}_3(l) &= \mathbb{E} \left[ g_l(\hat{Y}(\bx)) \cdot \mathbb{I}_{\{\hat{Y}(\bx) < a(l)\}} \right].
	\end{align*}
	By rewriting the expectations in integral form, we obtain
	\begin{align*}
		r_3(l) &= \int_0^{a(l)} g_l(y) \, dF_Y(y\,|\,\bx), \\
		\hat{r}_3(l) &= \int_0^{a(l)} g_l(y) \, dF_{\hat{Y}}(y\,|\,\bx).
	\end{align*}
	Hence,
	\[
	\hat{r}_3(l) - r_3(l)
	=  \int_0^{a(l)} g_l(y) \, d(F_{\hat{Y}} - F_Y)(y\,|\,\bx).
	\]

	Since \(g_l(\cdot)\) is Lipschitz continuous with constant \(K\) on $[0,a(\bar{l})]$, uniformly over \(l \in [0,\bar{l}]\), it follows that \(g_l(y)\) is differentiable a.e. on this domain, with \(\left|g_l'(y)\right| \leq K\) a.e.. Moreover, since \(g_l(\cdot)\) is Lipschitz, it is absolutely continuous, so the integration by parts formula can be applied under the Lebesgue integral.
	
	Applying integration by parts to $\hat{r}_3(l) - r_3(l) $ yields
	\begin{eqnarray*}
		\lefteqn{\int_0^{a(l)} g_l(y) \, d(F_{\hat{Y}} - F_Y)(y\,|\,\bx)} \\
		&=& \left.\left[ g_l(y) \left(F_{\hat{Y}}(y\,|\,\bx) - F_Y(y\,|\,\bx)\right) \right]\right|_0^{a(l)}  \\
		& &-  \int_0^{a(l)} \left(F_{\hat{Y}}(y\,|\,\bx) - F_Y(y\,|\,\bx)\right) g_l'(y) \, dy.
	\end{eqnarray*}
	Therefore,
	\begin{eqnarray*}
		\hat{r}_3(l) - r_3(l)	&=&  g_l(a(l)) \left(F_{\hat{Y}}(a(l)\,|\,\bx) - F_Y(a(l)\,|\,\bx)\right)\\
		& & -\  g_l(0) \left(F_{\hat{Y}}(0\,|\,\bx) - F_Y(0\,|\,\bx)\right) \\
		& & - \int_0^{a(l)} \left(F_{\hat{Y}}(y\,|\,\bx) - F_Y(y\,|\,\bx)\right) g_l'(y) \, dy.
	\end{eqnarray*}
	
	Let 
	\begin{equation}\label{eq:delta}\delta_{n,m} = \sup_{y \geq 0} \left|F_{\hat{Y}}(y\,|\,\bx) - F_Y(y\,|\,\bx)\right|,
	\end{equation} 
	and denote \( M = g_{\bar{l}}(0)\). 
	For the boundary terms, since \(\left|g_l(a(l))\right| \leq M\), \(\left|g_l(0)\right| \leq M\), and \(\left|F_{\hat{Y}}(y\,|\,\bx) - F_Y(y\,|\,\bx)\right| \leq \delta_{n,m}\), it follows that
	\begin{align*}
		\left|g_l(a(l))\right| \cdot \left|F_{\hat{Y}}(a(l)\,|\,\bx) - F_Y(a(l)\,|\,\bx)\right| \leq M \delta_{n,m}, \\
		\left|g_l(0)\right| \cdot \left|F_{\hat{Y}}(0\,|\,\bx) - F_Y(0\,|\,\bx)\right| \leq M \delta_{n,m}.
	\end{align*}
	For the integral term, using $\left|F_{\hat{Y}}(y\,|\,\bx) - F_Y(y\,|\,\bx)\right| \leq \delta_{n,m}$ and $\left|g_l'(y)\right| \leq K$ a.e., we obtain
	\begin{eqnarray*}
		\lefteqn{\int_0^{a(l)} \left|F_{\hat{Y}}(y\,|\,\bx) - F_Y(y\,|\,\bx)\right| \cdot \left|g_l'(y)\right| \, dy} \\
		&\leq&  \int_0^{a(l)} \delta_{n,m} K \, dy = K \delta_{n,m} a(l) \leq K \delta_{n,m} a(\bar{l}),
	\end{eqnarray*}
	as \(a(l) \leq a(\bar{l})\).
	Combining the bounds above gives
	\begin{eqnarray*}
		\left|\hat{r}_3(l) - r_3(l)\right| &\leq&  M \delta_{n,m} + M \delta_{n,m} + K \delta_{n,m} a(\bar{l})\\
		&=& \left(2M + K a(\bar{l})\right) \delta_{n,m}.
	\end{eqnarray*}
	Because the bound is independent of \(l\), it holds uniformly:
	\begin{equation}\label{eq:r3bound}
		\sup_{l \in [0, \bar{l}]} \left|\hat{r}_3(l) - r_3(l)\right| \leq (2M + K a(\bar{l})) \delta_{n,m}.
	\end{equation}
	By Lemma~\ref{thm:weakconvergence_Y}(b), we have \(\delta_{n,m} \to 0\) as \( n, m \to \infty \). Thus,
	\[
	\sup_{l \in [0, \bar{l}]} \left|\hat{r}_3(l) - r_3(l)\right| \to 0.
	\]
	This completes the proof.
\end{proof}

Theorem~\ref{thm:asym_r3} establishes the uniform consistency of the QRGMM estimator of $r_3(l)$. The convergence analysis is facilitated by the boundedness and smoothness of the integrand $g_l(\cdot)$, and the controlled behavior of the truncation threshold $a(l)$. These structural conditions are naturally satisfied in the SCF context. The uniform convergence of \( \hat{r}_3(l) \) and using different risk measures provides a comprehensive basis for reliable risk measures estimation across varying loan levels, supporting adaptive loan sizing.

\subsection{Practical Choice of $m$}\label{rate_m}

In the above subsections, we establish the uniform convergence of the estimated conditional distribution function and the estimated risk measures when using the QRGMM algorithm. For implementation, however, one must specify the number of quantile grid points $m$ relative to the sample size $n$. In what follows, we formalize the convergence rate on the main quantile region and translate it into a principled recommendation for $m$.

We work on the main quantile region $[\tau_\ell,\tau_u]\subset(0,1)$, where the conditional density $f_Y(\cdot\,|\,\bx)$ is bounded and bounded away from zero by Assumption~\ref{ass:cons2}	. Let $Q(\tau\,|\,\bx)=F_Y^{-1}(\tau\,|\,\bx)$ and $\hat Q(\tau\,|\,\bx)$ be the conditional quantile function and its QRGMM estimator. \citet{hong2023learning} proves the following convergence rate for the quantile curve on the main quantile region.
\begin{lemma}[Hong et.al, 2023]\label{lemma:quantile_middle}
	Under Assumptions~\ref{ass:cons1}--\ref{ass:cons3}, for any given \( \mathbf{x} \), as \( n, m \to \infty \), we have
	$$
	\max_{\tau_j\in[\tau_\ell,\tau_u]}\bigl|\hat Q(\tau_j\,|\,\bx)-Q(\tau_j\,|\,\bx)\bigr| \;=\;O_{\mathbb P}\!\left(\frac{1}{\sqrt{n}}\right).
	$$ 
	Moreover,
	$$
	\sup_{\tau\in[\tau_\ell+1/m,\tau_u-1/m]}
	\bigl|\hat Q(\tau\,|\,\bx)-Q(\tau\,|\,\bx)\bigr|
	\;=\;O_{\mathbb P}\!\left(\frac{1}{\sqrt{n}}\right)\;+\;O\!\left(\frac{1}{m}\right).
	$$
\end{lemma}
Here, the notation $O_{\mathbb P}(\cdot)$ is defined in the following way: for a sequence $Z_n$, $Z_n = O_{\mathbb P}(h_n)$ means $Z_n/h_n$ is bounded in probability. The $O_{\mathbb P}(n^{-1/2})$ term reflects statistical error bound from estimating the quantile regressions on the grid, while the $O(m^{-1})$ term is the deterministic interpolation error bound between adjacent quantile levels.

We next convert Lemma~\ref{lemma:quantile_middle} into a convergence rate for the estimated conditional distribution function on the corresponding main range of $y$. Define
$\mathcal{Y}_{m,n}\left(\bx\right)=\bigl[\hat Q(\tau_\ell+1/m\,|\,\bx),\,\hat Q(\tau_u-1/m\,|\,\bx)\bigr].$

\begin{proposition}\label{prop:cdf_middle}
	Under Assumptions~\ref{ass:cons1}--\ref{ass:cons3}, for any given \( \mathbf{x} \), as \( n, m \to \infty \), we have
	$$
	\sup_{y\in\mathcal{Y}_{m,n}(\bx)}\bigl| F_{\hat{Y}}(y\,|\,\bx)-F_Y(y\,|\,\bx)\bigr|
	\;=\;O_{\mathbb P}\!\left(\frac{1}{\sqrt{n}}\right)\;+\;O\!\left(\frac{1}{m}\right).
	$$
\end{proposition}

The proof of Proposition~\ref{prop:cdf_middle} is provided in the Online Appendix~A. The convergence rate of Proposition~\ref{prop:cdf_middle} can also transfer to the default probability estimation $\hat{r}_1(l)$. Recall $r_1(l)=\Pr\{L(l)>0\}=\Pr\{Y(\bx)<l/r\}=F_Y(l/r\,|\,\bx)$; for loan levels $l$ such that $l/r\in\mathcal{Y}_{m,n}(\bx)$, Proposition~\ref{prop:cdf_middle} yields
$$
\sup_{l:\,l/r\in\mathcal{Y}_{m,n}(\bx)}\bigl|\hat r_1(l)-r_1(l)\bigr|
\;=\;O_{\mathbb P}\!\left(\frac{1}{\sqrt{n}}\right)\;+\;O\!\left(\frac{1}{m}\right).
$$

This rate highlights two sources of error: the estimation error $O_{\mathbb P}(n^{-1/2})$ and the interpolation error $O(m^{-1})$. To prevent the interpolation term from dominating while keeping computation reasonable, Proposition~\ref{prop:cdf_middle} suggests the rule of thumb $m =O(\sqrt{n})$.

For more general risk measures, Proposition~\ref{prop:cdf_middle} does not apply as directly as it does for $\hat{r}_1(l)$. Even so, the inequality \eqref{eq:r3bound} together with \eqref{eq:delta} and the proof of Theorem~\ref{thm:asym_r3} indicate that the main part of the error of these risk measures is influenced by the same convergence rate $O_{\mathbb P}(n^{-1/2})+O(m^{-1})$. 

Through a practical example, we test the sensitivity of choosing $m =\sqrt{n}$ in the Online Appendix~B. The results show that the choice is robust and performs reasonably well.

\begin{remark}
	For the convergence rate analysis, we only work on the main quantile region for both theoretical and practical reasons. Theoretically, rate analysis near the extreme quantile ($\tau \to 0,1$) is substantially more involved: it requires additional stronger tail assumptions and specialized extreme value theory. In particular, when the conditional density goes to 0 near the boundaries $\tau \to 0,1$, standard empirical process arguments no longer apply, and current literature provides limited theoretical tools to establish uniform convergence rates in these regions \citep{chernozhukov2005extremal,hong2023learning}.
	Practically, quantile estimates are more stable in the main region, which covers the bulk of the data, whereas reliable estimation of extreme quantiles would require a far larger sample size than what is typically available in SCF applications. Moreover, in our SCF setting, we do not need to consider extreme quantiles of the sales distribution $Y(\bx)$: when $\tau\to 0$, sales are nearly zero and no loan would be issued; while when $\tau\to 1$, sales reach their maximum potential but we only care about whether they are sufficient to cover the loan $l$, not about the extreme upper tail itself. In other words, since the potential loss is naturally bounded above by $\bar{l}$, and negative losses are not of primary concern in credit risk evaluation, restricting attention to the main quantile region can be regarded as a reasonable simplification.
\end{remark}

\section{Extension of QRGMM for Practical Applicability in SCF Risk Management}
\label{sec:extension_qrgmm}

In the previous section, we focused on the QRGMM under relatively simple model assumption, linear quantile regression model. While this assumption facilitated theoretical derivations and provided valuable insights into convergence properties, they may be restrictive for real-world SCF scenarios. In practice, the relationships governing sales, especially in cross-border e-commerce, are often complex, involving intricate combinatorial interactions among products, stores, and other contextual factors. The linear model may struggle to capture such combinatorial effects, potentially undermining the accuracy and effectiveness of distributional predictions required for robust SCF risk management.

To overcome this limitation, this chapter integrates advanced techniques from recommendation systems, specifically Deep Factorization Machines \citep[DeepFM,][]{guo2017deepfm}, with the QRGMM framework. This integration allows us to model non-linear, intricate feature interactions and improve the realism and accuracy of generated sales distributions. We term this extended model as QRGMM$^\dagger$, highlighting its advanced capability to handle complex combinatorial features and non-linear relationships, essential for robust SCF applications.

\subsection{Overview of Related Techniques}
\label{sec:related_techniques}

This subsection introduces the key building blocks for our extended model. We first discuss Matrix Factorization \citep[MF,][]{lee2000algorithms} and Factorization Machines \citep[FM,][]{rendle2010factorization}. They are fundamental techniques in recommendation systems that enable efficient modeling of latent and combinatorial features. We then present the Deep Factorization Machines technique \citep[DeepFM,][]{guo2017deepfm}, which combines FM with deep learning techniques to capture second-order and higher-order feature interactions. Lastly, the Quantile Regression Neural Network \citep[QRNN,][]{cannon2011quantile} offers an effective way to extend quantile regression from linear models to neural networks, serving as a natural bridge that seamlessly integrates DeepFM with the QRGMM framework.

\subsubsection{Matrix Factorization (MF)}
\begin{figure*}[!ht]
	\centering
	\includegraphics[width=0.45\linewidth]{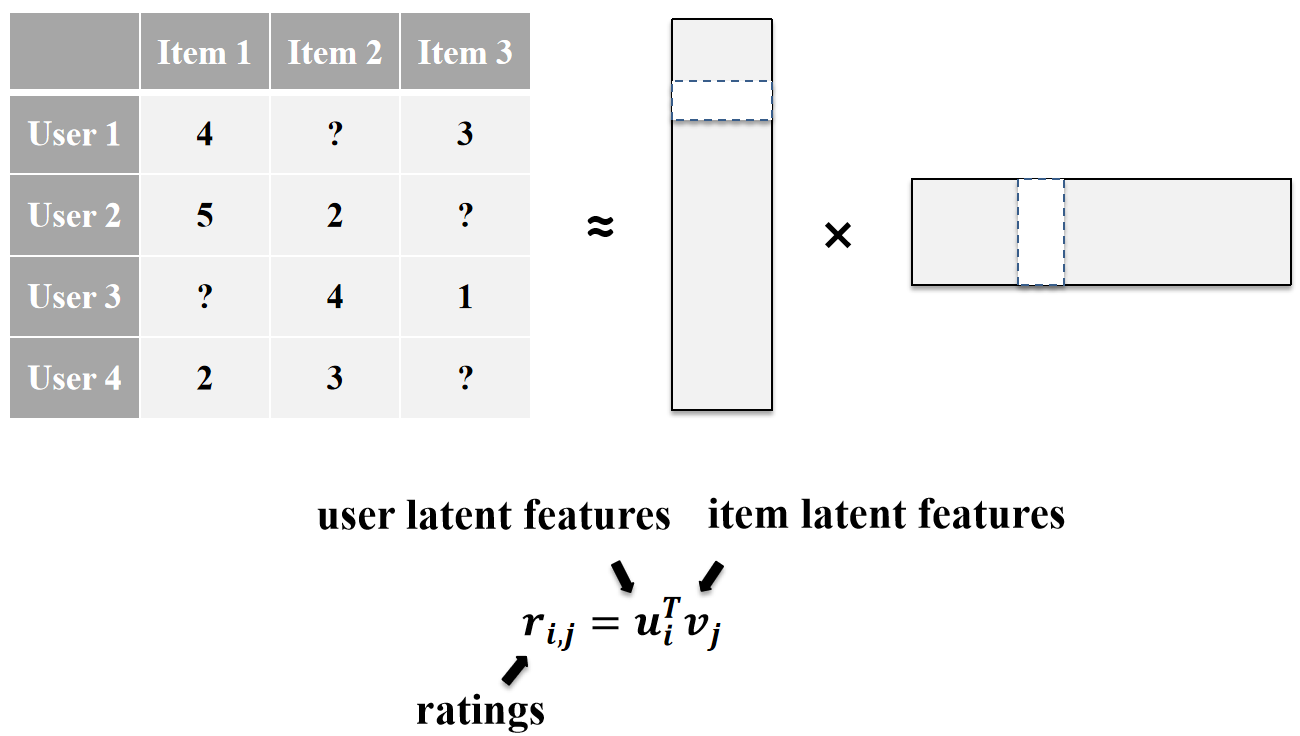}
	\hfill
	\includegraphics[width=0.45\linewidth]{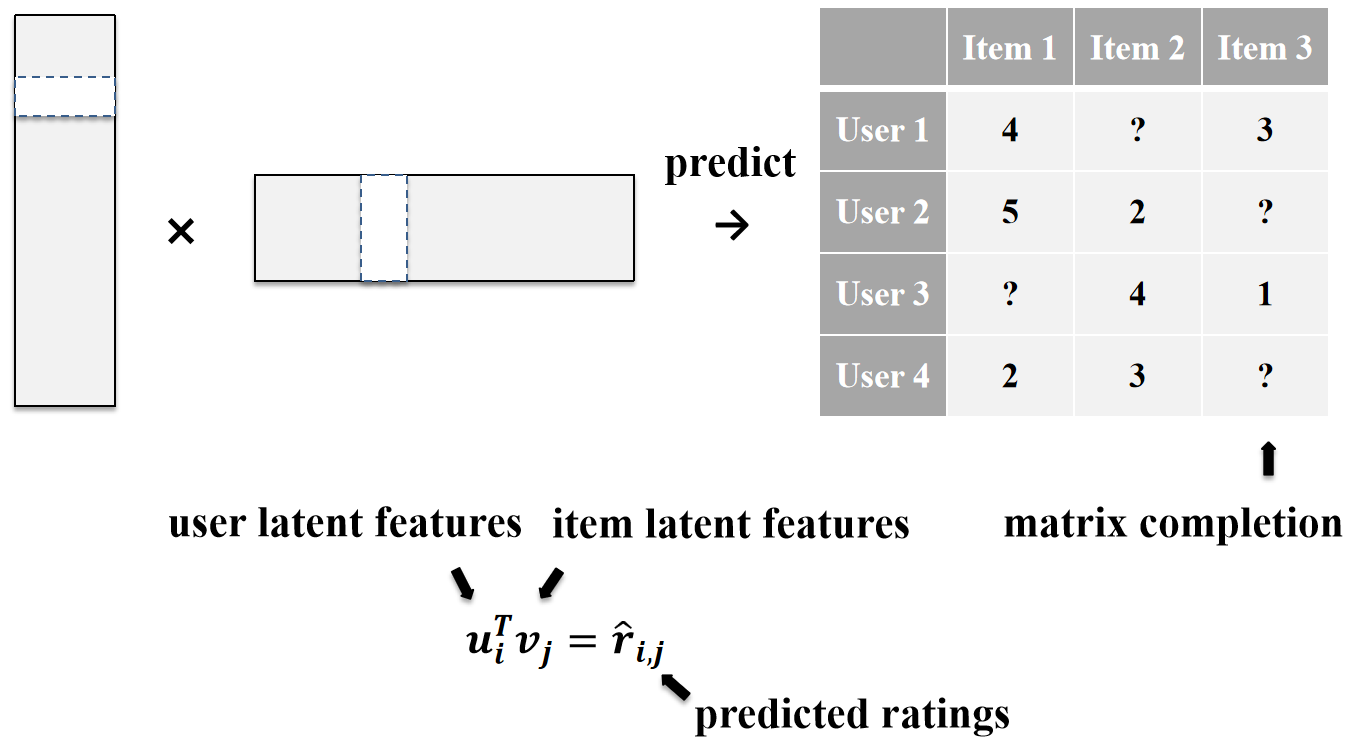}
	\caption{Illustration of the Matrix Factorization process.}
	\label{fig:mf}
\end{figure*}

Matrix Factorization is a foundational technique in recommendation systems and collaborative filtering. It decomposes a large, sparse interaction matrix (e.g., user-item ratings) into the product of two lower-rank latent factor matrices: $R \approx U^\top V,$
where \( R \) is the observed rating matrix, \( U \) and \( V \) are matrices of user and item latent factors, respectively. The model is trained by minimizing reconstruction error while incorporating regularization to prevent overfitting:
\begin{equation*}
	\min_{U,V}\sum_{(i,j)\in \mathcal{K}} \left(r_{ij}-U_i^\top V_j\right)^2+\lambda\left(\|U\|^2+\|V\|^2\right),
\end{equation*}
where \( \mathcal{K} \) is the set of observed entries and \(\lambda\) is a regularization parameter. The learned model $U^\top V$ can then be used to predict unknown ratings, effectively filling in the missing entries in the matrix (see Figure~\ref{fig:mf}).

In the SCF context, we can analogize users and items to entities such as stores and products, and ratings to observed sales. By identifying latent factors underlying these observed sales, MF provides insight into complex, hidden relationships that simpler linear models may fail to capture.

\subsubsection{Factorization Machines (FM)}
While MF focuses on a user-item dyad, the Factorization Machines technique extends this idea to a broader class of features and their interactions. FM efficiently captures pairwise feature interactions by introducing latent vectors for each feature:
\begin{equation*}
	\hat{y}(\mathbf{x}) = w_0+\sum_{i=1}^p w_i x_i + \sum_{i=1}^p \sum_{j=i+1}^p \langle \mathbf{v}_i, \mathbf{v}_j \rangle x_i x_j,
\end{equation*}
where \(\mathbf{v}_i \in \mathbb{R}^k\) is the latent vector for the \(i\)-th feature and \(\langle \cdot,\cdot \rangle\) denotes the inner product. 
One key insight of FM is that it models the coefficients of interactions between different features as the inner product of their latent vectors, effectively decomposing their interactions through these coefficients. 

\begin{figure*}[!ht]
	\centering
	\includegraphics[width=0.8\linewidth]{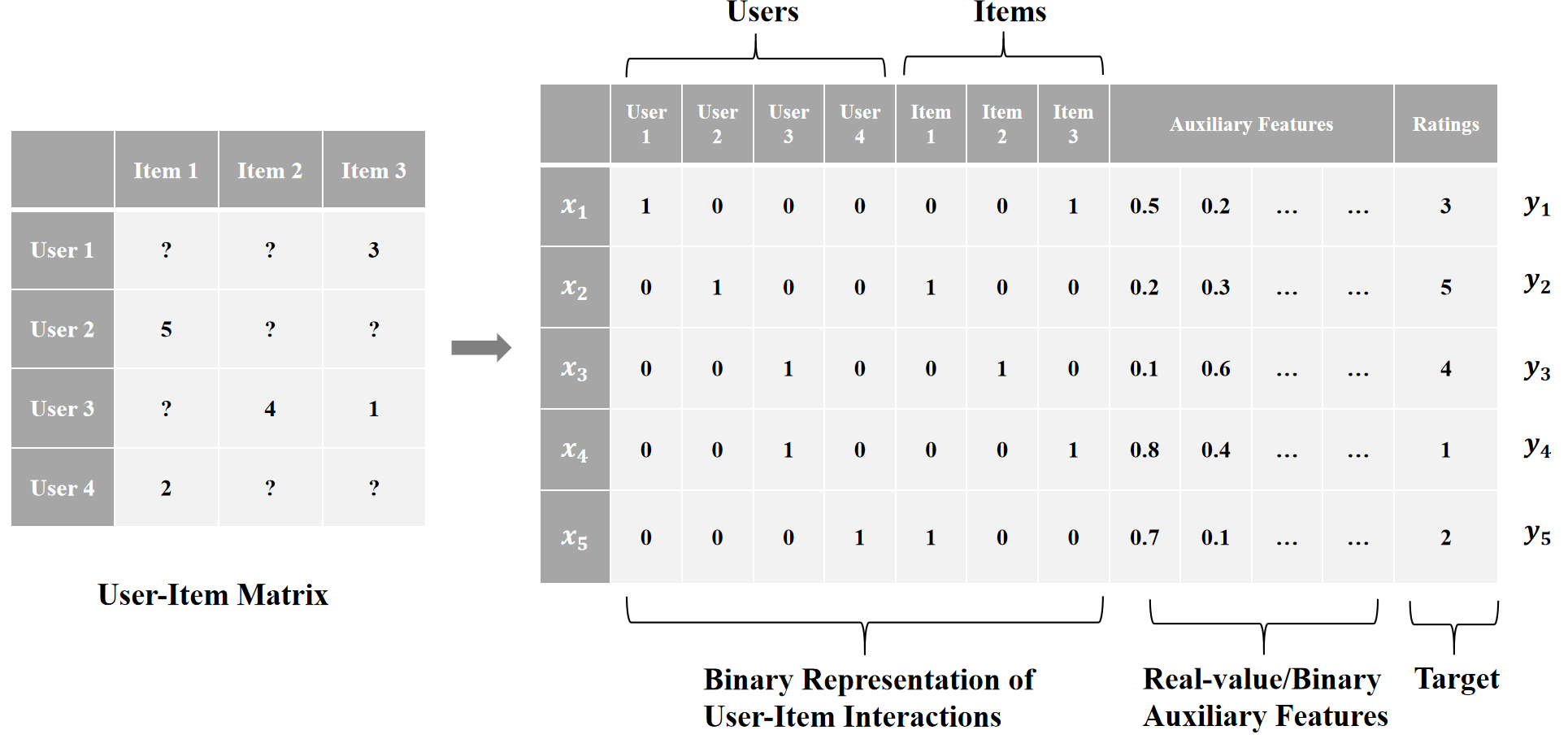}
	\caption{From Matrix Factorization to Factorization Machines.}
	\label{fig:FM}
\end{figure*}

As shown in Figure~\ref{fig:FM}, compared to MF, FM provides a more flexible structure that can incorporate a broader range of features, making it suitable for capturing complex relationships in diverse datasets. Additionally, compared to traditional polynomial kernel-based methods like Support Vector Machines (SVM), FM demonstrates superior generalization capabilities in sparse settings. The dense parametrization of SVMs relies on direct observations of feature interactions, which are often unavailable in sparse data, whereas FM can effectively estimate parameters even under such conditions (for more details, see \cite{rendle2010factorization}).

\subsubsection{Deep Factorization Machines (DeepFM)}
\begin{figure*}[!ht]
	\centering
	\includegraphics[width=0.8\linewidth]{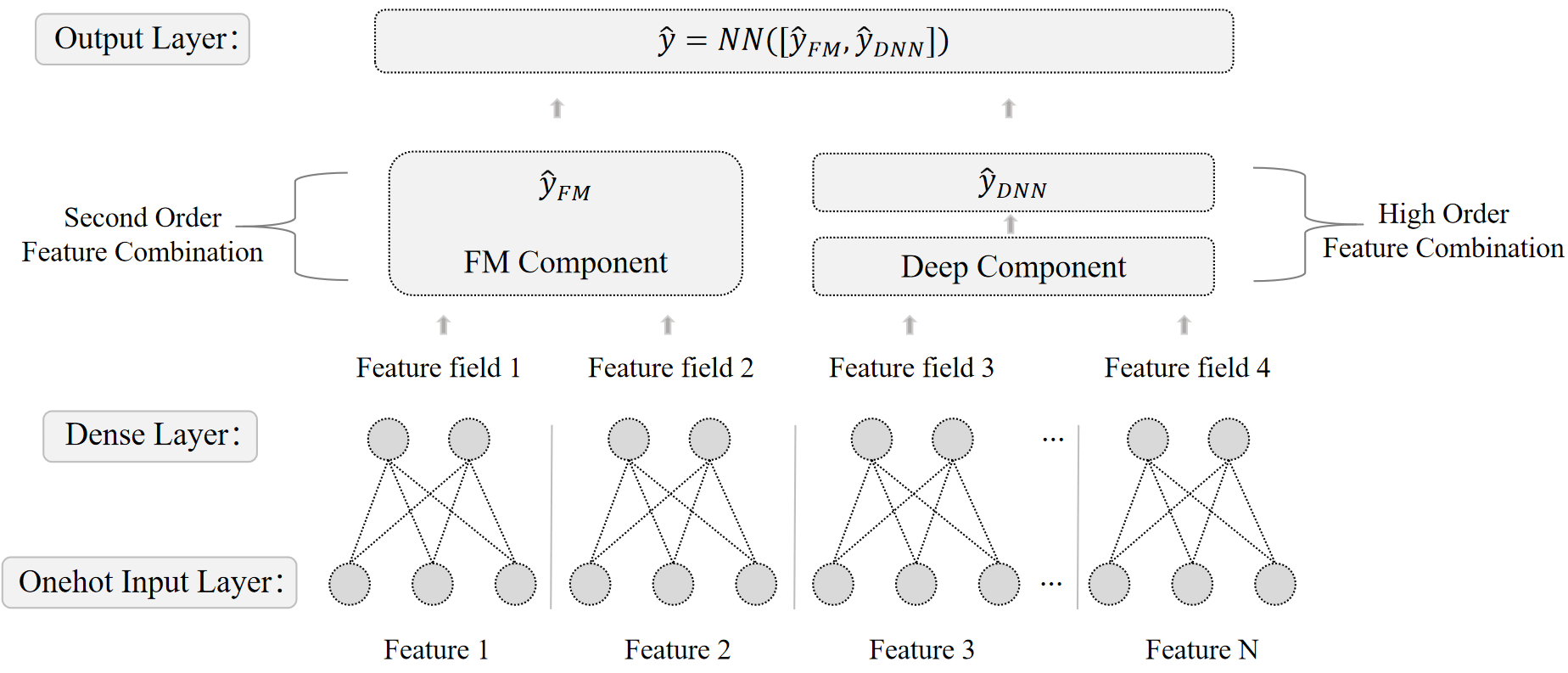}
	\caption{Architecture of Deep Factorization Machines.}
	\label{fig:deepfm}
\end{figure*}

The Deep Factorization Machines technique combines the strengths of FM with deep neural networks to model both low-order and high-order feature interactions. 
The FM component efficiently learns second-order feature interactions, and the deep neural network captures higher-order interactions, together reducing the need for extensive manual feature engineering.
The structure of DeepFM is shown in Figure~\ref{fig:deepfm}.

By integrating DeepFM, we can capture intricate relationships among various features of the covariates in SCF data (e.g., product attributes, store characteristics, advertisement factors), moving beyond linear approximations and paving the way for modeling complex sales distributions.

\subsubsection{Quantile Regression Neural Network (QRNN)}

The methods discussed earlier, including MF, FM, and DeepFM, are primarily designed for mean-based prediction tasks. 
However, our goal is to predict the entire distribution of sales, rather than just the mean, which is crucial for robust SCF risk management. Given that DeepFM is essentially a neural network architecture, it becomes necessary to extend the linear quantile regression model of QRGMM into a neural-network-based quantile regression model. QRNN provides an ideal tool for achieving this extension.

QRNN enhances conventional quantile regression by leveraging neural networks to approximate quantile functions directly. Instead of constraining the quantile function to a linear form, QRNN employs a flexible neural network architecture, often composed of multiple dense layers with nonlinear activation functions, to model the potentially intricate relationship between covariates and various quantiles of the response variable.
Unlike the loss functions used in MF, FM, or DeepFM, which typically minimize mean squared error, QRNN adopts the same pinball loss as the linear quantile regression model introduced earlier in this paper. In addition, QRNN provides an efficient way to fit multiple quantiles at different quantile levels \(\tau_j\), the total loss is typically the sum over all quantiles and observations:
\begin{equation*}
	\min_{\text{model parameters}} \sum_{j=1}^{m-1} \sum_{i=1}^{n} \rho_{\tau_j}(y_i - \hat{y}_{i,j}),
\end{equation*}
where \(\hat{y}_{i,j}\) denotes the predicted \(\tau_j\)-quantile for observation \(i\), obtained as the output from the neural network model. The model can be efficiently solved by SGD methods.

By integrating this quantile-specific loss with a neural network structure, QRNN can learn multiple quantiles simultaneously, capturing distinct aspects of the underlying conditional distribution. Neural network components such as nonlinear activations, dropout layers, and proper regularization enable the QRNN to represent complex, nonlinear patterns and maintain stability during training. 
In essence, QRNN extends quantile regression capabilities beyond linear assumptions, allowing for highly non-linear relationships and more realistic modeling of distributions.

\subsection{Integration of DeepFM and QRGMM: QRGMM$^\dagger$}
\label{sec:integration_dfm_qrnn}

\begin{figure*}[!ht]
	\centering
	\includegraphics[width=0.95\linewidth]{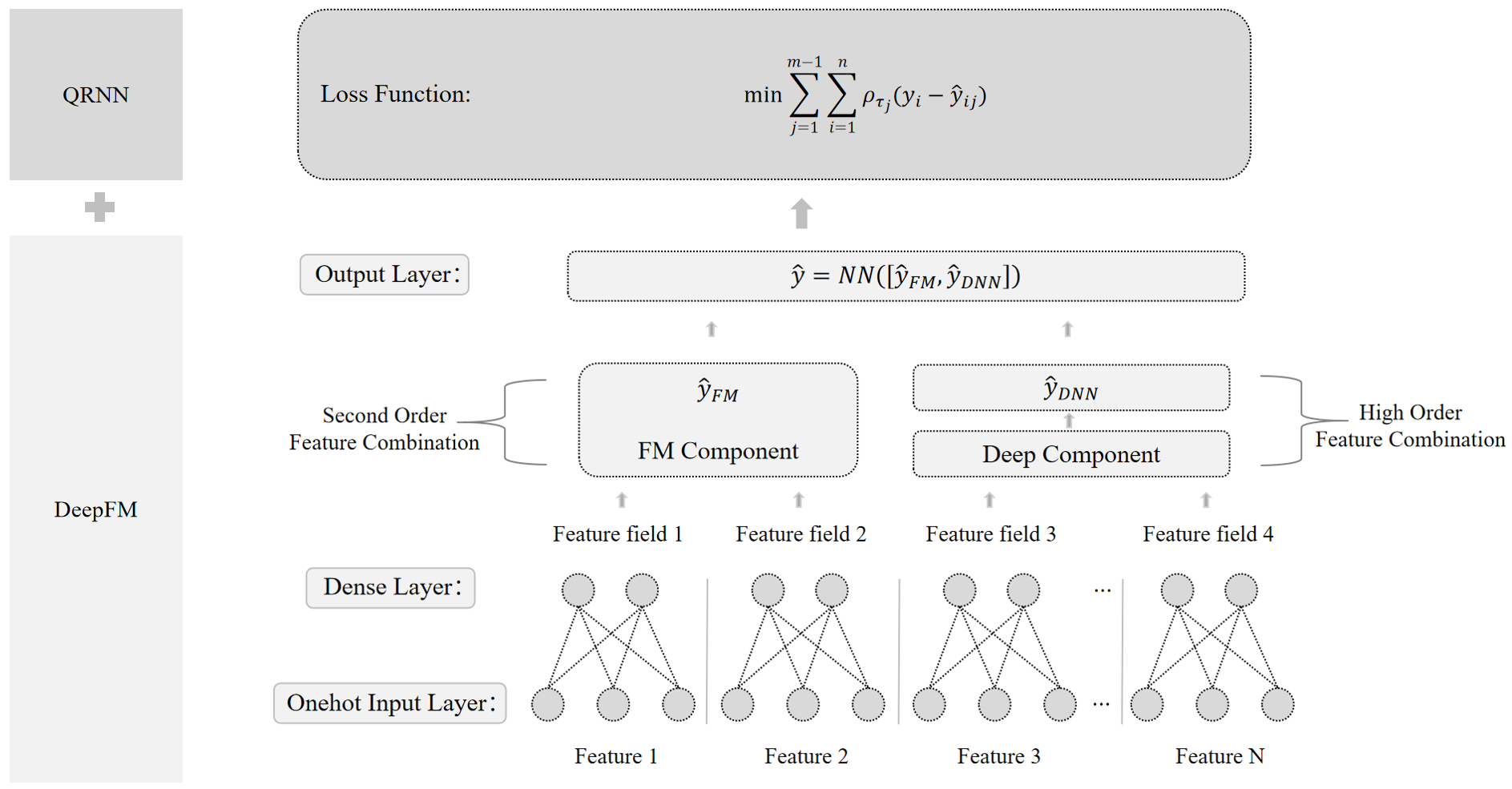}
	\caption{QRGMM$^\dagger$: integration of DeepFM and QRNN.}
	\label{fig:deepfmqrgmm}
\end{figure*}

Building on the techniques introduced above, we now present the integration of DeepFM and QRNN into the QRGMM framework, resulting in the QRGMM$^\dagger$ model. The idea is to leverage DeepFM's ability to capture intricate feature interactions and QRNN's flexibility in modeling quantiles to produce a generative model that can robustly simulate entire sales distributions under complex, realistic conditions. Its structure is shown in Figure~\ref{fig:deepfmqrgmm}. 

By using DeepFM as the neural backbone of the quantile regression network, we ensure that the learned quantile functions are sensitive to intricate second-order and higher-order feature interactions. Once the quantile functions are estimated, QRGMM's inverse transform approach is applied to generate observations. Specifically, for a given \(\bx\), we draw \(u_k \sim \text{Uniform}(0,1)\), and interpolate between the estimated quantiles to generate synthetic observations \(\hat{Y}_k(\bx)\) that approximate the conditional distribution of \(Y(\bx)\). The overall procedure follows Algorithm~\ref{alg:gmm}, except that the quantile regression model is replaced with the QRGMM$^\dagger$ model based on DeepFM.

By integrating DeepFM and QRNN within the QRGMM framework, the resulting QRGMM$^\dagger$ model more accurately captures the complex, non-linear interactions among features that characterize SCF sales patterns, bridging the gap between linear QRGMM and real-world complexities and leading to more realistic sales distributions.  
This enhanced generative capability supports improved estimation of risk measures, more informed lending decisions, and stronger risk management strategies.

\begin{remark}
	While our framework is motivated by credit risk management in 3PL-led CBEC SCF, its methodological core, constructing a reliable generative model of the underlying distribution and estimating risk measures via Monte Carlo simulation, naturally extends to other financial contexts. For example, the proposed approach can accommodate classical risk measures such as value-at-risk (VaR) and conditional value-at-risk (CVaR), which are widely used in insurance and financial risk management. In principle, once a reliable generative model is available, it can be used to simulate losses under different scenarios to estimate a wide spectrum of financial risk measures.
	
	What makes our setting distinctive, however, is the close linkage between sales distributions and collateral value in the CBEC supply chain finance. Here, the repayment of a loan depends directly on realized sales, making the accurate modeling of $Y(\bx)$ not only relevant but indispensable. In other SCF contexts, such as domestic trade credit, alternative forms of collateral (e.g., factories, equipments, or real estates) often mitigate the reliance on sales forecasts, reducing the need for such distributional modeling. By contrast, in cross-border e-commerce, the absence of tangible collateral elevates the importance of predictive analytics, which motivates our integration of QRGMM with DeepFM to better capture the intricate combinatorial patterns inherent in e-commerce data. Therefore, while the proposed framework is in principle extensible to other domains, its present formulation is particularly well aligned with the unique challenges of CBEC-SCF.
	
\end{remark}

\section{Numerical Experiments}\label{sec:numerical_experiments}

This section presents a comprehensive evaluation of our proposed QRGMM$^\dagger$ approach to assess its performance and practical value in SCF risk management. We begin with a synthetic scenario where the full data-generating process is known. In this controlled environment, we can thoroughly examine the model's ability to accurately estimate conditional quantiles, generate realistic distributions, and precisely estimate various risk measures. Following the synthetic analysis, we apply the method to two real-world datasets, thereby demonstrating its relevance and adaptability to practical SCF contexts.

\subsection{Experiments on Synthetic Data}\label{experiment_synthetic}

To create a challenging yet interpretable test environment, we construct a synthetic dataset of 10,000 observations using an FM-location-scale shift model:
\begin{align*}
	Y(\bx)=&w_0+\sum_{i=1}^p w_i x_i+\sum_{i=1}^p \sum_{j=i+1}^p\left\langle\mathbf{v}_i, \mathbf{v}_j\right\rangle x_i x_j \nonumber \\ 
	& + \left(r_0+\sum_{i=1}^p r_i x_i+\sum_{i=1}^p \sum_{j=i+1}^p\left\langle\mathbf{z}_i, \mathbf{z}_j\right\rangle x_i x_j\right)u,  
\end{align*}
where $\log(u) \sim N(0,1)$. Moreover,
the covariate vector \(\bx=(x_1,\ldots,x_{410})\) represents a mix of categorical and continuous features to mimic the complexity commonly encountered in SCF data. The first 100 features represent the one-hot encoding of product categories, the subsequent 300 features correspond to the one-hot encoding of seller IDs, and the final 10 features are continuous variables that capture additional product or market factors. This high-dimensional, heterogeneous covariate structure provides a realistic testbed for distribution-oriented modeling.

We randomly partition the data into a training set of 8,000 observations and a test set of 2,000 observations. Throughout all experiments, we follow the guideline in Section~\ref{rate_m} and set $m=\sqrt{n}$, which yields $m=90$ in this case. In addition, a sensitivity analysis of QRGMM$^\dagger$ with respect to $m$ is presented in the Online Appendix~B, confirming that the choice $m=\sqrt{n}$ is reasonable in practice. 
After the model is fitted, we assess its performance in terms of quantile estimation accuracy, generative quality, and risk measures estimation through subsequent evaluations, and compare it with a popular generative model, the conditional wasserstein generative adversarial networks with gradient penalty \citep[CWGAN-GP,][hereafter referred to as CWGAN]{athey2021using}.

\subsubsection{Quantile Estimation and Distribution Generation}

\begin{figure}[ht]
	\centering
	\includegraphics[width=1\linewidth]{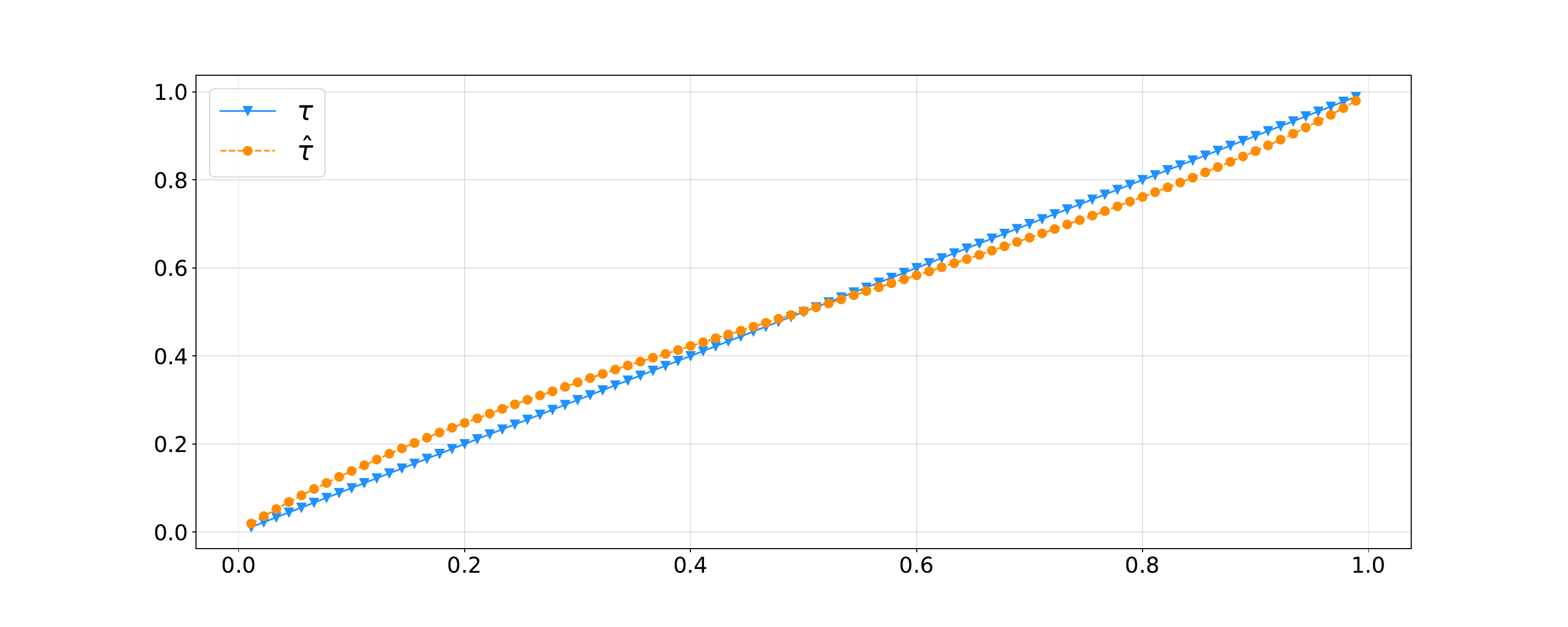}
	\caption{Alignment between nominal and empirical quantile levels on the synthetic data.}\label{fig:tauhat_syn}
\end{figure}

We first assess the accuracy of the model's quantile estimation. For each test observation $\bx_i$, we predict \(\tau_j\)-quantiles, denoted as $\hat{y}_{ij}$, and compare the empirical proportion of test responses that fall below these predicted quantiles, \(\hat{\tau}_j = \frac{1}{N_{\text{test}}}\sum_{i=1}^{N_{\text{test}}}\mathbb{I}\left(y_i \leq \hat{y}_{ij}\right)\), with the nominal level \(\tau_j\), for $j=1,2,\cdots,m-1$. If the model perfectly estimates the \(\tau_j\)-quantile, \(\hat{\tau}_j\) should be close to \(\tau_j\), for $j=1,2,\cdots,m-1$. We conduct the experiment over 100 replications and plot the average $\hat{\tau}_j$ across these replications in Figure~\ref{fig:tauhat_syn}. The result demonstrates a close alignment between $\hat{\tau}_j$ and $\tau_j$ for all $j=1,2,\cdots,m-1$, confirming that the model provides accurate quantile estimates.

Next, we evaluate the generative capabilities of the QRGMM$^\dagger$ model through both unconditional and conditional tests, comparing the distributions of generated data against the true test set. Our comparisons include not only the CWGAN baseline, but also the original linear version of QRGMM, in order to highlight the gains brought by integrating the DeepFM component into QRGMM$^\dagger$. Additionally, to better understand the individual contributions of the deep neural network component and the factorization machine component in QRGMM$^\dagger$, we also perform ablation studies with two simplified variants: QRGMM-D (with only the deep neural network component) and QRGMM-F (with only the factorization machine component). 

For CWGAN, we follow the hyperparameter choices recommended in \cite{athey2021using}, and conduct a grid search over the sensitive parameter \texttt{max\_epochs}. For QRGMM$^\dagger$, we choose $m=\sqrt{n}$, with the remaining hyperparameters handled as in ordinary neural network training. QRGMM, QRGMM-D, and QRGMM-F adopt the same choice of $m$, and all hyperparameters of QRGMM-D and QRGMM-F are kept identical to those of QRGMM$^\dagger$; the sole distinction lies in the output layer in Figure~\ref{fig:deepfmqrgmm}, where only a single component output is passed into the output layer instead of concatenating both components.

\paragraph*{Unconditional Test}

\begin{figure}[t]
	\centering
	\includegraphics[width=0.49\textwidth]{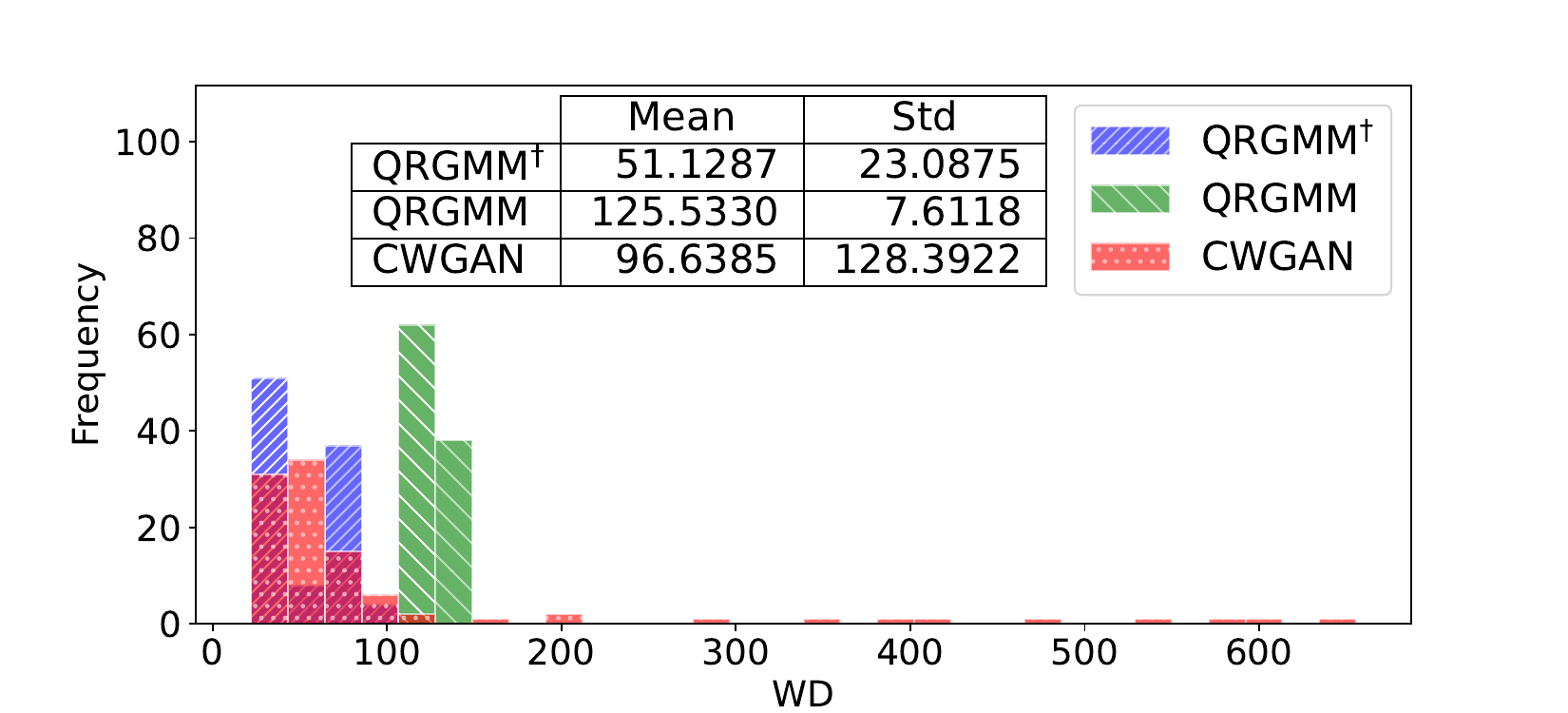} 
	\includegraphics[width=0.49\textwidth]{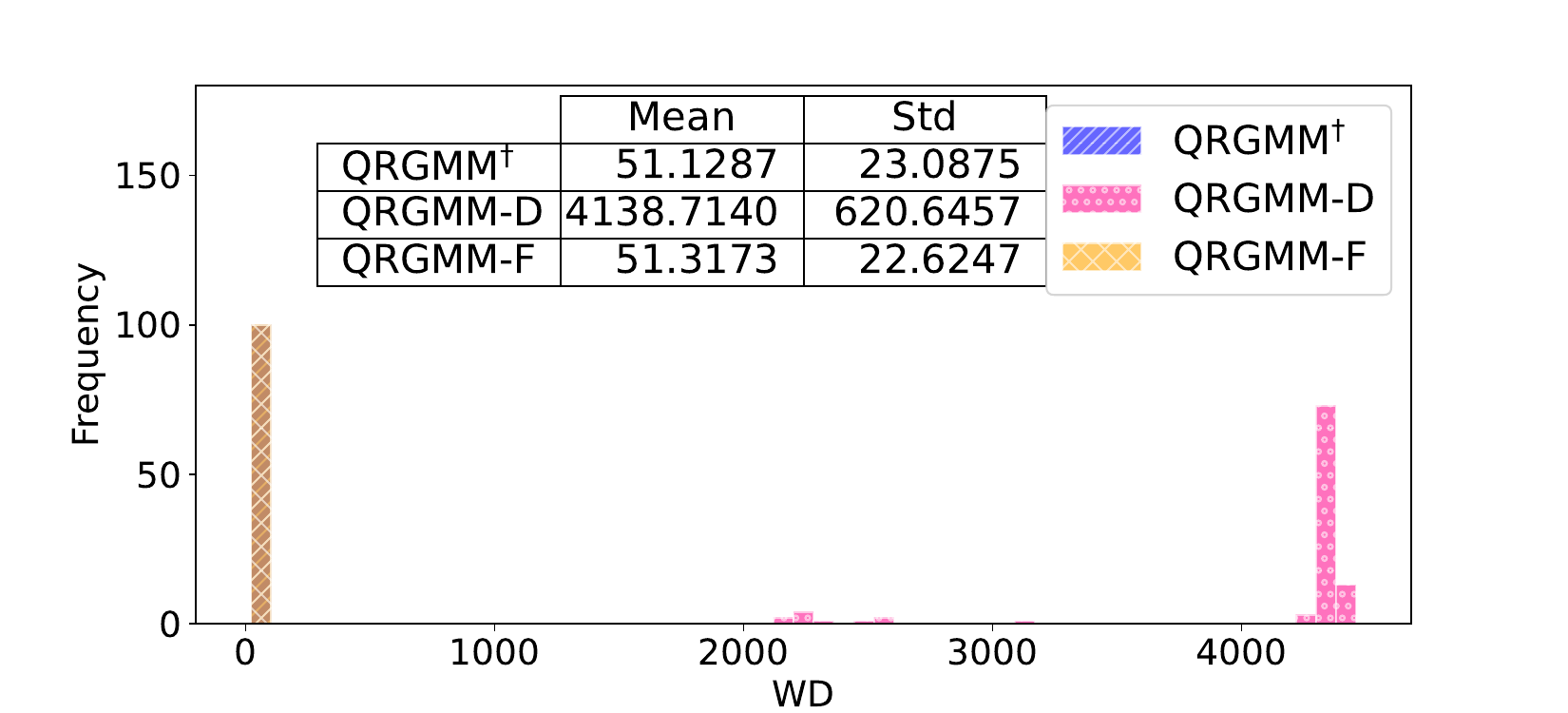} 
	\caption{WD for unconditional test.}\label{fig:wd_unconditional}
\end{figure}

\begin{figure}[t]
	\centering
	\includegraphics[width=0.49\textwidth]{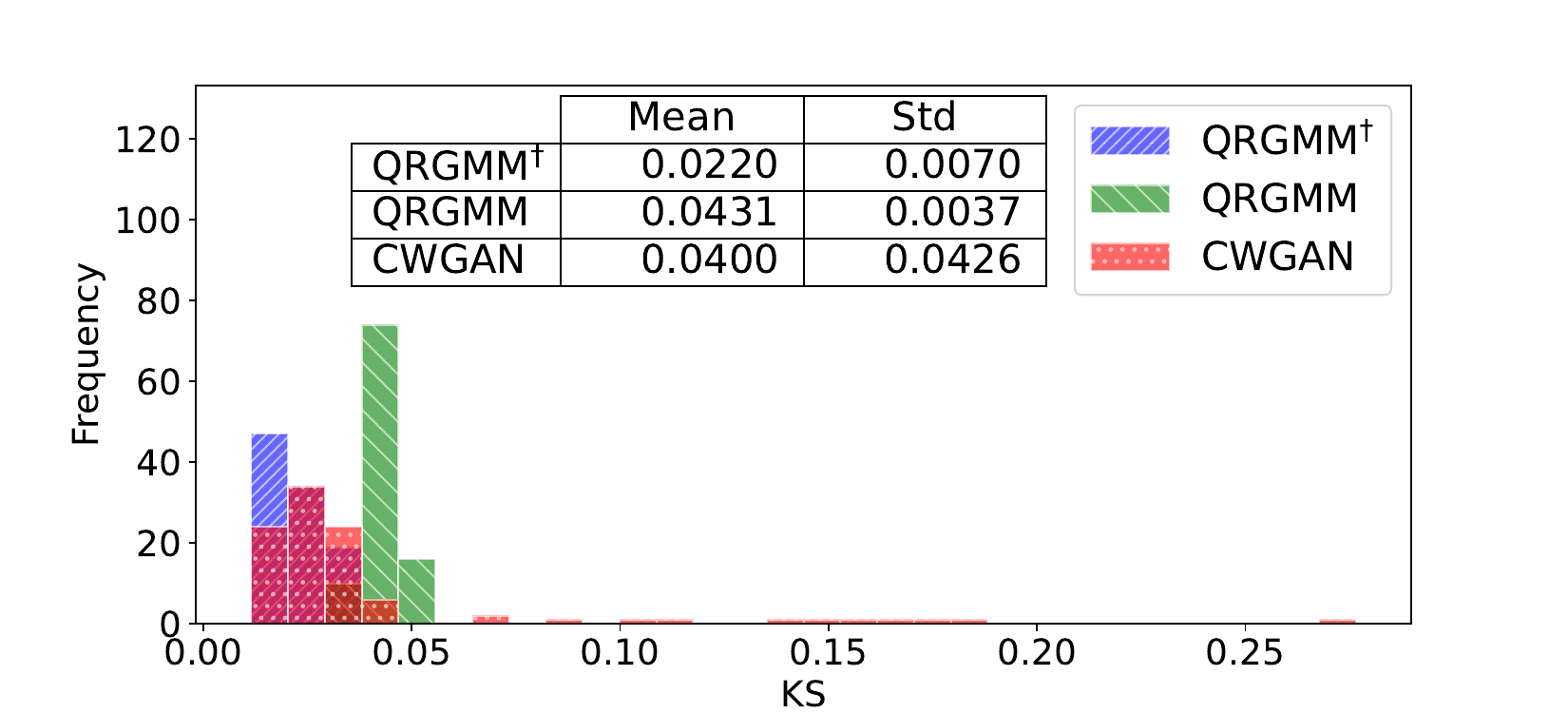}
	\includegraphics[width=0.49\textwidth]{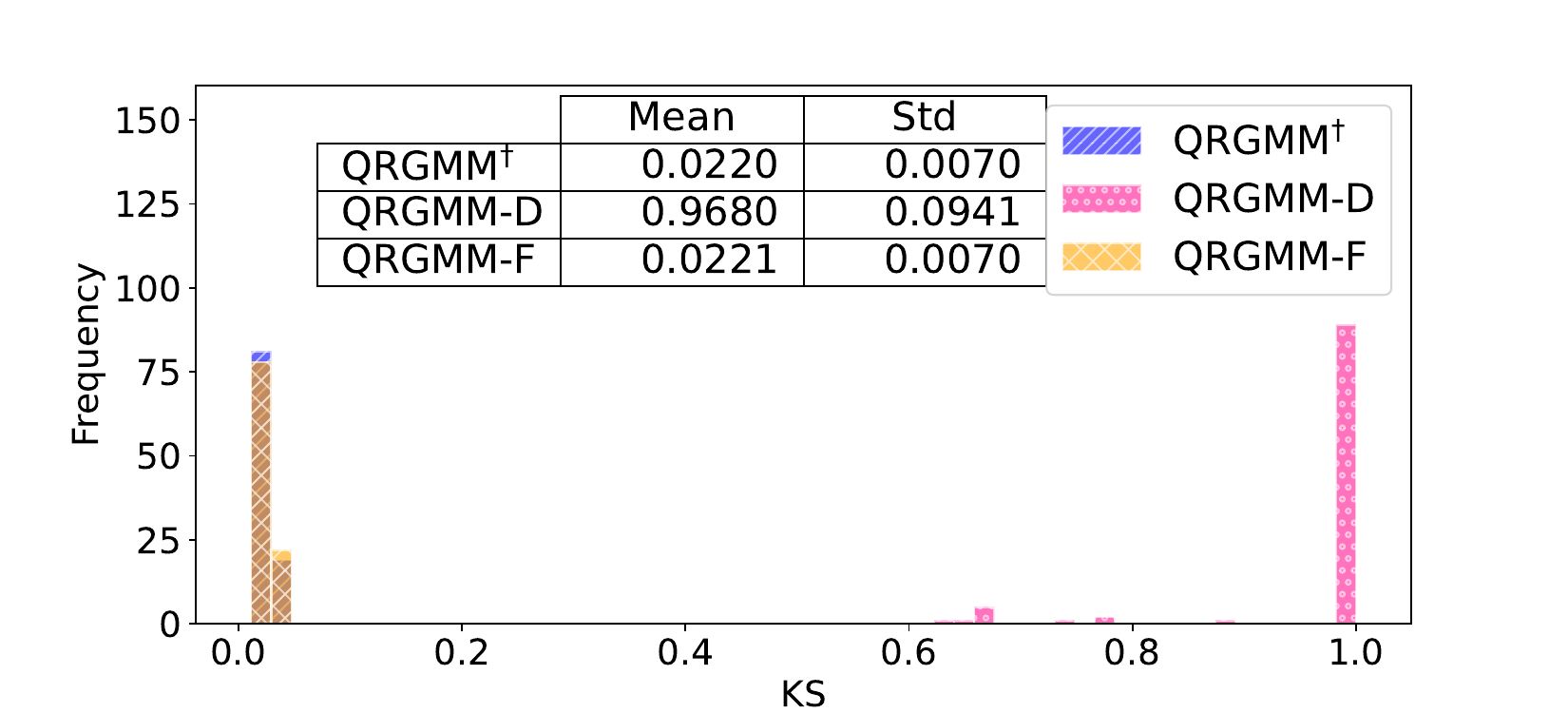}
	\caption{KS for unconditional test.}\label{fig:ks_unconditional}
\end{figure}

Using the trained models, we generate random observations \(\hat{y}^{\prime}_i\) for each test covariate vector \(\bx_i^{\prime}\), thus constructing generated test datasets. Then we can compare the overall distribution of the generated test observations \(\{\hat{y}^{\prime}_i : i = 1, \ldots, N_{\text{test}}\}\) with the true test observations \(\{y^{\prime}_i : i = 1, \ldots, N_{\text{test}}\}\). We conduct 100 experimental replications, and calculate the sample means and standard deviations for the generated datasets and the true test dataset in each run. The averages of these values across the 100 replications are reported in Table~\ref{table:sync}. To quantify distributional similarity, we also computed the Wasserstein distance (WD) and the Kolmogorov-Smirnov statistic (KS) between the generated datasets and the true test dataset for each replication. The histograms of these measures across replications are illustrated in Figures~\ref{fig:wd_unconditional} and \ref{fig:ks_unconditional}.

\paragraph*{Conditional Test}

\begin{figure}[t]
	\centering
	\includegraphics[width=0.49\textwidth]{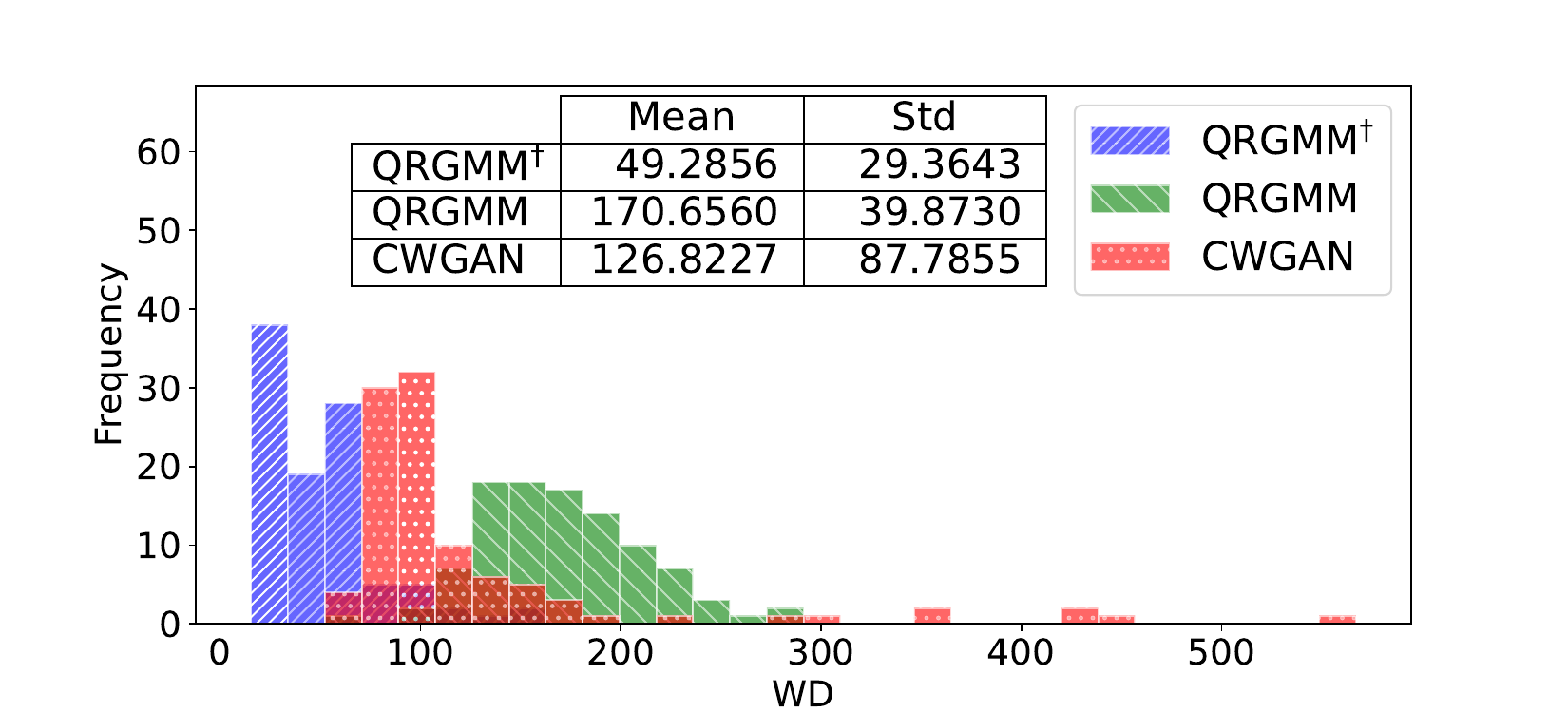} 
	\includegraphics[width=0.49\textwidth]{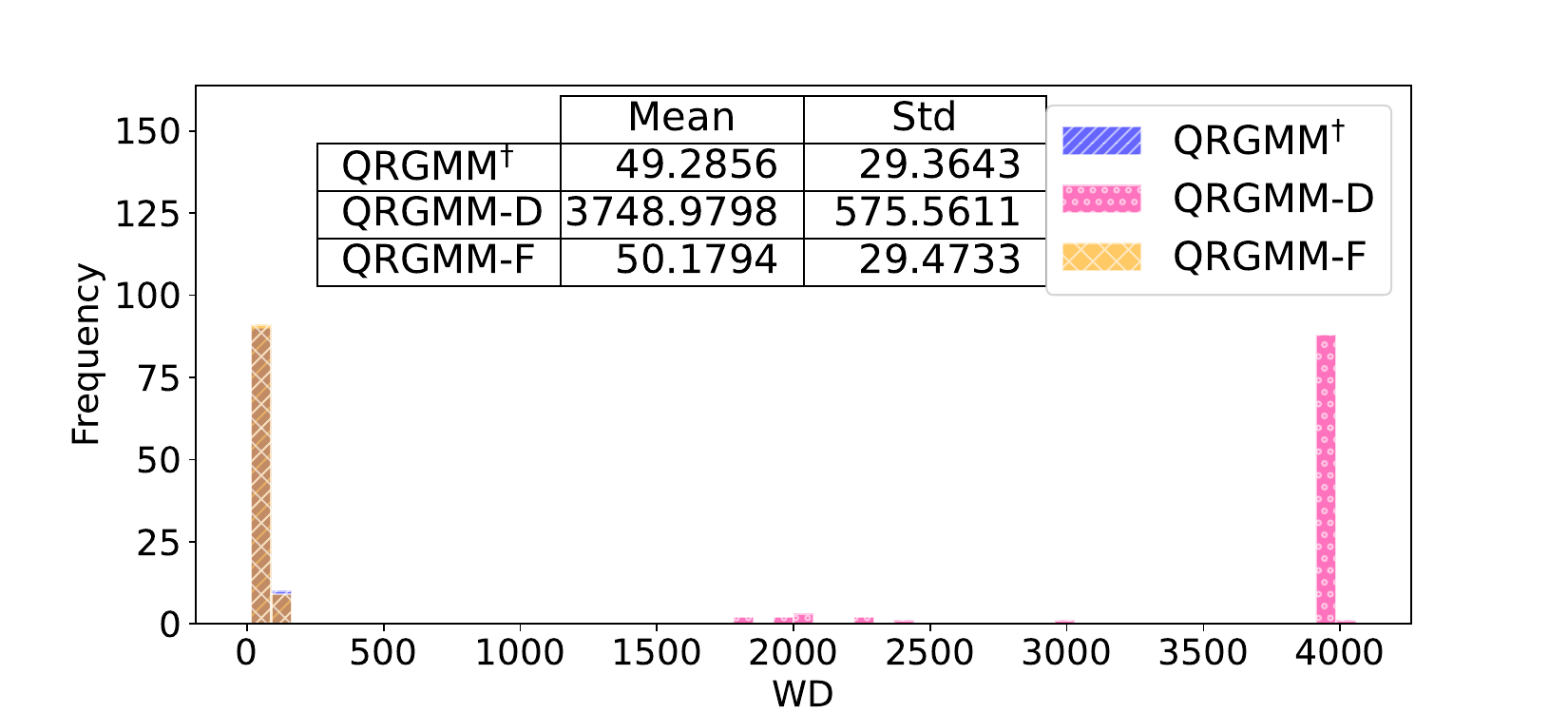}
	\caption{WD for conditional test.}\label{fig:wd_conditional}
\end{figure}

\begin{figure}[t]
	\centering
	\includegraphics[width=0.49\textwidth]{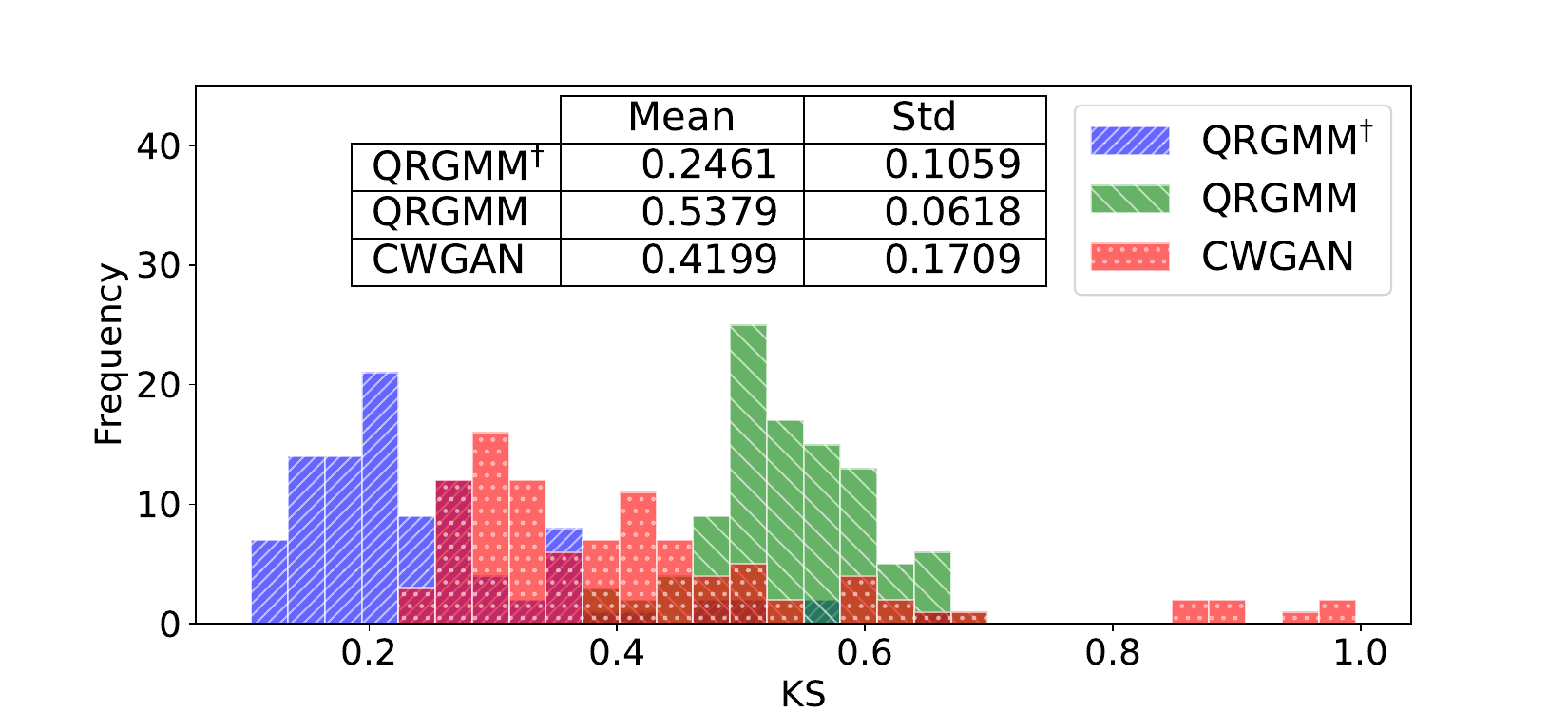}
	\includegraphics[width=0.49\textwidth]{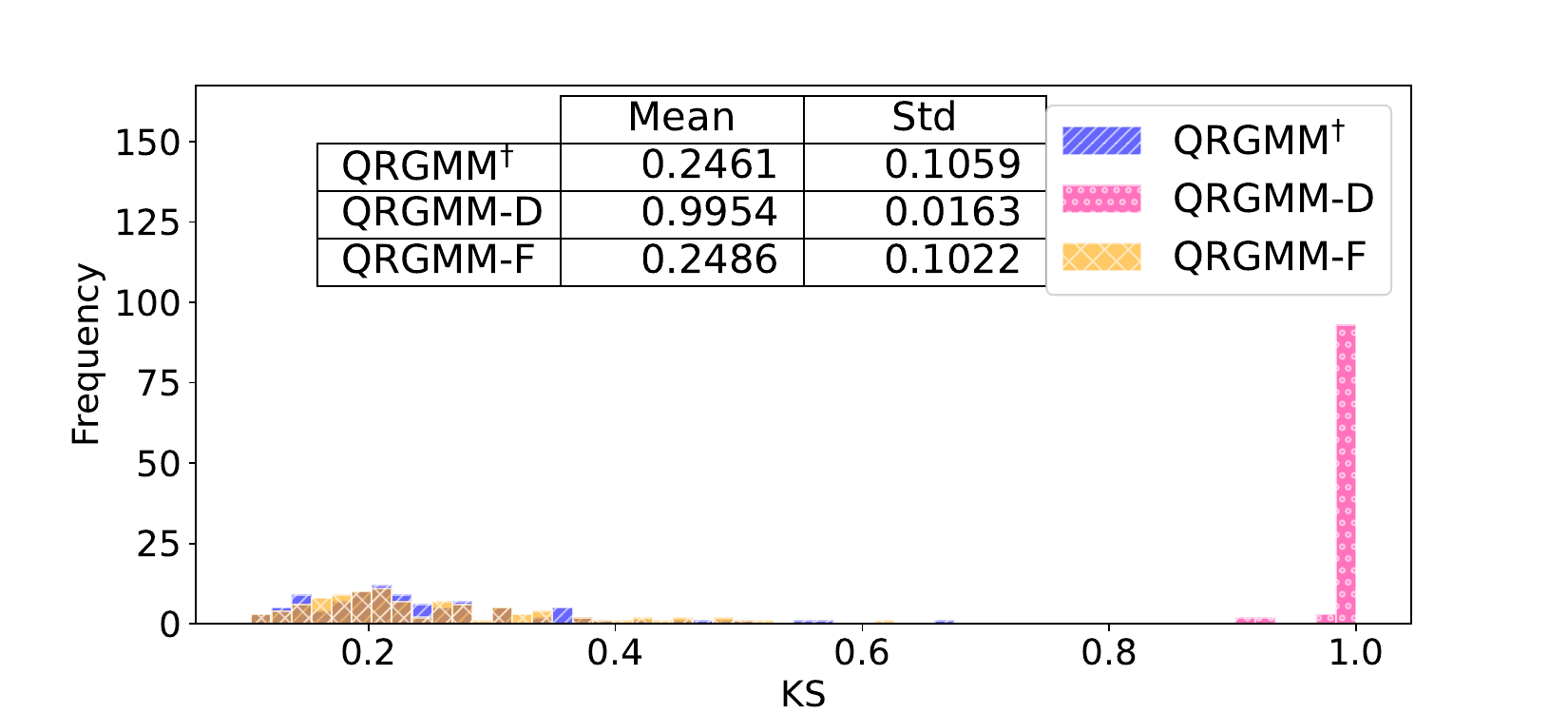}
	\caption{KS for conditional test.}\label{fig:ks_conditional}
\end{figure}

The above comparison is unconditional, assessing overall distributional similarity. Since we know the true data-generating process in the synthetic setting, we can also perform a conditional test to evaluate the generative performance conditioned on a specific covariate vector. For a given covariate vector \(\bx\), we generate conditional datasets of 10,000 observations of \(\hat{Y}(\bx)\) using trained models and compare them with the true conditional dataset of 10,000 observations of \(Y(\bx)\) sampled from the true model. Like the unconditional test, we perform 100 replications, calculating the sample means and standard deviations for each conditional dataset. The averaged results are also included in Table~\ref{table:sync}. The WD and KS values for the conditional test across replications are shown in Figures~\ref{fig:wd_conditional} and \ref{fig:ks_conditional}.

\setlength{\tabcolsep}{3pt} 
\begin{table}[ht]
	\centering
	\caption{Mean and standard deviation (SD) of generated observations.}\label{table:sync}
	\begin{tabular}{lcccc}
		\toprule 
		\multirow{2}{*}{\textbf{Model}} & \multicolumn{2}{c}{\textbf{Unconditional Test}} & \multicolumn{2}{c}{\textbf{Conditional Test}} \\ 
		\cmidrule(l){2-3} \cmidrule(l){4-5} 
		& \textbf{Mean} & \textbf{SD} & \textbf{Mean} & \textbf{SD} \\ 
		\midrule
		\textbf{Truth} & $4355.4783$ & $1430.8676$ & $3948.9500$ & $150.6615$ \\
		\textbf{QRGMM}$^{\dagger}$ & $4350.6580$ & $1412.5260$ & $3948.7998$ & $150.1433$ \\
		\textbf{QRGMM} & $4355.9730$ & $1421.0893$ & $4114.5597$ & $221.1148$ \\
		\textbf{QRGMM-D} & $216.7717$ & $117.6981$ & $205.5038$ & $96.5551$ \\
		\textbf{QRGMM-F} & $4351.5033$ & $1412.2491$ & $3950.5061$ & $150.8685$ \\
		\textbf{CWGAN} & $4320.6589$ & $1394.2039$ & $3915.1583$ & $190.8393$ \\
		
		\bottomrule
	\end{tabular}
\end{table}
\setlength{\tabcolsep}{6pt}

\begin{figure}[ht]
	\centering
	\includegraphics[width=1\linewidth]{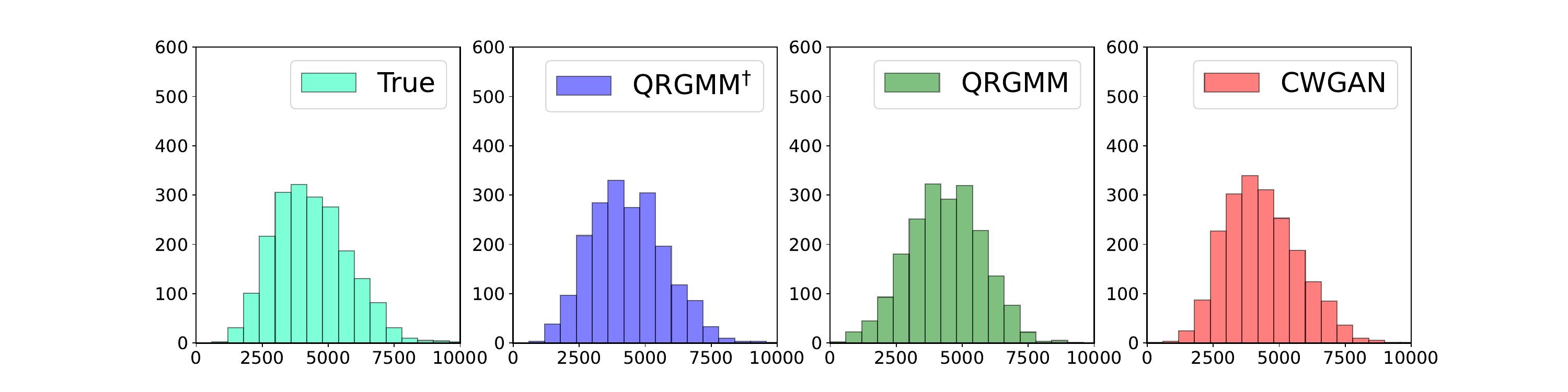}
	\includegraphics[width=1\linewidth]{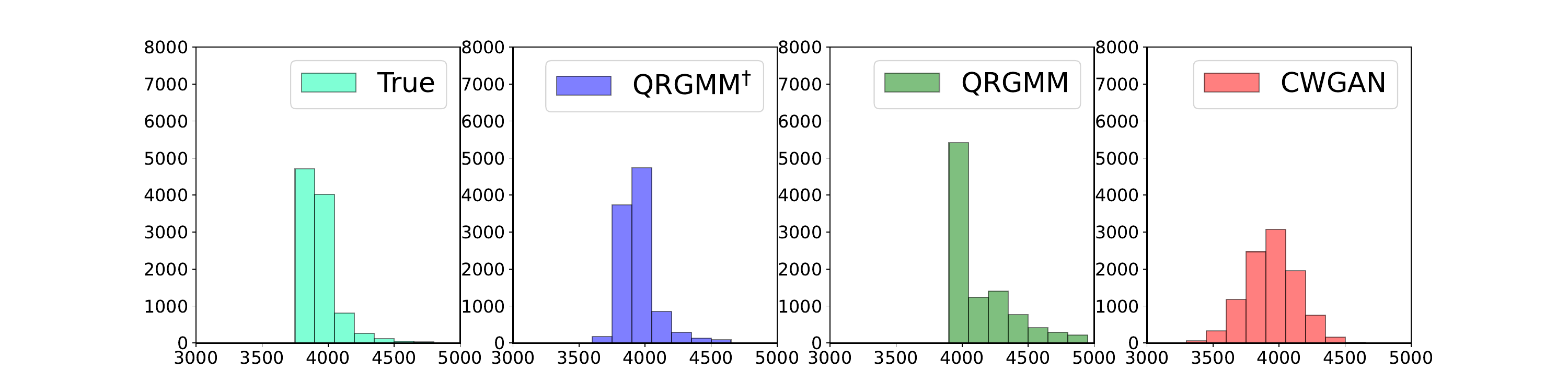}
	\caption{Histograms of generated vs.\ true data for unconditional (upper) and conditional (lower) test.}\label{fig:synthetic_data_histograms}

\end{figure}

Additionally, we visualize the performance of generative models by presenting histograms of the generated and true data distributions from one replication in Figure~\ref{fig:synthetic_data_histograms} for both the unconditional test and conditional test.

\paragraph*{Summary of the Test Results}

Our results demonstrate that QRGMM$^\dagger$ consistently achieves significantly lower WD and KS values compared to CWGAN in both unconditional and conditional tests (see Table~\ref{table:sync} and Figures~\ref{fig:wd_unconditional}--\ref{fig:ks_conditional}). This indicates that QRGMM$^\dagger$ provides a more accurate and robust reproduction of the true data-generating process, whereas CWGAN exhibits considerable variability in performance across different replications. Furthermore, the histograms in Figure~\ref{fig:synthetic_data_histograms} illustrate that QRGMM$^\dagger$-generated observations closely match the true data distribution in both unconditional and conditional settings, whereas CWGAN-generated observations exhibit noticeable discrepancies in conditional settings.

Although the original linear QRGMM attains slightly better mean and standard deviation alignment in the unconditional test, WD/KS values and visual plots reveal that its learned distribution is less accurate than QRGMM$^\dagger$. In conditional tests, QRGMM$^\dagger$ clearly outperforms all other methods, highlighting its reliability in decision-making contexts where conditioning is essential. The results also demonstrate that it is essential to incorporate the DeepFM component in the model.

The ablation studies further shed light on the role of the deep neural network and factorization machine components. The results suggest that the factorization machine component plays a crucial role in enabling the model to quickly capture salient structural information. While the deep neural network component is more expressive and in principle subsumes the factorization machine component, using it alone makes it difficult for the model to learn accurate distributions. QRGMM-F performs much better than QRGMM-D but still falls short of the full QRGMM$^\dagger$. 
Overall, the results highlight the benefit of incorporating DeepFM into QRGMM$^\dagger$, while underscoring its robustness and fidelity in capturing complex data distributions.

\subsubsection{Risk Measures Estimation}\label{sec:riskapproximation}

In this synthetic-data experiment, for a specified $\bx$, the data-generating process of \( Y(\bx) \) is known. It enables us to analytically compute true risk measures as benchmarks for assessing the performance of QRGMM$^\dagger$ estimators. In our setting, the loss function \( L(l) \) defined in Equation \eqref{eq:loss} is used, and the maximum loan amount \( \bar{l} \) is set to the 99.9th percentile of \( rY(\bx) \), with net unit revenue \( r = 1 \). Loan levels \( l \) are evaluated at 100 equally spaced points across \( [2000, \bar{l}] \). 

\begin{figure*}[ht]
	\centering
	\includegraphics[width=0.8\linewidth]{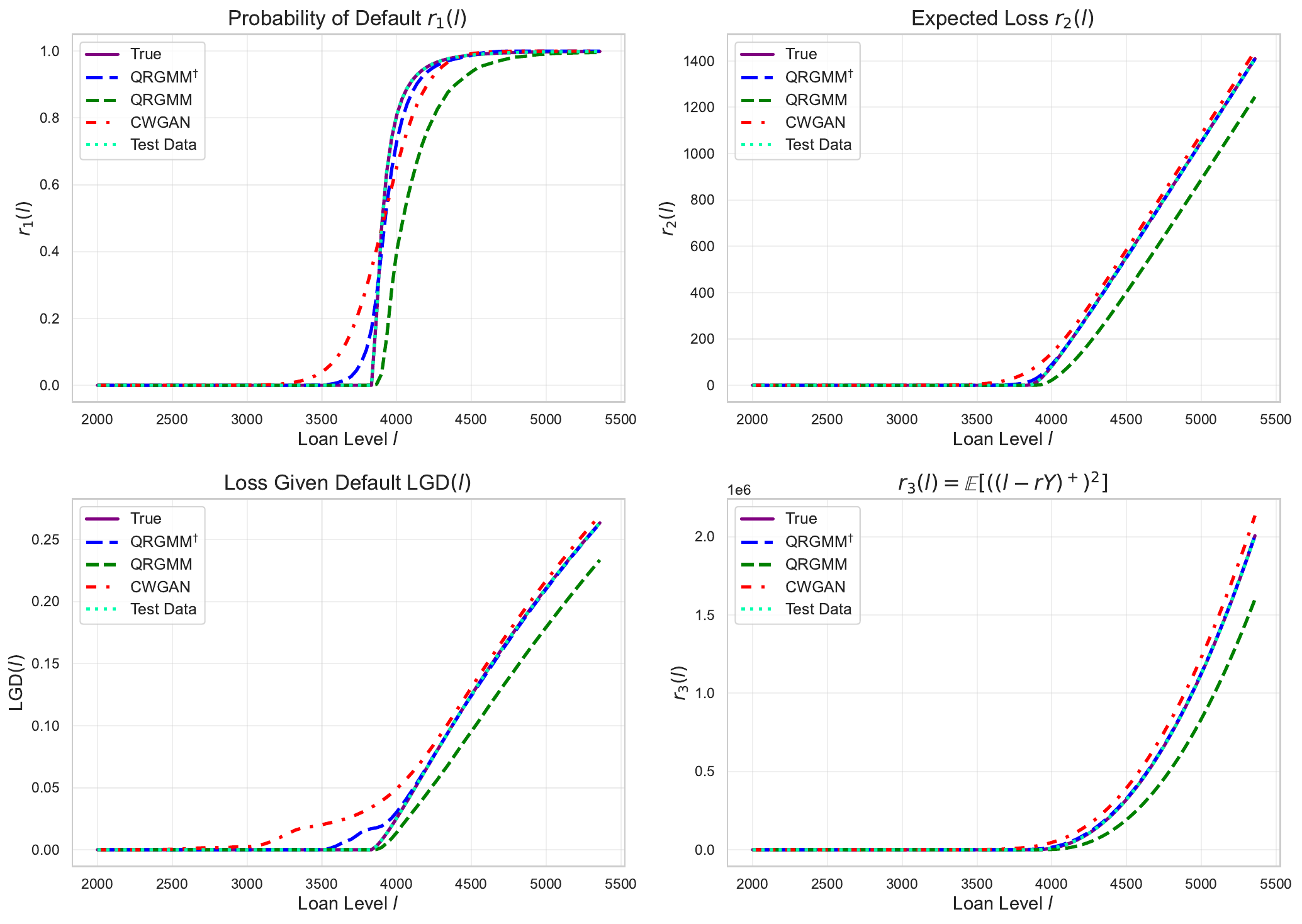}
	\caption{Estimation of risk measures across loan levels \( l \).}\label{fig:risk_measures_plot}
\end{figure*}

We compute four true risk measures: \( r_1(l) \) (probability of default), \( r_2(l) \) (expected loss), and \( \text{LGD}(l) \) (loss given default), as specified in Section \ref{sec:riskformulation}, and \( r_3(l) = \mathbb{E}\left[\left(\left(l - rY(\bx)\right)^+\right)^2\right] \), a specific form of the generalized risk measure highlighting severe losses. QRGMM$^\dagger$, QRGMM, and CWGAN estimators are calculated over 100 replications, each with 10,000 observations, and their means are compared to the true theoretical values, as well as the means of the estimators using the true test data observations calculated over 100 replications. These results are depicted in Figure \ref{fig:risk_measures_plot}, with subplots for four risk measures \( r_1(l) \), \( r_2(l) \), \( \text{LGD}(l) \), and \( r_3(l) \).

The results are highly encouraging: the QRGMM$^\dagger$ estimators consistently track the true values across all risk measures, demonstrating robust accuracy. By contrast, QRGMM and CWGAN estimators show noticeable discrepancies overall, indicating weaker performance.
This comparison underscores QRGMM$^\dagger$'s superior capability in estimating complex risk measures, demonstrating its applicability in SCF credit risk management. 

Furthermore, Figure \ref{fig:risk_measures_plot} also offers a clear, immediate view of how risk measures evolve with loan amount \( l \), providing decision-makers with an intuitive tool for understanding the risk dynamics, thereby greatly facilitating loan sizing decisions.

The Online Appendix~C examines additional conditional tests at different covariate points. The results confirm that changes in covariates lead to noticeable shifts in conditional distributions and loan sizing outcomes, while QRGMM$^\dagger$ estimators remain stable and accurate across these variations.

\subsection{Real-World Data Experiments}\label{sec:real}

\subsubsection{Big Mart Sales Dataset}\label{sec:bigmart}

To evaluate QRGMM$^\dagger$ in a practical setting, we apply it to the Big Mart sales dataset \citep{shrivas2013}. This dataset contains 8,523 observations of 1,559 products across 10 stores, along with various product-level and store-level attributes. 
We randomly split the data into training and test subsets with a 4:1 ratio across 100 replications, and set \(m=80\). 

\begin{figure}[ht]
	\centering
	\includegraphics[width=1\linewidth]{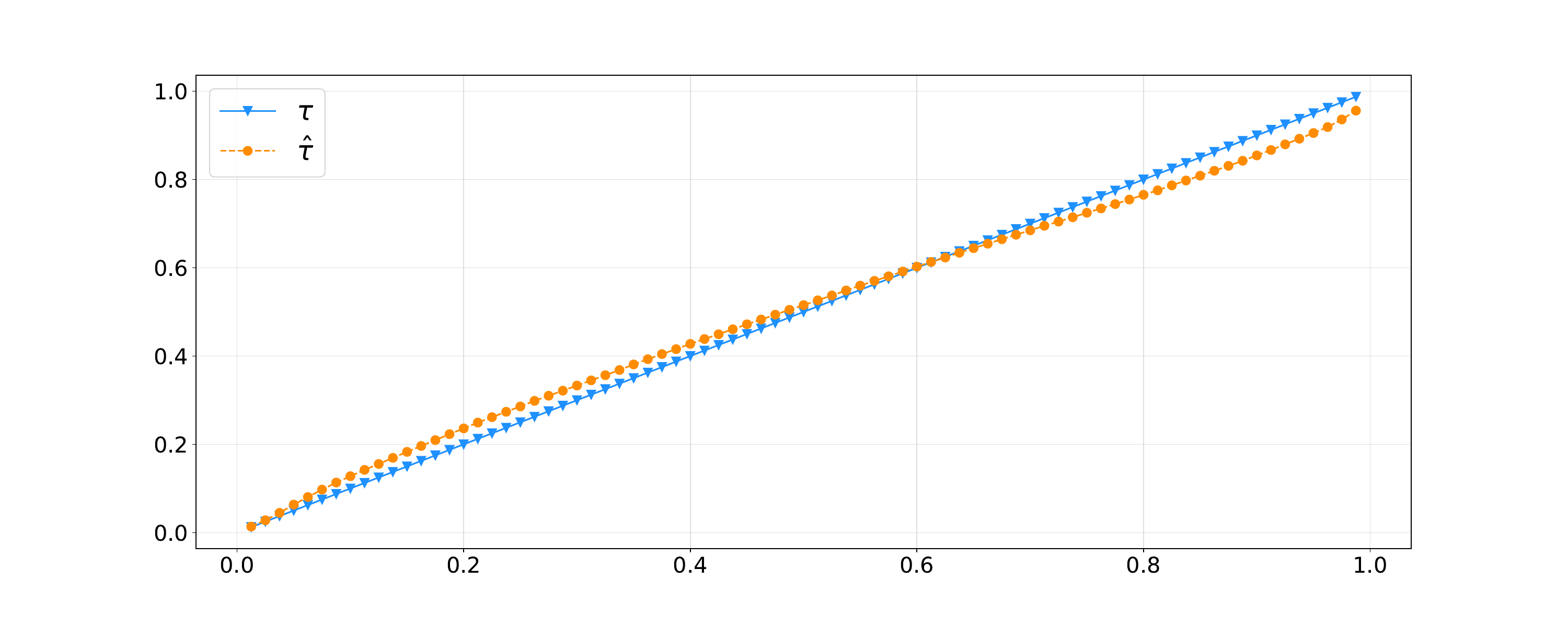}
	\caption{Alignment of nominal and empirical quantile levels on the Big Mart sales data.}\label{fig:tauhat_bigmart}
\end{figure}

\paragraph*{Quantile Estimation}
As in Section~\ref{experiment_synthetic}, we first examine the accuracy of the model's quantile estimation. For each test observation, we estimate \(\tau\)-quantiles and compute the empirical proportion of actual sales below these predicted values, \(\hat{\tau}\). Figure~\ref{fig:tauhat_bigmart} shows that \(\hat{\tau}\) closely aligns with \(\tau\) across 100 replications, indicating that QRGMM$^\dagger$ continues to offer reliable quantile estimates on real-world data.

\paragraph*{Unconditional Distribution Comparison}
Since the true data-generating mechanism is unknown in real-world datasets, we cannot perform a conditional test as in the synthetic experiment. Instead, we focus on an unconditional comparison of the generated and observed sales distributions. We trained the models on the training set and subsequently constructed a generated test dataset for each model across 100 replications.

\begin{figure}[ht]
	\centering
	\includegraphics[width=0.49\textwidth]{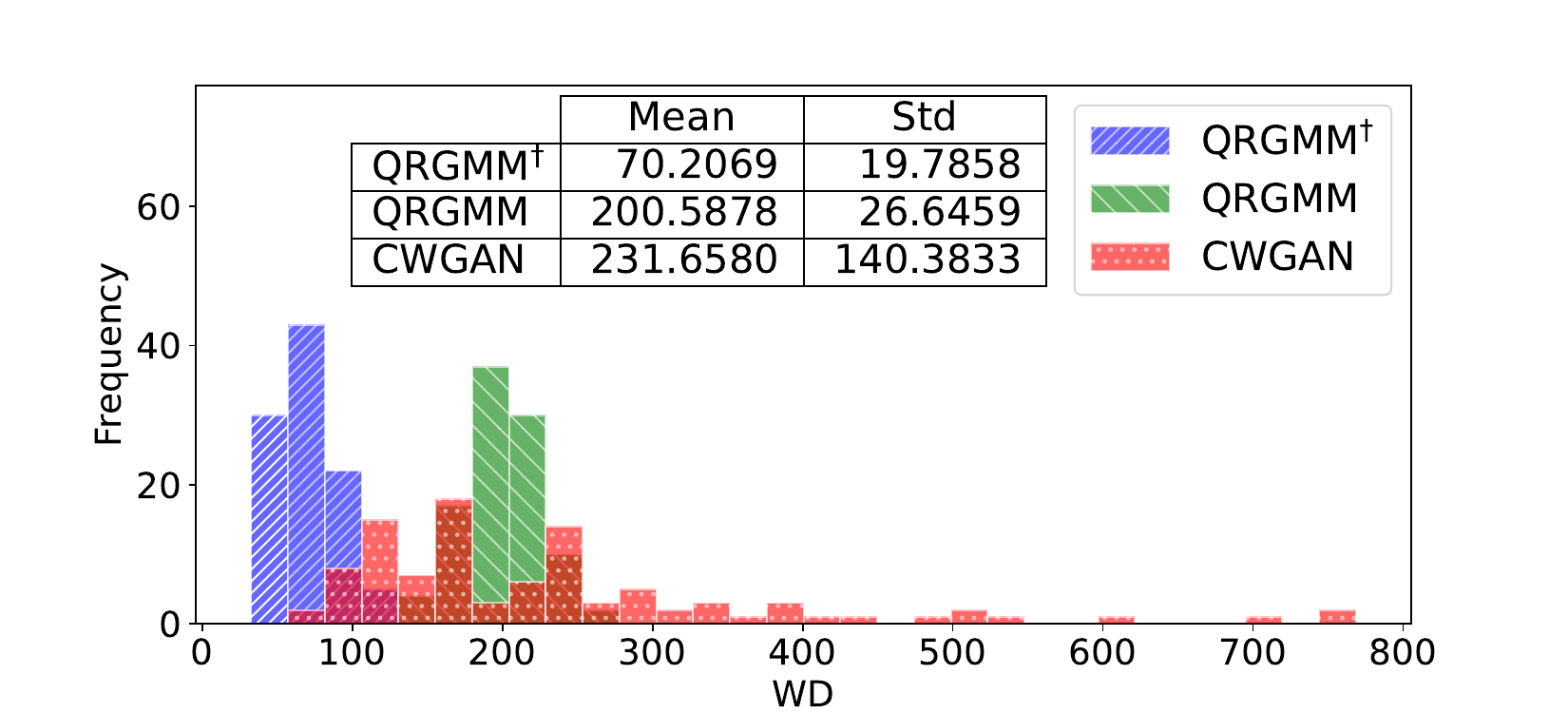} 
	\includegraphics[width=0.49\textwidth]{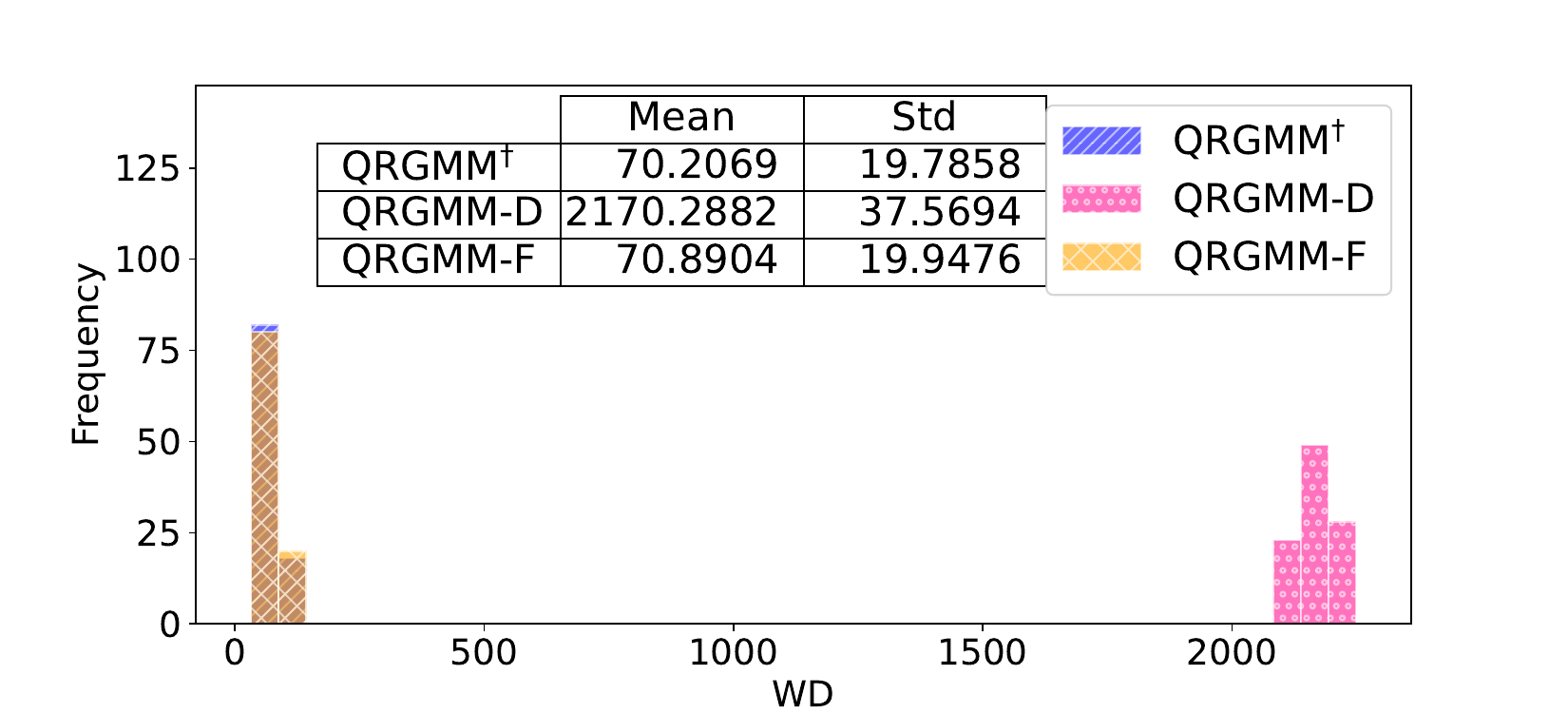}
	\caption{WD for Big Mart sales.}\label{fig:bigmart_WD_uc}
\end{figure}

\begin{figure}[ht]
	\centering
	\includegraphics[width=0.49\textwidth]{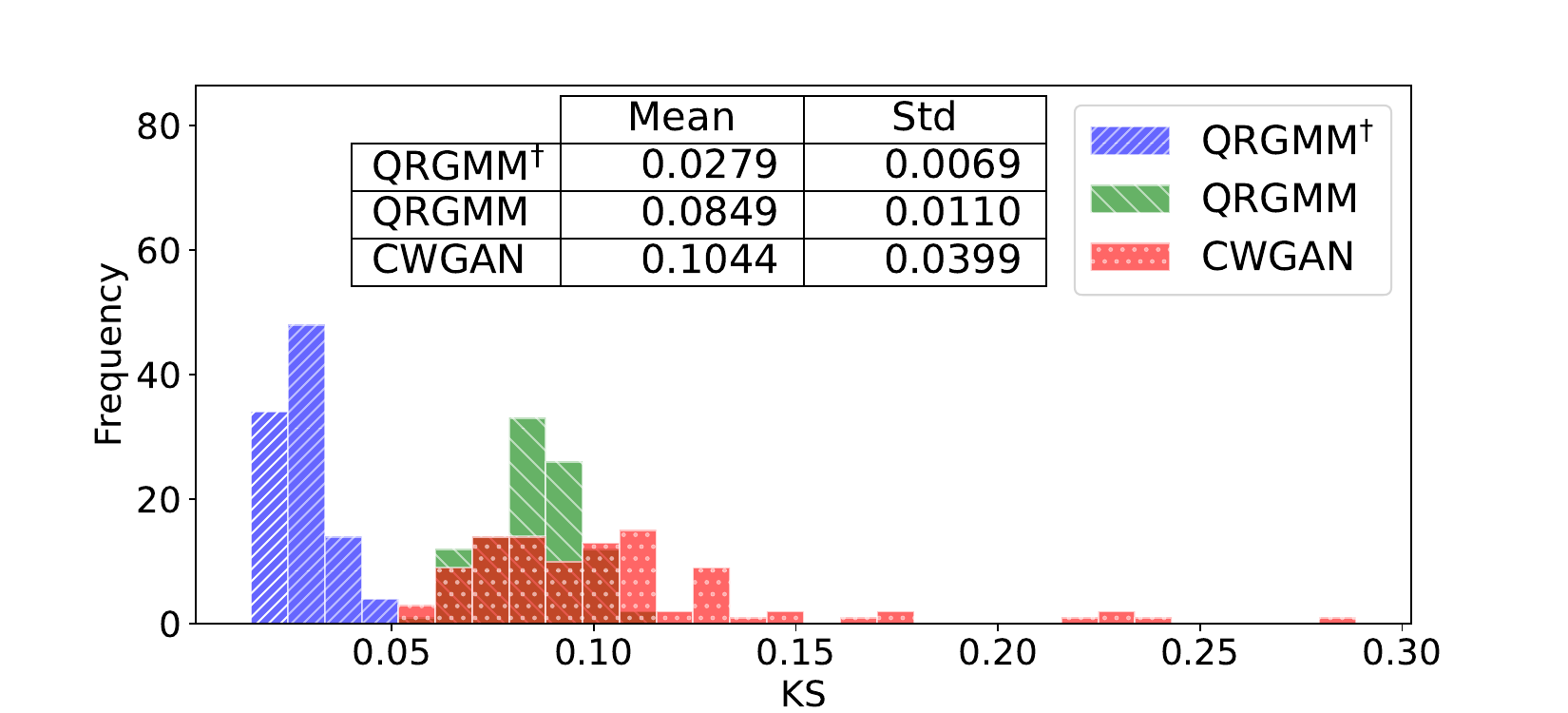} 
	\includegraphics[width=0.49\textwidth]{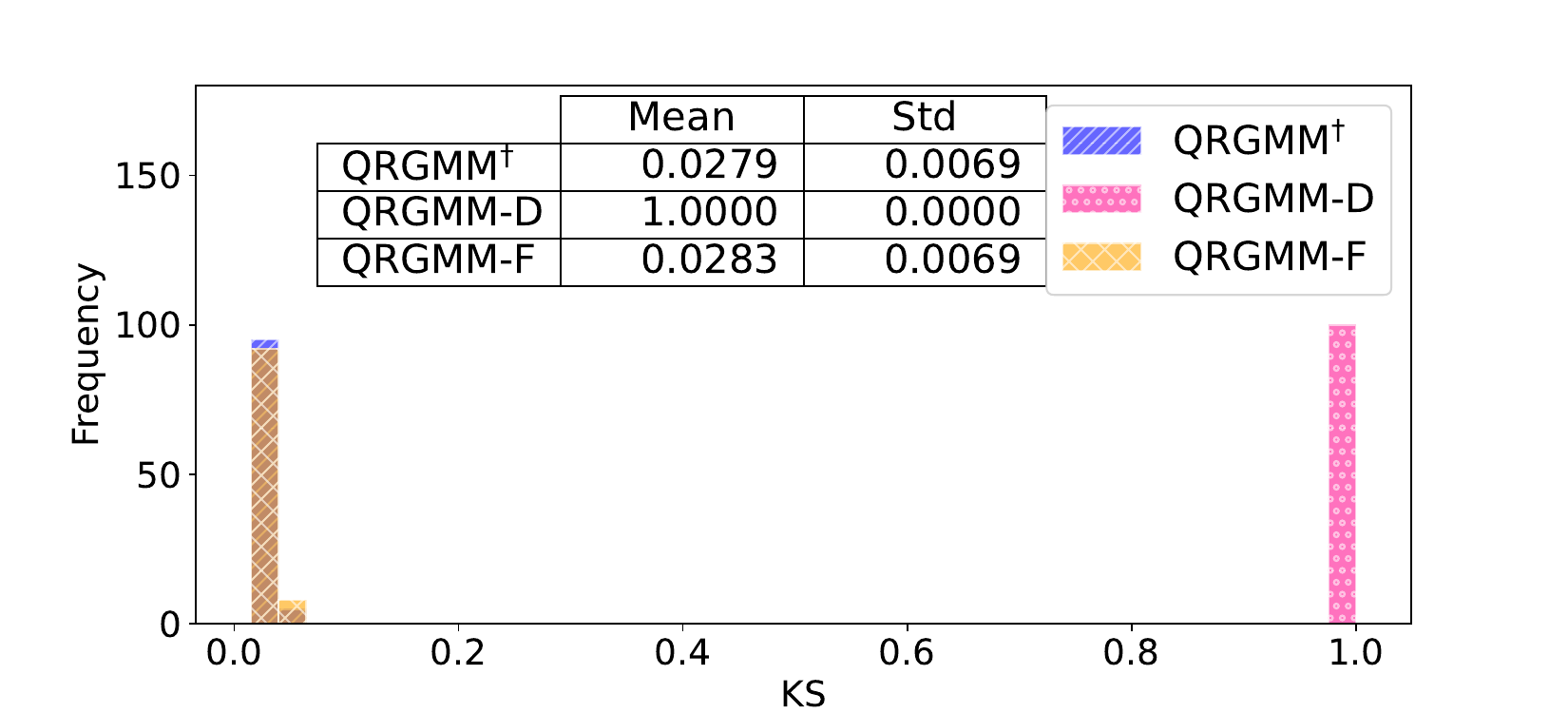}
	\caption{KS for Big Mart sales.}\label{fig:bigmart_KS_uc}
\end{figure}

Table~\ref{table:bigmart} summarizes the means and standard deviations (averaged over the 100 replications) of both the generated and observed test sales. QRGMM$^\dagger$ and QRGMM show closer matches to the true data than CWGAN. To further quantify distributional similarity, we compute the WD and KS values for each replication, and plot their distributions in Figures~\ref{fig:bigmart_WD_uc} and \ref{fig:bigmart_KS_uc}. QRGMM$^\dagger$ again exhibits lower WD and KS values than CWGAN and QRGMM, confirming its superior generative ability in this real-world scenario.

Consistent with the experiments in Section~\ref{experiment_synthetic}, the results here reaffirm that the factorization machine component is essential while the deep neural network component provides complementary refinement, and their integration in QRGMM$^\dagger$ delivers the most accurate generative performance.

\setlength{\tabcolsep}{16pt}

\begin{table}[ht]
	\centering
	\caption{Mean and standard deviation of generated observations.}\label{table:bigmart}
	\begin{tabular}{lcc}
		\toprule 
		\textbf{Model} & \textbf{Mean} & \textbf{SD} \\ 
		\midrule
		\textbf{Truth} & $2177.0654$ & $1708.2110$ \\
		\textbf{QRGMM}$^{\dagger}$ & $2168.6467$ & $1682.5005$ \\
		\textbf{QRGMM} & $2173.0349$ & $1637.7796$ \\
		\textbf{QRGMM-D} & $6.7772$ & $3.9158$ \\
		\textbf{QRGMM-F} & $2168.0144$ & $1681.5131$ \\
		\textbf{CWGAN} & $2202.2440$ & $1647.7403$ \\
		\bottomrule
	\end{tabular}
\end{table}

\setlength{\tabcolsep}{6pt}

\begin{figure}[ht]
	\centering
	\includegraphics[width=1\linewidth]{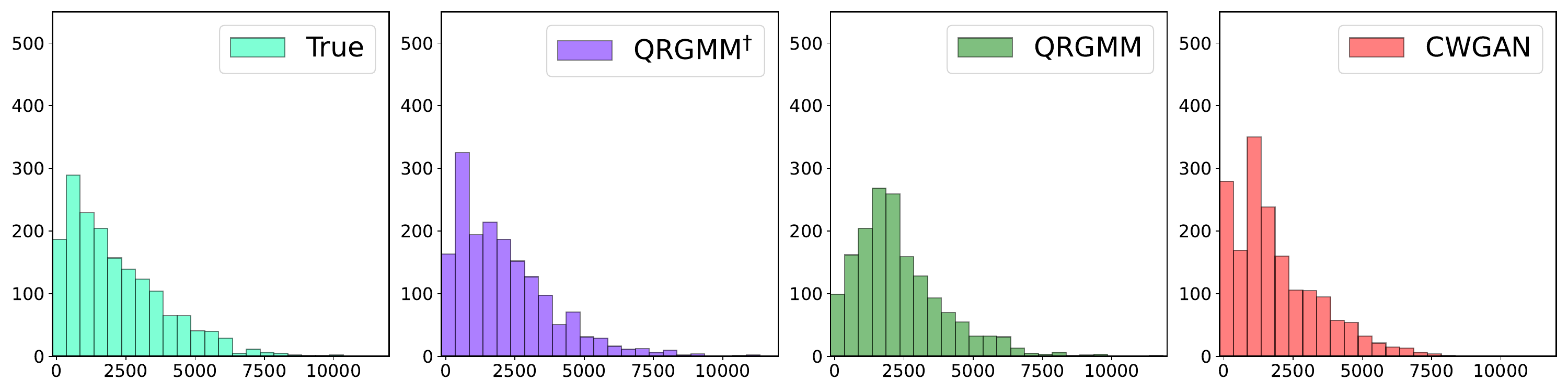}
	\caption{Histogram comparison of observed vs.\ generated Big Mart sales (unconditional test).}\label{fig:bigmart_hist_uc}
\end{figure}

Figure~\ref{fig:bigmart_hist_uc} illustrates the observed and generated sales distributions in one representative replication. The distribution generated by QRGMM$^\dagger$ aligns more closely with the empirical distribution than those generated by CWGAN and QRGMM.

\paragraph*{Risk Measures Estimation}

For the Big Mart sales dataset, the conditional data-generating process of $Y(\bx)$ is not directly available, which prevents us from analytically computing the true conditional risk measures. To address this, we compute risk measures using the unconditional distribution of the original sales data and use them as benchmarks. Specifically, we evaluate the same four risk measures, $r_1(l)$, $r_2(l)$, $\text{LGD}(l)$, and $r_3(l)$, as described in Section~\ref{sec:riskapproximation}, using $\bar{l}$ as the 99th percentile of sales, and compare the estimators from QRGMM$^\dagger$, QRGMM, and CWGAN with the benchmarks.

\begin{figure*}[ht]
	\centering
	\includegraphics[width=0.8\linewidth]{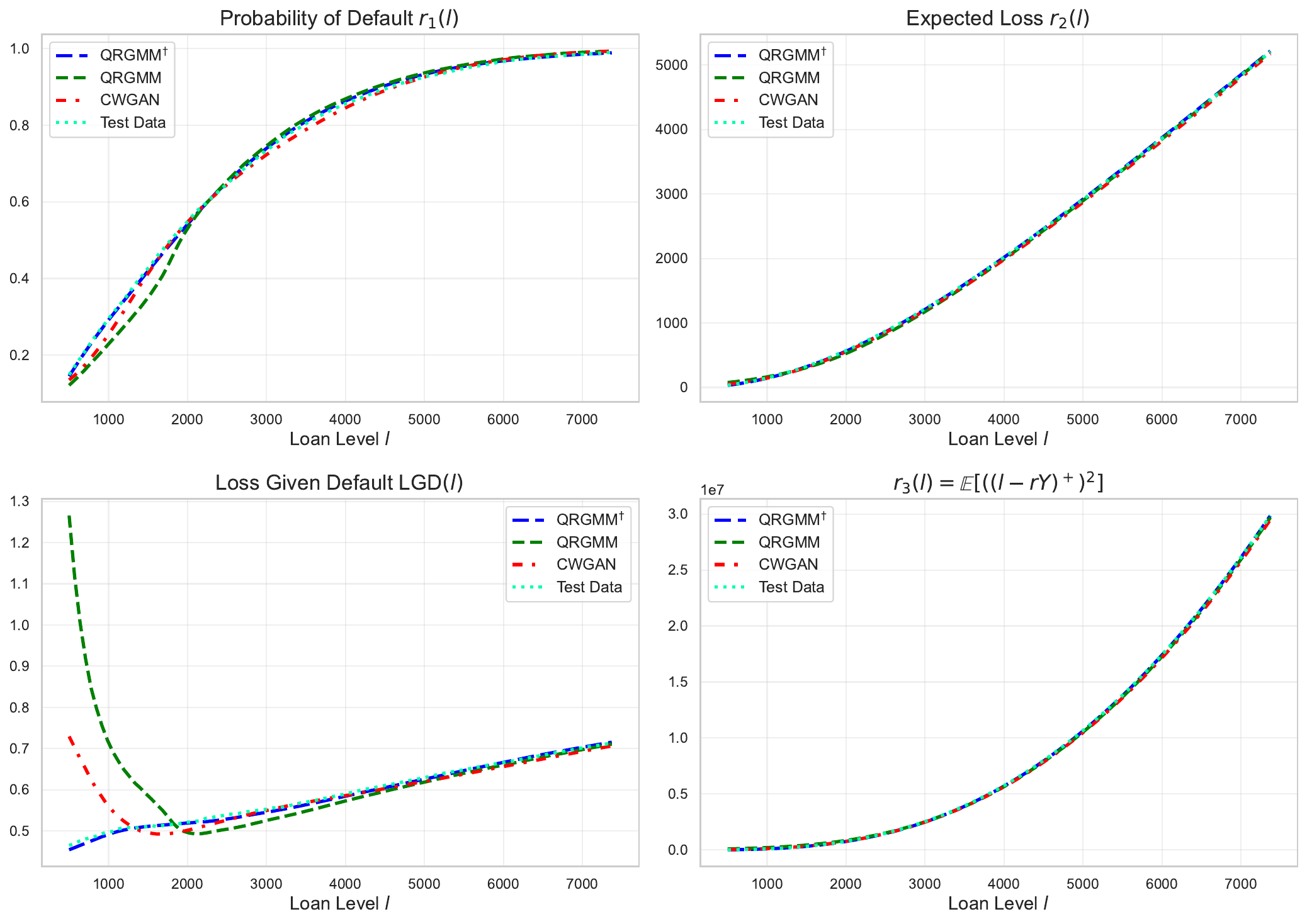}
	\caption{Estimation of risk measures across loan levels \( l \).}\label{fig:bigmart_unconditional_risk_measures_plot}
\end{figure*}

The results, reported in Figure~\ref{fig:bigmart_unconditional_risk_measures_plot}, show that QRGMM$^\dagger$ overall provides the closest approximations to the empirical benchmarks across all measures, consistently outperforming QRGMM and CWGAN. In particular, the estimators of $r_1(l)$ and $\text{LGD}(l)$ from QRGMM and CWGAN deteriorate when $l < 2000$, as these models fail to adequately capture the left tail of the sales distribution (see also Figure~\ref{fig:bigmart_hist_uc}). By contrast, QRGMM$^\dagger$ maintains strong performance in this region and achieves superior fidelity across the entire range, highlighting its robustness and accuracy in estimating risk measures for real-world sales data.

\paragraph*{Discussion}

In summary, QRGMM$^\dagger$ demonstrates consistently strong performance on the Big Mart sales dataset, excelling in quantile estimation, distribution generation, and risk measures estimation. Although a conditional test is infeasible due to the unknown data-generating mechanism, unconditional evaluations, including distributional measures, visual diagnostics, and risk measure benchmarks, consistently show QRGMM$^\dagger$ outperforming both QRGMM and CWGAN. These findings highlight the practicality of QRGMM$^\dagger$ for complex SCF applications.

\subsubsection{Amazon CBEC Dataset}\label{sec:amazon}

To further assess the effectiveness of QRGMM$^\dagger$ in a real-world cross-border e-commerce (CBEC) context, we also conduct experiments on a dataset from a company conducting China-to-U.S. CBEC on Amazon. This dataset includes 3,495 sales records from the first quarter of 2023, covering 99 product categories across 131 stores. The products are managed by 23 distinct product selection teams and 23 sales teams, forming various team pairings. As noted earlier, these pairings exhibit combinatorial effects due to differences in team capability and cross-team collaboration. The dataset also features seven continuous variables: advertising expenses, clicks, impressions, sale price, product cost, average rating, and number of reviews. Again, we randomly split the data into training and test subsets with a 4:1 ratio across 100 replications. And we set \( m = 60 \).

\begin{figure}[ht]
	\centering
	\includegraphics[width=1\linewidth]{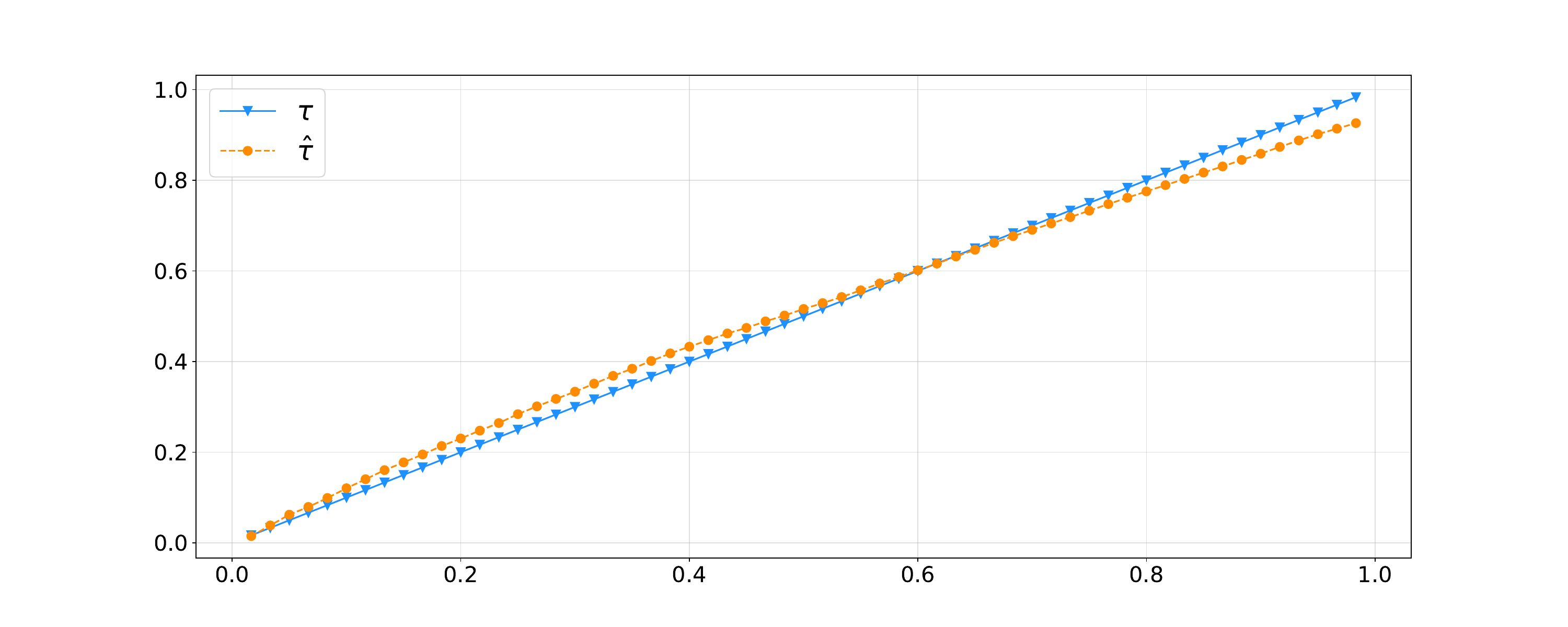}
	\caption{Alignment of nominal and empirical quantile levels on the Amazon CBEC data.}\label{fig:tauhat_amazon}
\end{figure}

\paragraph*{Quantile Estimation}
Similarly to Section~\ref{sec:bigmart}, we begin by evaluating the model's quantile estimation accuracy. For each test observation, we compute the empirical coverage \(\hat{\tau}\) of the predicted \(\tau\)-quantiles across 100 replications. As shown in Figure~\ref{fig:tauhat_amazon}, \(\hat{\tau}\) aligns well with the nominal \(\tau\), confirming that QRGMM$^\dagger$ maintains reliable quantile estimation on this CBEC dataset.

\paragraph*{Unconditional Distribution Comparison}
Since the true conditional distribution remains unknown, we again adopt an unconditional evaluation strategy. All models are trained on the training set and used to construct generated test sales data over 100 replications.

\begin{figure}[ht]
	\centering
	\includegraphics[width=0.49\textwidth]{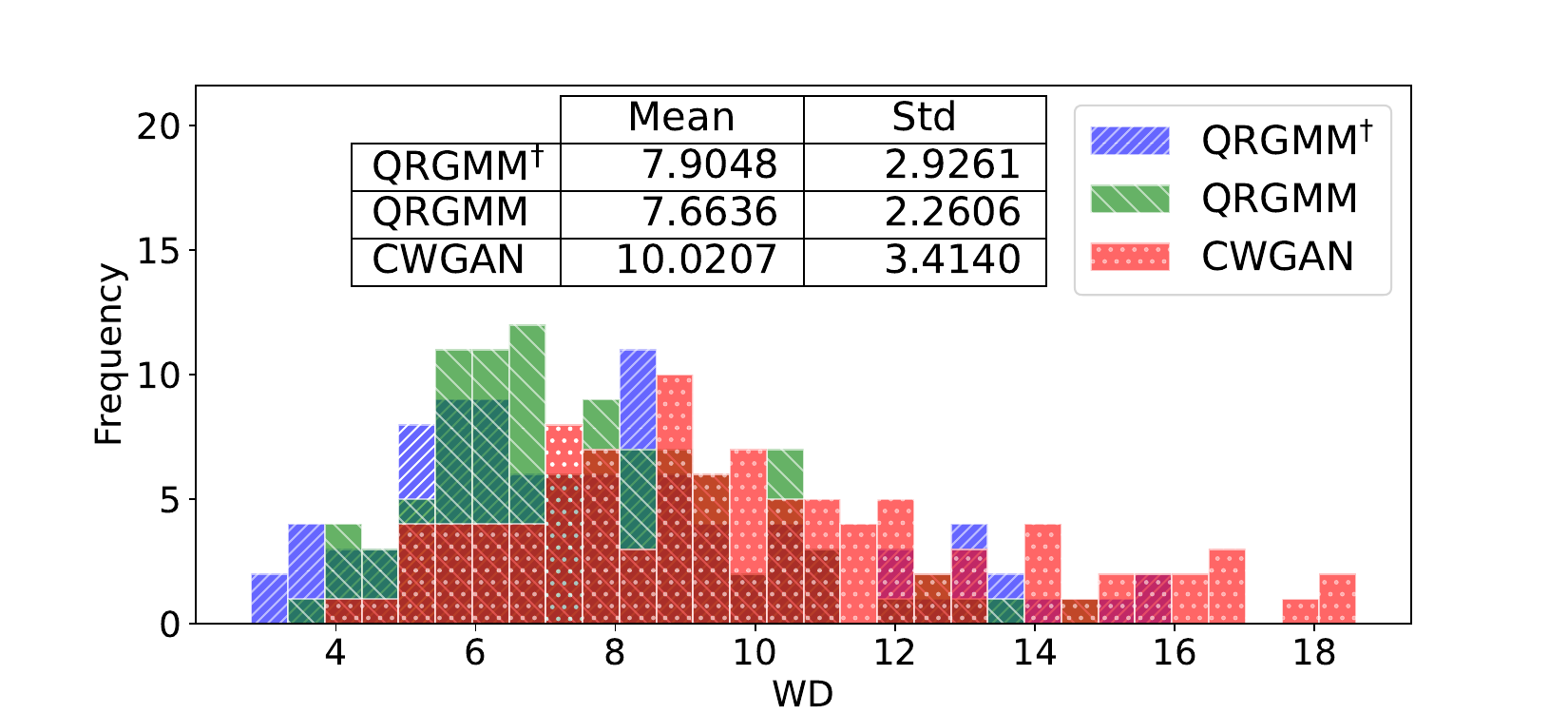} 
	\includegraphics[width=0.49\textwidth]{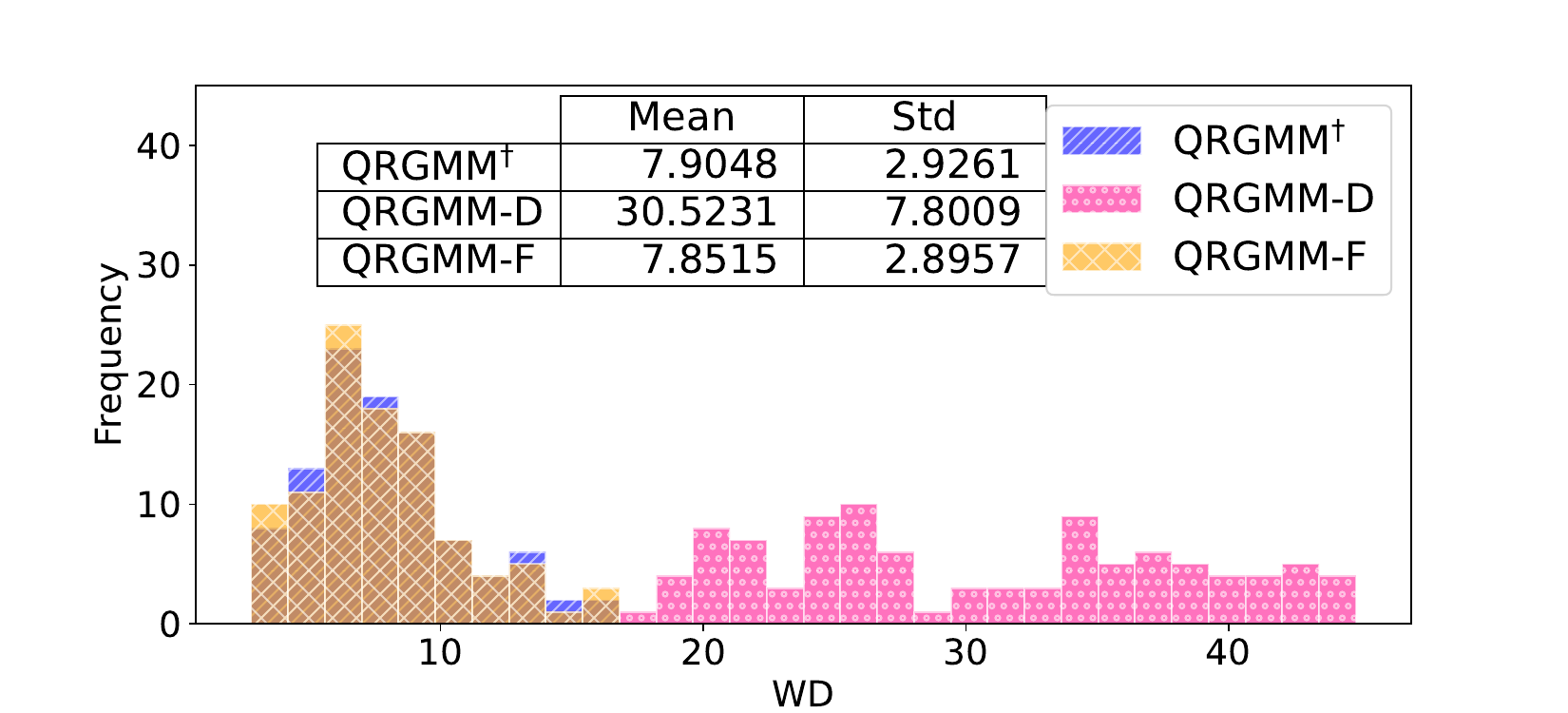}
	\caption{WD for Amazon CBEC sales.}\label{fig:amazon_WD_uc}
\end{figure}

\begin{figure}[ht]
	\centering
	\includegraphics[width=0.49\textwidth]{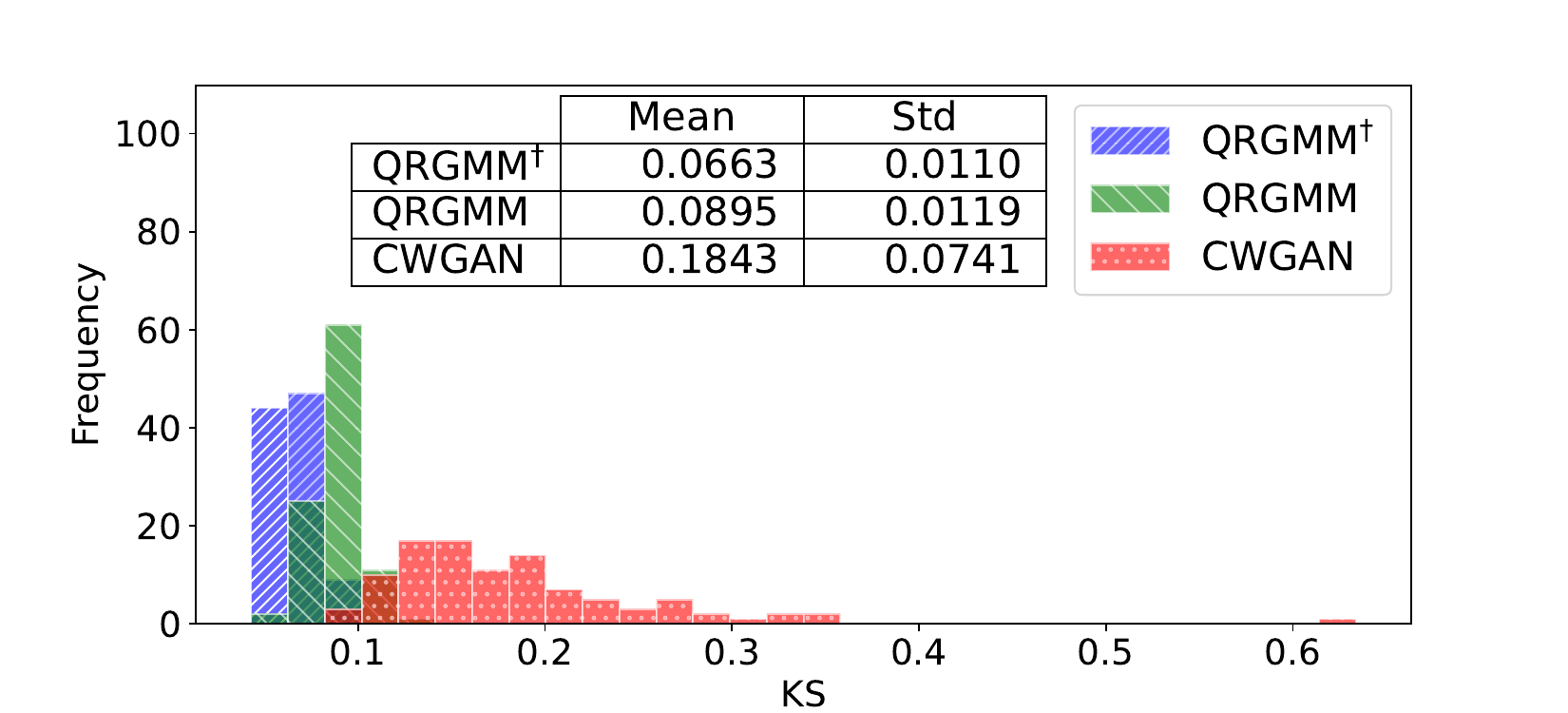} 
	\includegraphics[width=0.49\textwidth]{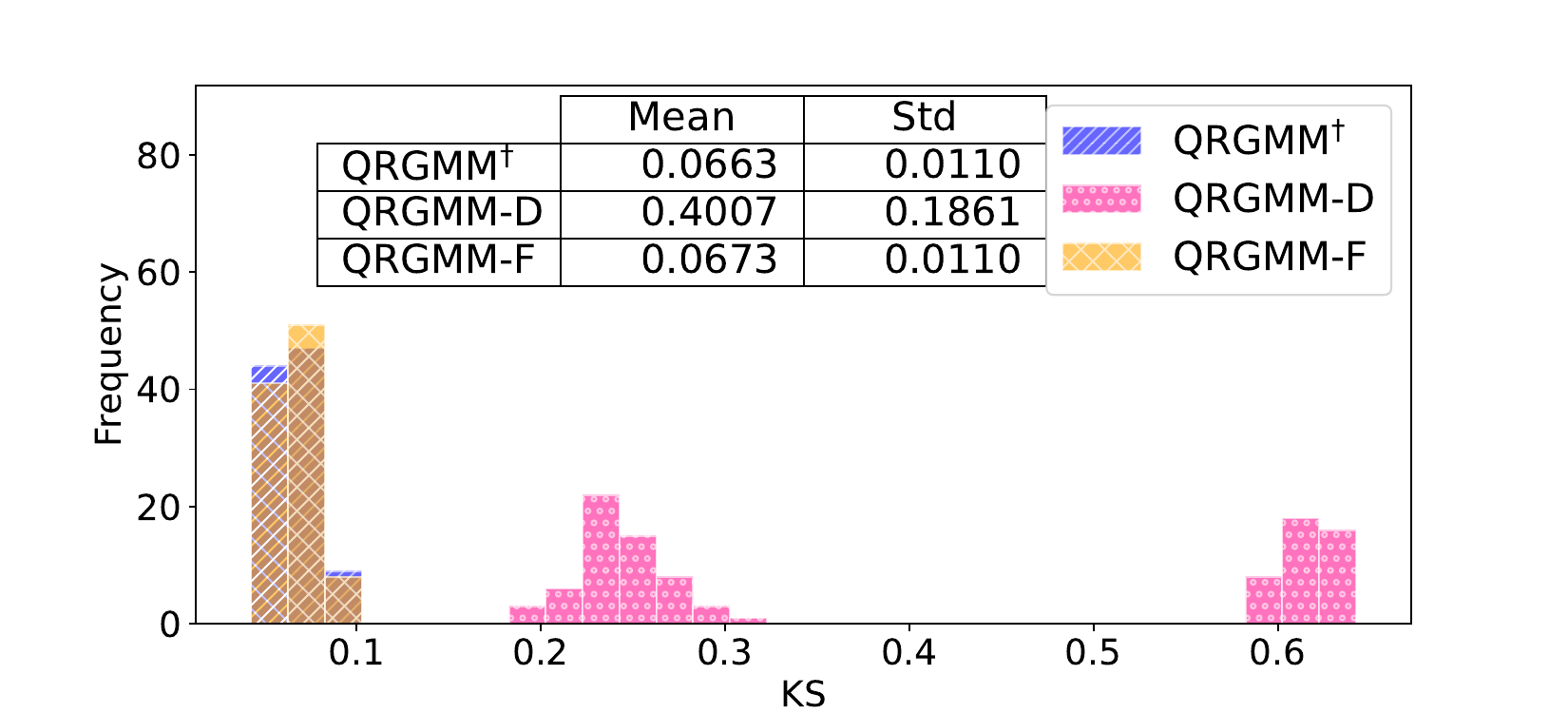}
	\caption{KS for Amazon CBEC sales.}\label{fig:amazon_KS_uc}
\end{figure}

Table~\ref{table:amazon} reports the means and standard deviations (averaged across 100 replications) of the generated and observed test sales, while Figures~\ref{fig:amazon_WD_uc} and \ref{fig:amazon_KS_uc} compare the WD and KS values across replications. Overall, QRGMM$^\dagger$ continues to perform strongly, aligning closely with the true data and consistently yielding lower WD and KS values than CWGAN. In this case, QRGMM also performs reasonably well, indicating that the basic version remains effective in data-limited settings. These results confirm the robustness of QRGMM$^\dagger$ across different data conditions, while providing empirical evidence that the simpler QRGMM can also be competitive in practice when the sample size is limited.

Consistent with the synthetic and Big Mart results, the results on the Amazon CBEC dataset further confirm that the factorization machine is critical, which provides efficient second-order feature interaction learning in SCF applications, while the deep neural network provides complementary refinement, and their integration in QRGMM$^\dagger$ achieves the best performance.

\setlength{\tabcolsep}{20pt}

\begin{table}[ht]
	\centering
	\caption{Mean and standard deviation of generated observations.}\label{table:amazon}
	\begin{tabular}{lcc}
		\toprule 
		\textbf{Model} & \textbf{Mean} & \textbf{SD} \\ 
		\midrule
		\textbf{Truth} & $44.2296$ & $102.2226$ \\
		\textbf{QRGMM}$^{\dagger}$ & $39.3147$ & $76.7319$ \\
		\textbf{QRGMM} & $43.3484$ & $94.6634$ \\
		\textbf{QRGMM-D} & $13.8928$ & $29.8903$ \\
		\textbf{QRGMM-F} & $39.2903$ & $76.7562$ \\
		\textbf{CWGAN} & $39.8529$ & $76.0303$ \\
		\bottomrule
	\end{tabular}
\end{table}

\begin{figure}[ht]
	\centering
	\includegraphics[width=1\linewidth]{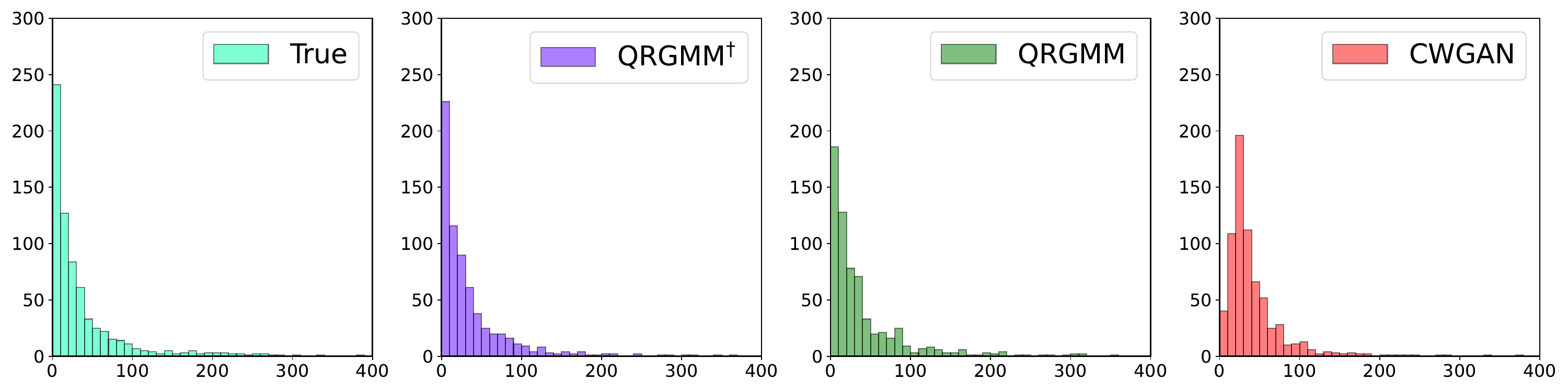}
	\caption{Histogram comparison of observed vs.\ generated Amazon CBEC sales (unconditional test).}\label{fig:amazon_hist_uc}
\end{figure}

Figure~\ref{fig:amazon_hist_uc} presents a histogram comparison of observed and generated sales distributions from one representative replication. As with the Big Mart dataset, the QRGMM$^\dagger$-generated observations align more closely with the observed sales, whereas CWGAN shows noticeable deviation.

\paragraph*{Risk Measures Estimation}

\begin{figure*}[ht]
	\centering
	\includegraphics[width=0.8\linewidth]{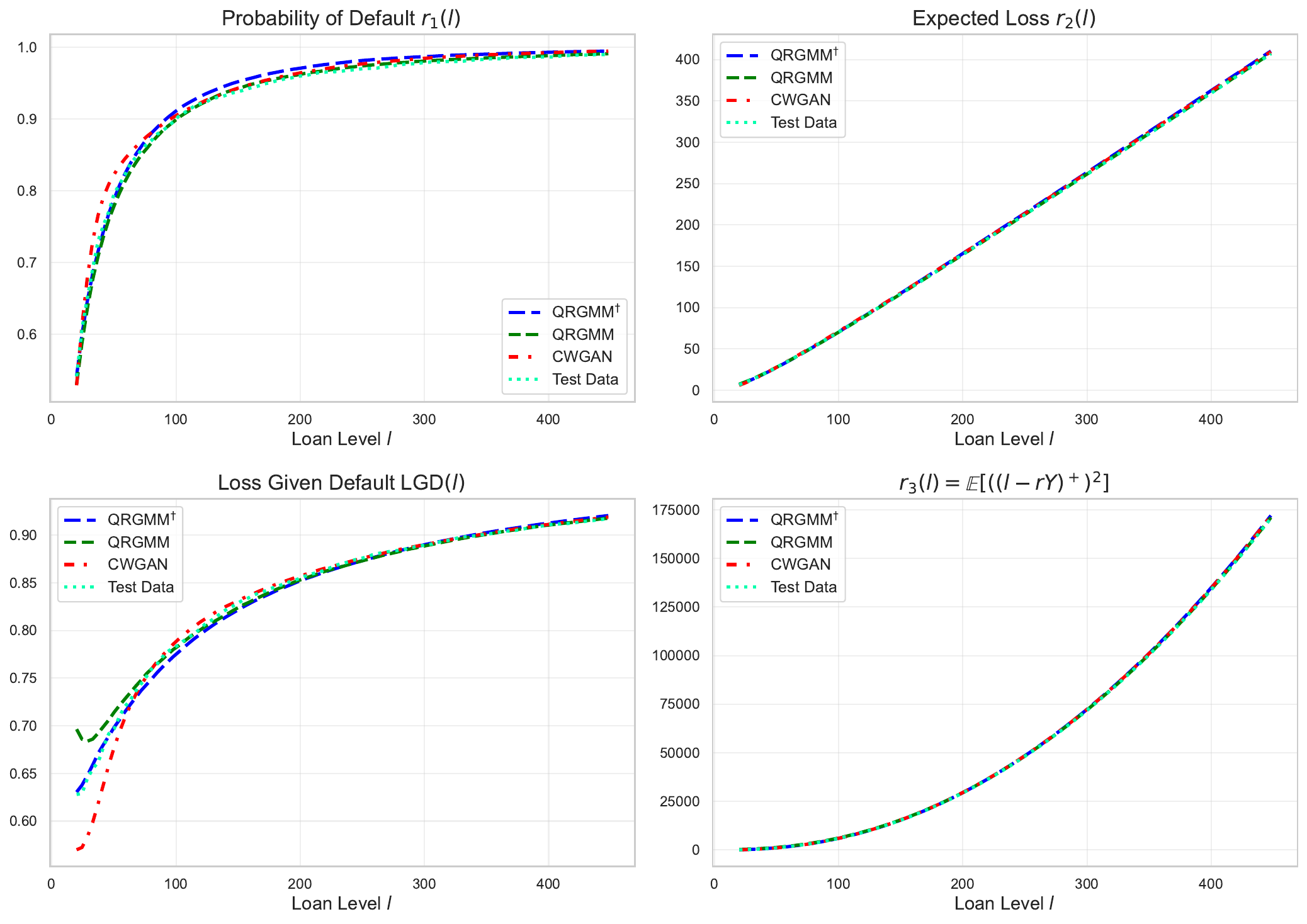}
	\caption{Estimation of risk measures across loan levels \( l \).}\label{fig:Amazon_unconditional_risk_measures_plot}
\end{figure*}

Similar to the Big Mart sales dataset, we also compute risk measures for the Amazon CBEC dataset using the unconditional distribution of the observed sales data and use them as benchmarks. As illustrated in Figure~\ref{fig:Amazon_unconditional_risk_measures_plot}, overall, QRGMM$^\dagger$ appears to yield closer approximations to the empirical benchmarks, with performance generally exceeding those of QRGMM and CWGAN. The performance gap is most evident when $l < 100$, where QRGMM and CWGAN fail to capture the left tail of the sales distribution, whereas QRGMM$^\dagger$ preserves stable and accurate estimation.

\paragraph*{Discussion}

Overall, QRGMM$^\dagger$ continues to demonstrate robust performance on the Amazon CBEC dataset across quantile estimation, unconditional distribution generation, and risk measures estimation. Despite the relatively small sample size and the complex interaction effects among product selection teams and sales teams, QRGMM$^\dagger$ effectively captures the underlying data structure and provides stable estimation across all measures, underscoring its practical utility in complex CBEC environments and validating its applicability to SCF risk management. 

\begin{remark}
	
	From a deployment perspective, the proposed framework can be integrated into existing credit assessment pipelines with only minor adjustments, as it primarily requires generative models of sales distributions that can be coupled with the standard risk evaluation modules already employed by financial institutions. Moreover, the data requirements are realistic within the SCF context, and the datasets used in our experiments reflect practical availability. On the one hand, lenders can draw upon large-scale historical sales datasets from e-commerce platforms such as Amazon, whether accumulated through prior lending experience, dedicated collection efforts, or licensed purchases from data providers. On the other hand, individual small- and medium-sized sellers applying for financing are typically required to authorize access to their sales and settlement information through official platform APIs (e.g., the Amazon Selling Partner API), thereby ensuring the availability of borrower-specific information and real-time performance data. This dual channel of data availability supports both robust calibration of generative models and tailored credit assessments. Potential regulatory implications primarily concern data privacy, cross-border information sharing, and model transparency. However, these challenges can be systematically managed through secure API protocols, repayment-account control mechanisms, and adherence to established financial data governance standards. Taken together, these considerations suggest that the proposed framework is not only technically deployable but also aligned with the broader evolution of data-driven, inclusive finance.			
	
\end{remark}

\section{Conclusion}\label{sec:conclusion}

This paper introduced a novel SCF risk management framework that leverages a generative model to provide a full distributional understanding of sales, developing a unified analytical framework that simultaneously addresses credit risk assessment and loan size determination, objectives traditionally treated separately. By integrating QRGMM, a quantile-regression-based generative modeling method, with DeepFM to capture complex feature interactions, we have developed a flexible approach capable of estimating entire conditional distributions in complex SCF settings. This allows practitioners to move beyond conventional point estimation and simple risk classification methods, offering a data-driven, distribution-oriented perspective on SCF risk.

Building on the linear quantile regression framework of \cite{hong2023learning}, we establish rigorous asymptotic guarantees for the key risk measures, which serve as universal building blocks for deriving a wide range of risk measures. Numerical experiments on synthetic data confirm that QRGMM$^\dagger$ accurately and stably approximates the underlying distribution, surpasses CWGAN in unconditional and conditional distribution matching, and recovers risk measures that align closely with true benchmarks across the range of loan level space, greatly facilitating loan-sizing decision making in an intuitive way. On real-world datasets, QRGMM$^\dagger$ demonstrates practical applicability, accurately estimating quantiles, generating distributions that mirror real sales patterns, and accurately estimating risk measures.

In summary, this work represents an initial exploration of the use of generative models for operations management, advancing existing SCF risk management methodologies by employing generative models that provide richer distributional insights. However, much work remains to be done in this area. Future research should extend the framework to handle time-series data with seasonal effects and tackle challenges related to high-dimensional datasets. Additionally, while this study mainly focuses on a single product for a seller, future efforts could explore loan portfolio problems by accounting for intricate interdependencies among products, including complementary and substitutable relationships. Moreover, applications in other domains, such as financial risk management and supply chain optimization, could also be investigated. By continuing to refine these generative distributional approaches, the ultimate goal is to support more informed, resilient, and data-driven decision-making in various fields of operations management.



\bibliographystyle{elsarticle-harv} 
\bibliography{ref1}

\appendix

	\section{Proof of Proposition 1}

\begin{proof}[Proof]
	Recall that $\tau_j=\frac{j}{m}$ with $\tau_{j+1}-\tau_j=\frac{1}{m}$. For any $j \in \{1, \ldots, m-1\}$ such that $[\tau_j, \tau_{j+1}] \cap [\tau_{\mathsf{L}} + \frac{1}{m}, \tau_{\mathsf{U}} - \frac{1}{m}] \neq \emptyset$, let $\tau \in [\tau_j, \tau_{j+1}] \cap [\tau_{\mathsf{L}} + \frac{1}{m}, \tau_{\mathsf{U}} - \frac{1}{m}]$.
	For any given $\bx$, define
	\[
	\tilde\alpha
	=\frac{\hat Q(\tau_{j+1}\mid\bx)-\hat Q(\tau\mid\bx)}{\hat Q(\tau_{j+1}\mid\bx)-\hat Q(\tau_j\mid\bx)},
	\qquad
	1-\tilde\alpha
	=\frac{\hat Q(\tau\mid\bx)-\hat Q(\tau_j\mid\bx)}{\hat Q(\tau_{j+1}\mid\bx)-\hat Q(\tau_j\mid\bx)},
	\]
	and because $F_{\hat{Y}}\left(\hat{Q}(\tau\mid\bx)\mid\bx\right)$ is linearly interpolated between $\tau_j=F_{\hat{Y}}\left(\hat{Q}(\tau_j\mid\bx)\mid\bx\right)$ and $\tau_{j+1}=F_{\hat{Y}}\left(\hat{Q}(\tau_{j+1}\mid\bx)\mid\bx\right)$, we have
	$
	F_{\hat Y}\!\left(\hat Q(\tau\mid\bx)\mid\bx\right)
	=\tilde\alpha\,\tau_j+(1-\tilde\alpha)\,\tau_{j+1}.
	$
	Then
	\[
	\begin{aligned}
		&F_{\hat Y}\!\left(\hat Q(\tau\mid\bx)\mid\bx\right)-F_Y\!\left(\hat Q(\tau\mid\bx)\mid\bx\right) \\
		&= \tilde\alpha\bigl[\tau_j-F_Y(\hat Q(\tau_j\mid\bx)\mid\bx)\bigr]
		+(1-\tilde\alpha)\bigl[\tau_{j+1}-F_Y(\hat Q(\tau_{j+1}\mid\bx)\mid\bx)\bigr] \\
		&\quad+ \tilde\alpha\!\left\{F_Y(\hat Q(\tau_j\mid\bx)\mid\bx)-F_Y(\hat Q(\tau\mid\bx)\mid\bx)\right\}
		+(1-\tilde\alpha)\!\left\{F_Y(\hat Q(\tau_{j+1}\mid\bx)\mid\bx)-F_Y(\hat Q(\tau\mid\bx)\mid\bx)\right\}.
	\end{aligned}
	\]
	Using the identity
	\[
	F_Y(\hat Q(\tau_{j+1}\mid\bx)\mid\bx)-F_Y(\hat Q(\tau_j\mid\bx)\mid\bx)
	=\bigl[F_Y(\hat Q(\tau_{j+1}\mid\bx)\mid\bx)-\tau_{j+1}\bigr]+\tfrac{1}{m}
	+\bigl[\tau_j-F_Y(\hat Q(\tau_j\mid\bx)\mid\bx)\bigr],
	\]
	and the triangle inequality, we obtain the bound
	\begin{equation}\label{eq:noC-key}
		\begin{aligned}
			&\bigl|F_{\hat Y}(\hat Q(\tau\mid\bx)\mid\bx)-F_Y(\hat Q(\tau\mid\bx)\mid\bx)\bigr|\\
			&\le\;\; 3\max\!\Bigl\{\bigl|\tau_j-F_Y(\hat Q(\tau_j\mid\bx)\mid\bx)\bigr|,
			\;\bigl|\tau_{j+1}-F_Y(\hat Q(\tau_{j+1}\mid\bx)\mid\bx)\bigr|\Bigr\}
			+\frac{1}{m}.
		\end{aligned}
	\end{equation}

	By Assumption~(2.a), for each grid point $\tau_j$ there exists $\eta_j$ between $Q(\tau_j\mid\bx)$ and $\hat Q(\tau_j\mid\bx)$ such that
	\[
	\bigl|\tau_j-F_Y(\hat Q(\tau_j\mid\bx)\mid\bx)\bigr|= \bigl|F_Y( Q(\tau_j\mid\bx)\mid\bx)-F_Y(\hat Q(\tau_j\mid\bx)\mid\bx)\bigr|
	= f_Y(\eta_j\mid\bx)\,\bigl|Q(\tau_j\mid\bx)-\hat Q(\tau_j\mid\bx)\bigr|.
	\]
	Let
	\[
	\underline f:=\inf_{\tau\in[\tau_\ell,\tau_u]}\inf_{\bx\in\mathcal X} f_Y(F_Y^{-1}(\tau\mid\bx)\mid\bx)>0,
	\qquad
	\overline f:=\sup_{\tau\in[\tau_\ell,\tau_u]}\sup_{\bx\in\mathcal X} f_Y(F_Y^{-1}(\tau\mid\bx)\mid\bx)<\infty,
	\]
	as guaranteed by Assumption~(2.b). Then, for all $j$ with $\tau_j\in[\tau_\ell,\tau_u]$,
	\[
	\bigl|\tau_j-F_Y(\hat Q(\tau_j\mid\bx)\mid\bx)\bigr|
	\le \overline f\;\bigl|Q(\tau_j\mid\bx)-\hat Q(\tau_j\mid\bx)\bigr|.
	\]
	Hence,
	\begin{equation}\label{eq:noC-endpoint}
		\max_{\tau_j\in[\tau_\ell,\tau_u]}
		\bigl|\tau_j-F_Y(\hat Q(\tau_j\mid\bx)\mid\bx)\bigr|
		\;\le\;
		\overline f\;\max_{\tau_j\in[\tau_\ell,\tau_u]}\bigl| Q(\tau_j\mid\bx)-\hat Q(\tau_j\mid\bx)\bigr|.
	\end{equation}
	By Lemma~2,
	$$
	\max_{\tau_j\in[\tau_\ell,\tau_u]}\bigl| Q(\tau_j\mid\bx)-\hat Q(\tau_j\mid\bx)\bigr| \;=\;O_{\mathbb P}\!\left(\frac{1}{\sqrt{n}}\right).
	$$
	Combining this with \eqref{eq:noC-key}–\eqref{eq:noC-endpoint} and taking the supremum over $\tau\in[\tau_\ell+1/m,\tau_u-1/m]$ yields
	\[
	\sup_{\tau\in[\tau_\ell+1/m,\tau_u-1/m]}\bigl|F_{\hat Y}(\hat Q(\tau\mid\bx)\mid\bx)-F_Y(\hat Q(\tau\mid\bx)\mid\bx)\bigr|
	\;=\;O_{\mathbb P}\!\left(n^{-1/2}\right)+O\!\left(m^{-1}\right).
	\]
	Finally, since $y=\hat Q(\tau\mid\bx)$ ranges over the corresponding support $\mathcal{Y}_{m,n}(\bx)$, we conclude
	\[
	\sup_{y\in\mathcal{Y}_{m,n}(\bx)}\bigl|F_{\hat{Y}}(y\mid\bx)-F_Y(y\mid\bx)\bigr|
	\;=\;O_{\mathbb P}\!\left(n^{-1/2}\right)+O\!\left(m^{-1}\right),
	\]
	\qedhere
\end{proof}

\section{Sensitivity of $m$}

\begin{figure}[ht]
	\centering
	\begin{subfigure}{0.48\textwidth}
		\centering
		\includegraphics[width=\textwidth]{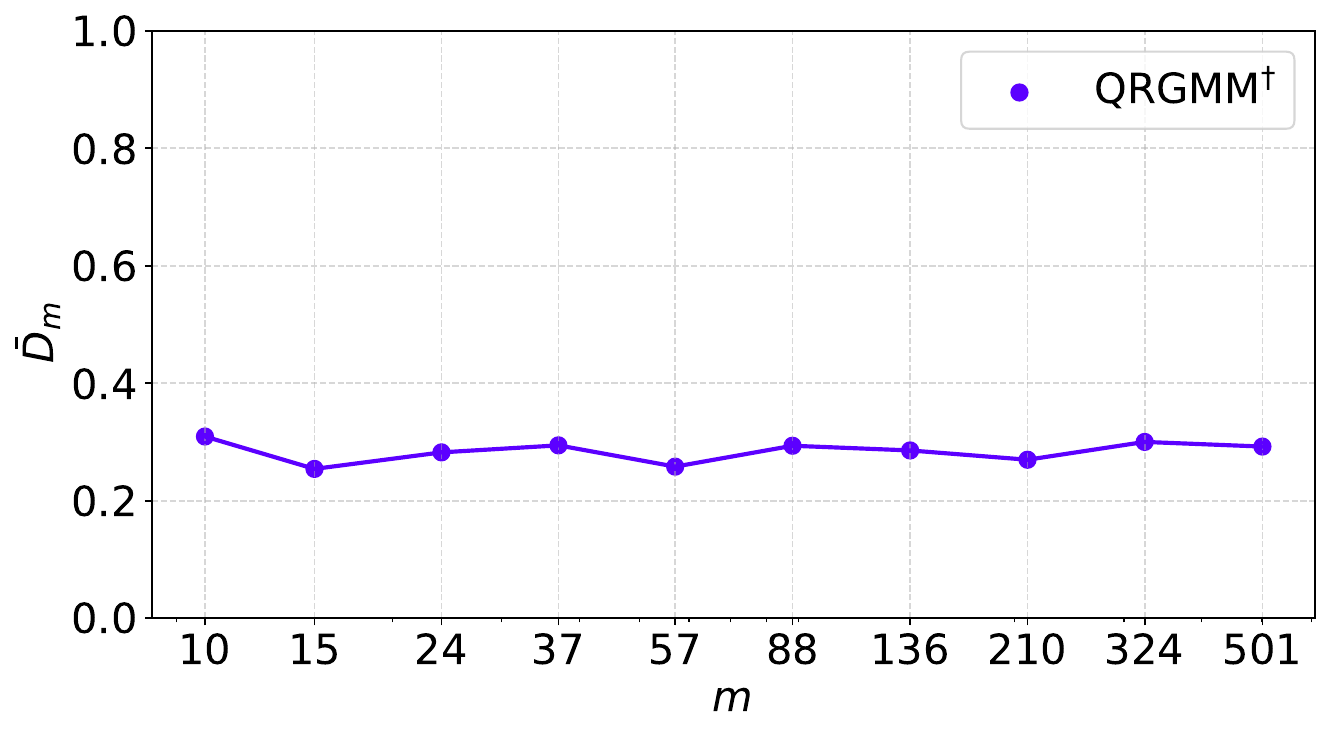}
		\caption{$n=8000$}
		\label{fig:D_vs_m_synthetic}
	\end{subfigure}
	\hfill
	\begin{subfigure}{0.48\textwidth}
		\centering
		\includegraphics[width=\textwidth]{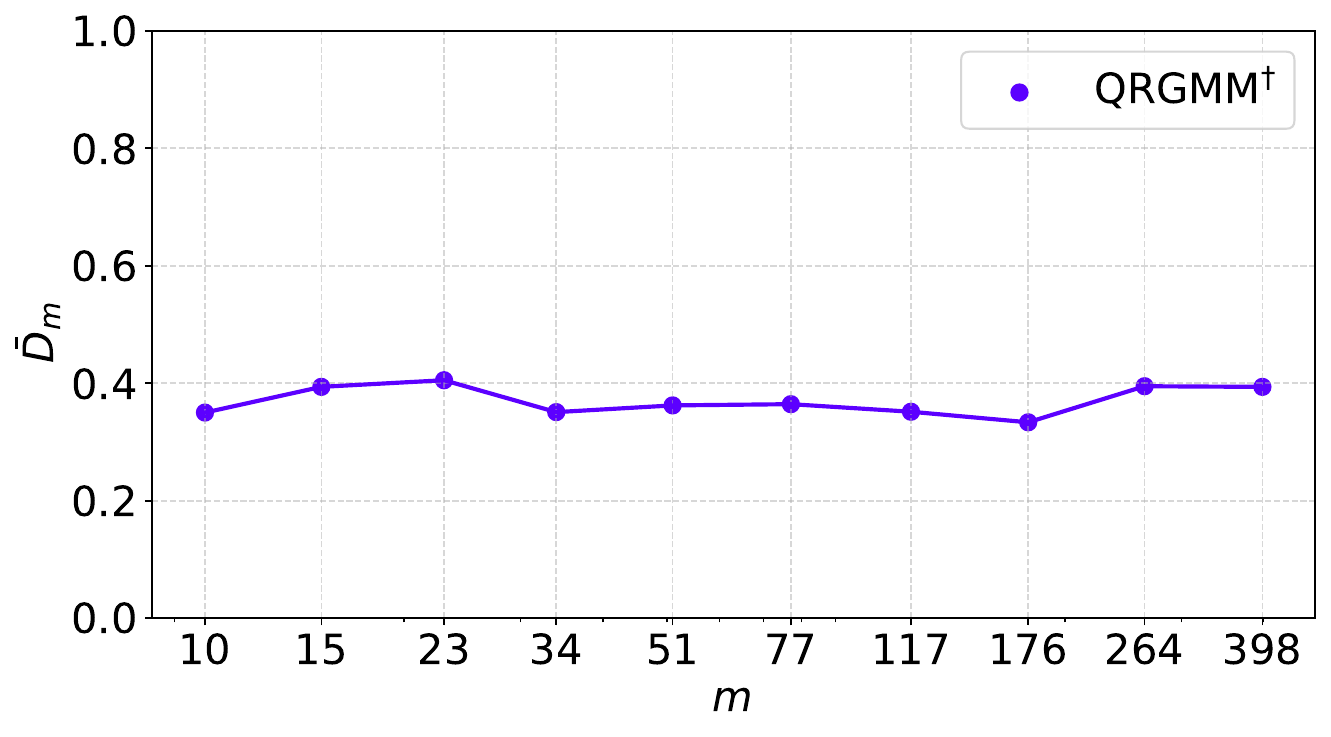}
		\caption{$n=2800$}
		\label{fig:D_vs_m_synthetic_smalldata}
	\end{subfigure}
	\caption{Sensitivity of QRGMM$^\dagger$ to the number of quantile grid points $m$ under different training sample sizes.}
	\label{fig:D_vs_m_sensitivity}
\end{figure}

We examine the sensitivity of QRGMM$^\dagger$ with respect to the number of quantile grid points $m$ using the synthetic FM location–scale shift model from Section~6.1. For each replication, the model was trained under different values of $m$ (logarithmically spaced), with training data sizes $n=8{,}000$ and $n=2{,}800$, corresponding to the training data sizes used in Sections~6.1 and 6.3, respectively. Evaluation was conducted on $10{,}000$ observations of $\hat{Y}(\mathbf{x})$ generated by QRGMM$^\dagger$ at a fixed covariate $\mathbf{x}$, where the Kolmogorov–Smirnov statistic was computed against the true distribution. Each experiment was repeated for 50 replications, and the results were averaged over these replications, denoted by $\bar{D}_m$. Figure~\ref{fig:D_vs_m_sensitivity} displays $\bar{D}_m$ as a function of $m$ for QRGMM$^\dagger$.

The convergence rate analysis in Section~4.3 shows that the error of the linear QRGMM follows the rate $O_{\mathbb P}(n^{-1/2})+O(m^{-1})$, suggesting $m$ should be chosen on the order of $\sqrt{n}$. Our experiments indicate that, after incorporating the deep neural network and factorization machine components in QRGMM$^\dagger$, the performance is not highly sensitive to the choice of $m$. Hence, we adopt $m=\sqrt{n}$ as a robust and convenient choice throughout the paper.

\section{Effects of Covariates Changes on Conditional Distributions and Loan Decisions}

	We use the conditional test evaluations in Section~6.1 to illustrate how covariates influence conditional distributions and subsequent loan sizing decisions. Specifically, we choose two covariate vectors randomly drawn from a uniform distribution over the covariate space to conduct the conditional test. In Section~6.1, the conditional test is conducted at
\[
\mathbf{x}_1 = (0.65,\,0.51,\,0.53,\,0.90,\,0.70,\,0.71,\,0.72,\,0.22,\,0.18,\,0.46,\,50,\,97),
\] 
while here we consider
\[
\mathbf{x}_2 = (0.67,\,0.49,\,0.83,\,0.03,\,0.81,\,0.57,\,0.30,\,0.05,\,0.99,\,0.01,\,51,\,194).
\]

Although both $\mathbf{x}_1$ and $\mathbf{x}_2$ are randomly drawn, they represent substantially different feature combinations, enabling us to examine how such shifts affect the conditional distribution of $Y(\mathbf{x})$ and the resulting loan sizing decision.

\begin{figure}[ht]
	\centering
	\begin{subfigure}{0.48\textwidth}
		\centering
		\includegraphics[width=\textwidth]{Figures/SyntheticData_WD_Y0}
		\caption{$\mathbf{x}_1$}
	\end{subfigure}
	\hfill
	\begin{subfigure}{0.48\textwidth}
		\centering
		\includegraphics[width=\textwidth]{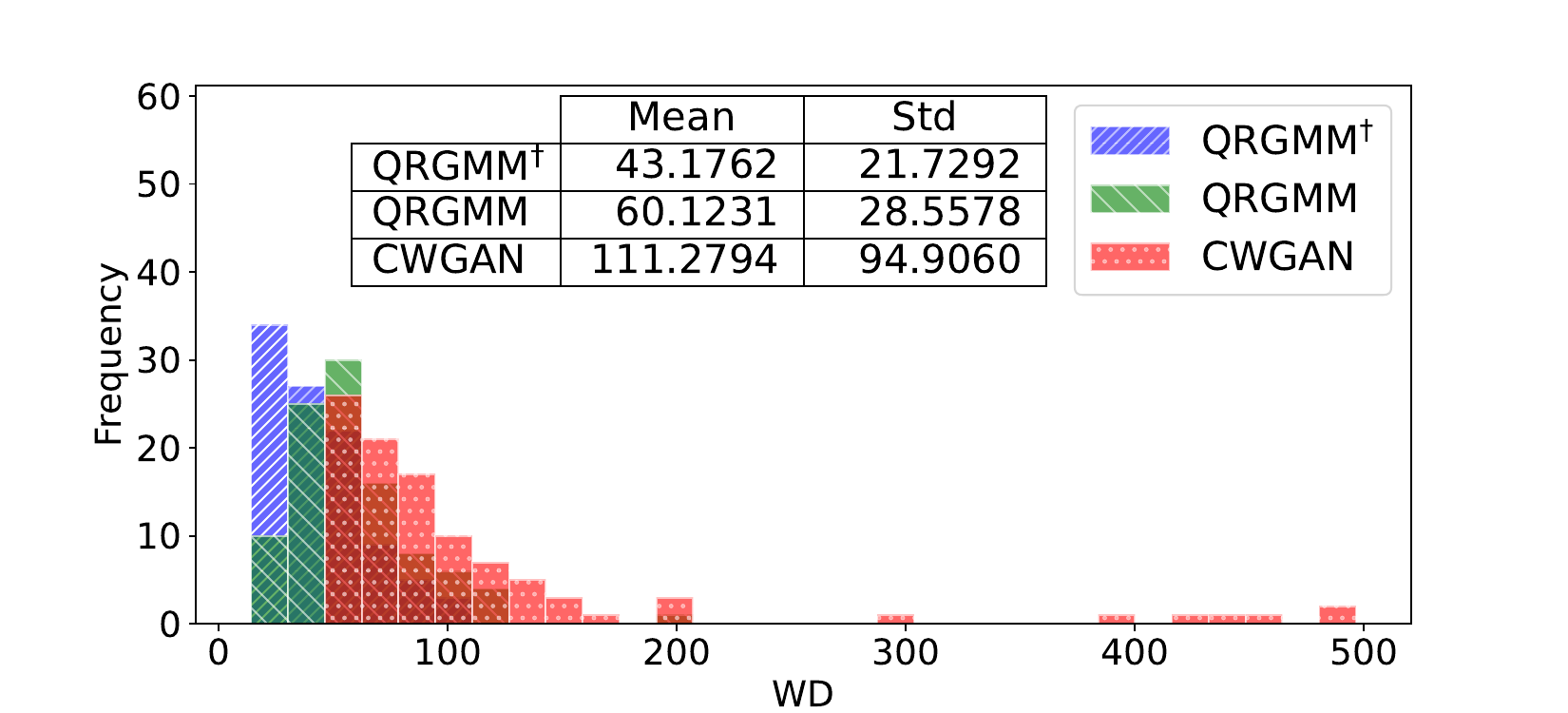}
		\caption{$\mathbf{x}_2$}
	\end{subfigure}
	\caption{WD for conditional test under different covariates.}\label{fig:wd_conditional_2points}
\end{figure}

\begin{figure}[ht]
	\centering
	\begin{subfigure}{0.48\textwidth}
		\centering
		\includegraphics[width=\textwidth]{Figures/SyntheticData_KS_Y0}
		\caption{$\mathbf{x}_1$}
	\end{subfigure}
	\hfill
	\begin{subfigure}{0.48\textwidth}
		\centering
		\includegraphics[width=\textwidth]{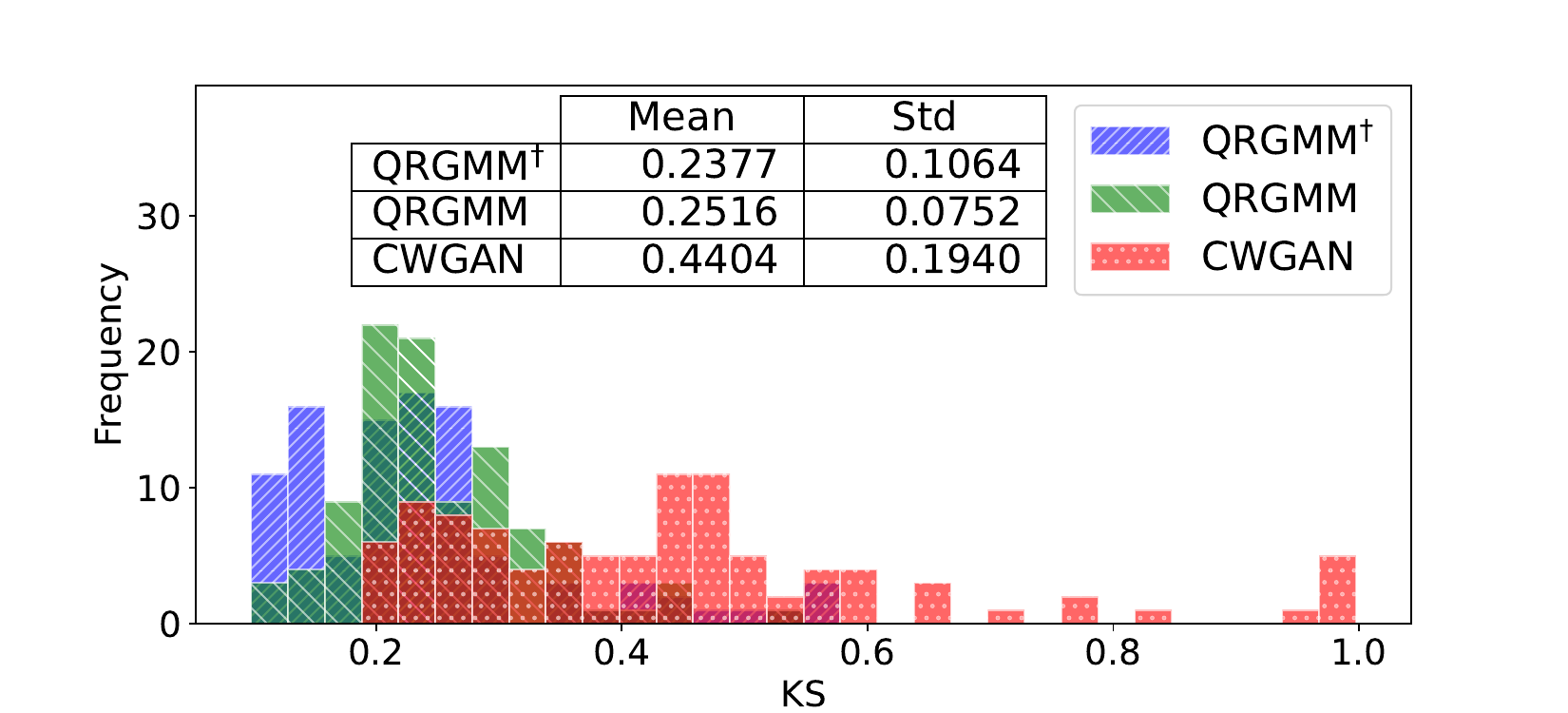}
		\caption{$\mathbf{x}_2$}
	\end{subfigure}
	\caption{KS for conditional test under different covariates.}\label{fig:ks_conditional_2points}
\end{figure}

\begin{figure}[ht]
	\centering
	\begin{subfigure}{1\textwidth}
		\centering
		\includegraphics[width=\textwidth]{Figures/Conditional_Hist_Synthetic}
		\caption{$\mathbf{x}_1$}
	\end{subfigure}
	\hfill
	\begin{subfigure}{1\textwidth}
		\centering
		\includegraphics[width=\textwidth]{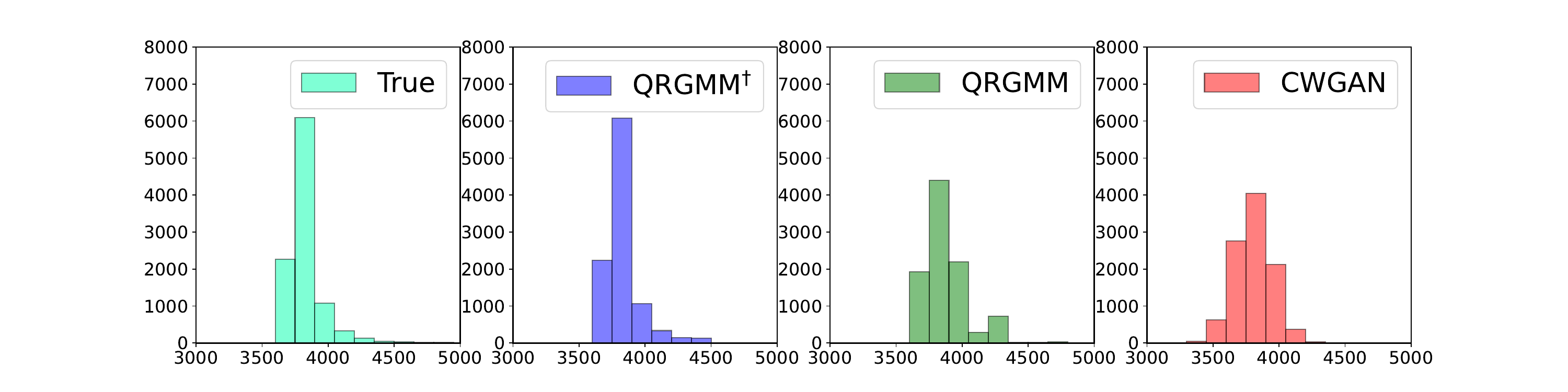}
		\caption{$\mathbf{x}_2$}
	\end{subfigure}
	\caption{Histograms of generated vs.\ true data under different covariates.}\label{fig:hist_2points}
\end{figure}

For this new covariate $\mathbf{x}_2$, we repeat the conditional test evaluations described in Section~6.1. Specifically, we report the sample means and standard deviations of generated conditional datasets (Table~\ref{table:sync_2points}), together with Wasserstein distance (WD) and Kolmogorov--Smirnov statistics (KS), averaged over 100 replications (Figures~\ref{fig:wd_conditional_2points} and \ref{fig:ks_conditional_2points}). Visual comparisons between generated and true conditional distributions (Figure~\ref{fig:hist_2points}), as well as the estimation of risk measures $\{r_1(l), r_2(l), \text{LGD}(l), r_3(l)\}$ (Figure~\ref{fig:risk_2points}), are also presented below.

\setlength{\tabcolsep}{3pt} 
\begin{table}[ht]
	\centering
	\caption{Mean and standard deviation (SD) of generated samples.}\label{table:sync_2points}
	\begin{tabular}{lcccccc}
		\toprule 
		\multirow{2}{*}{\textbf{Model}} 
		& \multicolumn{2}{c}{\textbf{Unconditional Test}} 
		& \multicolumn{2}{c}{\textbf{Conditional $Y(\mathbf{x}_1)$}} 
		& \multicolumn{2}{c}{\textbf{Conditional $Y(\mathbf{x}_2)$}} \\ 
		\cmidrule(l){2-3} \cmidrule(l){4-5} \cmidrule(l){6-7}
		& \textbf{Mean} & \textbf{SD} 
		& \textbf{Mean} & \textbf{SD} 
		& \textbf{Mean} & \textbf{SD} \\ 
		\midrule
		\textbf{Truth} & $4355.4783$ & $1430.8676$ & $3948.9500$ & $150.6615$ & $3832.1420$ & $148.9017$ \\
		\textbf{QRGMM}$^{\dagger}$ & $4350.6580$ & $1412.5260$ & $3948.7998$ & $150.1433$ & $3814.0224$ & $145.0772$ \\
		\textbf{QRGMM} & $4355.9730$ & $1421.0893$ & $4114.5597$ & $221.1148$ & $3867.7647$ & $165.1972$ \\
		\textbf{CWGAN} & $4320.6589$ & $1394.2039$ & $3915.1583$ & $190.8393$ & $3761.0674$ & $141.0797$ \\
		
		\bottomrule
	\end{tabular}
\end{table}
\setlength{\tabcolsep}{6pt}

\begin{figure}[ht]
	\centering
	\begin{subfigure}{1\textwidth}
		\centering
		\includegraphics[width=\textwidth]{Figures/Synthetic_conditional_risk_measure}
		\caption{$\mathbf{x}_1$}
	\end{subfigure}
	\hfill
	\begin{subfigure}{1\textwidth}
		\centering
		\includegraphics[width=\textwidth]{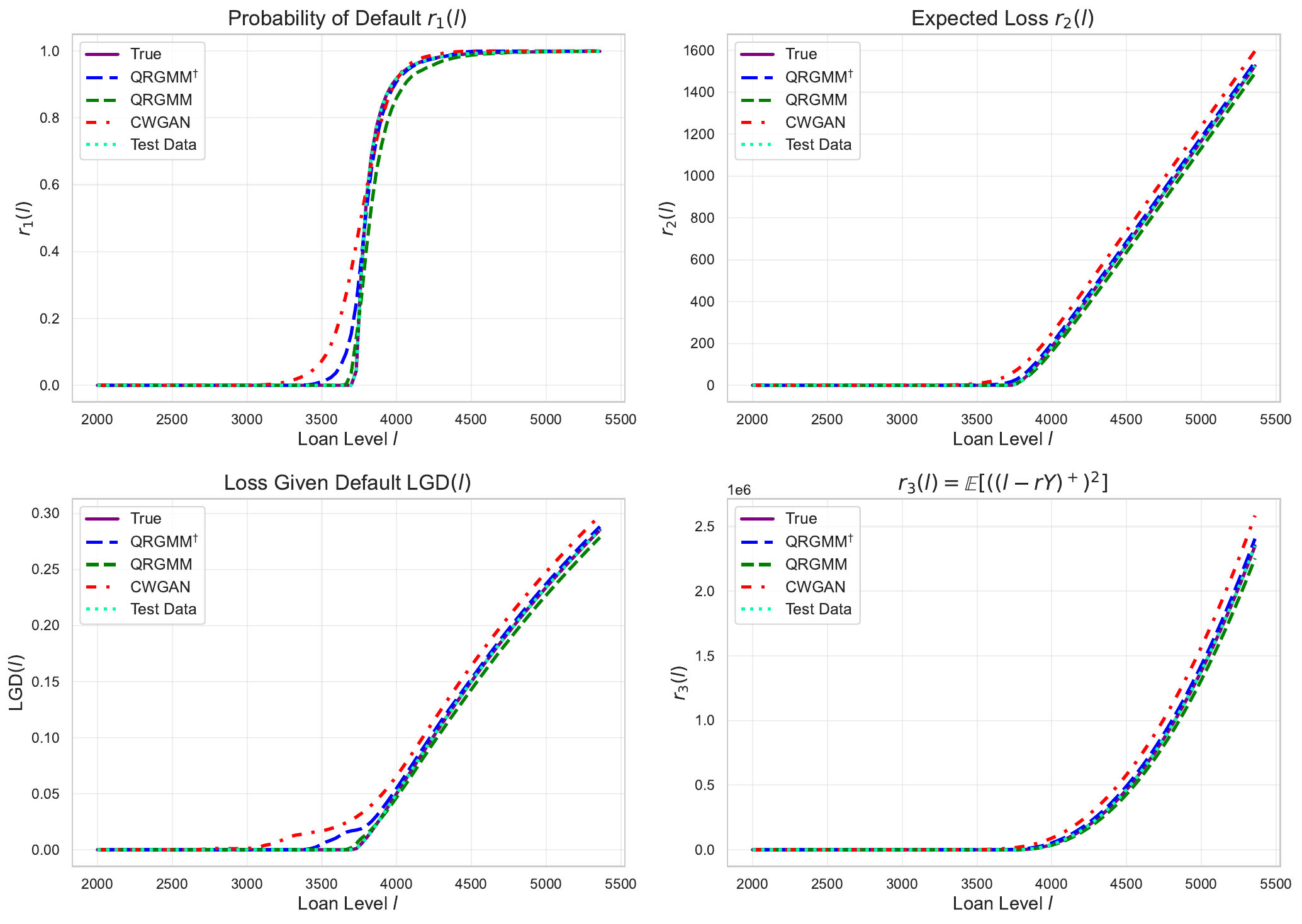}
		\caption{$\mathbf{x}_2$}
	\end{subfigure}
	\caption{Estimation of risk measures across loan levels $l$ under different covariates.}\label{fig:risk_2points}
\end{figure}

\paragraph*{Summary of Results.}  
The results clearly show that the conditional distribution shifts when covariates change, leading to different loan sizing outcomes. For example, the mean and standard deviation of the conditional distribution under $\mathbf{x}_2$ differ from those observed under $\mathbf{x}_1$ (see Table~\ref{table:sync_2points}), which in turn alters the estimated default probabilities and loss measures (see Figure~\ref{fig:risk_2points}). Nevertheless, QRGMM$^\dagger$ continues to deliver stable and accurate estimation: its generated observations align closely with the true conditional distribution across all metrics, and its risk measure curves consistently track the analytical benchmarks. In contrast, both QRGMM and CWGAN exhibit unstable and poor performance, indicating notable limitations in their effectiveness.
These findings underscore both the sensitivity of loan decisions to covariates changes and the robustness of QRGMM$^\dagger$ across these variations.

%
%




%
%
%

\end{document}